\documentclass{article}

\usepackage[main, final]{neurips_2026}

\usepackage[utf8]{inputenc} 
\usepackage[T1]{fontenc}    
\usepackage{hyperref}       
\usepackage{url}            
\usepackage{booktabs}       
\usepackage{amsfonts}       
\usepackage{nicefrac}       
\usepackage{microtype}      
\usepackage{xcolor}         
\usepackage{titletoc}

\title{Forgetting to Improve: \\Principled Data Removal in Active Learning}

\usepackage{amsmath,amsthm}
\usepackage{graphicx}
\usepackage[T1]{fontenc}
\usepackage[utf8]{inputenc}
\usepackage[inline]{enumitem}
\usepackage{dsfont}

\usepackage[english]{babel}
\usepackage{amssymb,amsfonts}
\usepackage{bbm}
\usepackage{nicefrac}
\usepackage{xfrac}
\usepackage{subcaption}
\usepackage{booktabs}
\usepackage{multirow}
\usepackage{tikz}
\usepackage{tikz-3dplot}
\usetikzlibrary{3d,arrows.meta,patterns,decorations.markings}

\usepackage[colorinlistoftodos,bordercolor=purple,backgroundcolor=pink!20,linecolor=pink,textsize=scriptsize]{todonotes}
\usepackage{wrapfig}
\usepackage{microtype}      
\usepackage{parskip}
\usepackage{caption}

\usepackage{thm-restate}
\usepackage{empheq}
\usepackage{color}
\definecolor{cool-blue}{HTML}{1D6996}
\definecolor{cool-purple}{HTML}{5F4690}
\definecolor{cool-teal}{HTML}{38A6A5}
\definecolor{cool-green}{HTML}{0F8554}
\definecolor{dark-green}{rgb}{0.0,0.5,0.0}

\newcommand{\algname}[1]{{\small\sf#1}}
\newcommand{\algnamef}[1]{{\small\textsf{\textbf{#1}}}}
\usepackage{algorithm}
\usepackage[noend]{algpseudocode}
\usepackage[capitalize]{cleveref}
\usepackage[theorems,skins]{tcolorbox}
\usepackage{accents}
\usepackage{tikz-3dplot}
\tdplotsetmaincoords{60}{130}
\usepackage{stackengine}
\stackMath          

\newcommand{\set}[1]{\mathcal{#1}}

\newcommand{\E}{\mathbb{E}}

\newcommand{\tr}{\operatorname{tr}}
\newcommand{\diag}{\operatorname{diag}}

\newcommand{\obj}{\mathcal{J}}
\newcommand{\comp}{\bar{\mathcal{J}}}

\newtheorem{definition}{Definition}
\crefname{definition}{definition}{definitions}
\newtheorem{assumption}{Assumption}
\crefname{assumption}{assumption}{assumptions}

\crefname{theorem}{theorem}{theorems}

\crefname{proposition}{proposition}{propositions}

\crefname{corollary}{corollary}{corollaries}
\newtheorem{lemma}{Lemma}
\crefname{lemma}{lemma}{lemmas}
\theoremstyle{remark}

\crefname{remark}{remark}{remarks}

\usepackage[colorinlistoftodos,bordercolor=purple,backgroundcolor=pink!20,linecolor=pink,textsize=scriptsize]{todonotes}

\definecolor{blue}{RGB}{29, 105, 150}

\newcounter{relctr} 
\everydisplay\expandafter{\the\everydisplay\setcounter{relctr}{0}} 
\newcommand\labelrel[2]{%
  \begingroup
    \refstepcounter{relctr}%
    \stackrel{\textnormal{(\text{\roman{relctr})}}}{\mathstrut{#1}}%
    \originallabel{#2}%
  \endgroup
}
\AtBeginDocument{\let\originallabel\label} 

\definecolor{cool-blue}{HTML}{1D6996}
\definecolor{cool-purple}{HTML}{5F4690}
\definecolor{cool-teal}{HTML}{38A6A5}
\definecolor{cool-green}{HTML}{0F8554}
\definecolor{dark-green}{rgb}{0.0,0.5,0.0}
\hypersetup{
	colorlinks=true,
	linkcolor=cool-blue,
	filecolor=cool-blue,
    citecolor=cool-green,
	urlcolor=cool-blue
}

\author{%
  Manuel Wendl \thanks{Correspondence to: \texttt{mwendl@ethz.ch}.} $^{\,,1,2}$\quad Erik Englesson $^{2}$ \quad Andreas Krause $^{1}$ \quad Carl Henrik Ek $^{2,3}$\\
  $^{1}$ ETH Zürich \quad $^{2}$ University of Cambridge \quad $^{3}$ Karolinska Institutet\\ 
}

\begin{document}

\maketitle

\begin{abstract}
\looseness=-1
The uncertainty of a statistical model is most commonly factorized into an aleatoric and an epistemic part. This factorization changes how predictions are interpreted in downstream decision tasks. Importantly, except for the idealistic scenario with no model mismatch, the quantification is a characteristic of the model and not the data generating process. In this paper, we propose \algname{Forgetting to Improve} a method that reduces this discrepancy by incorporating the task into the modeling framework. Our key insight is to acknowledge that in scenarios of model mismatch, data can have a detrimental effect on the modeling for a specific task. Based on this insight, we propose an influence function for Gaussian process models that allows for principled removal of detrimental data samples. 
We showcase the flexibility of this approach by demonstrating significant improvements across a range of tasks, including Bayesian optimization, model-based reinforcement learning, and transductive learning.
\end{abstract}

\section{Introduction}

\looseness=-1
Gaussian process (GP) models \cite{rasmussen2005gaussian} are a popular choice for probabilistic modeling. Given their strong priors they are especially popular in low data regimes and their interpretable uncertainties make them ideally suited for sequential decision tasks such as Bayesian optimization \cite{chowdhury2017kernelized,garnett2023bayesian} and Reinforcement Learning \cite{deisenroth2011pilco,kakade2020information}. For the modeling assumption of GPs, the inference procedure leads to a natural factorization of the uncertainty into an \emph{aleatoric} and an \emph{epistemic} part. As we often specify the model before seeing the data, there are scenarios where the model does not correspond to the data generation process, which we refer to as \emph{model mismatch}. Box~\cite{box-1976-scien-statis} famously stated ``all models are wrong but some are useful'' which means that these assumptions very rarely match the data generation process. In \emph{active learning} scenarios such as Bayesian optimization, modeling and decision making are often seen as two separate tasks. We first model the data, then make decisions on where to model next. In scenarios of model mismatch this can have severely limiting effects, as we might introduce data, which has detrimental effects on the model that is relevant to the task.

More formally, the interaction of the \emph{aleatoric} and \emph{epistemic} components of the predictive uncertainty is not intrinsic to the data alone, but is induced by the choice of model and the task \cite{smith2025rethinking}. First, in most cases, the choice of model is made a~priori, and may be misaligned for the considered task, while we retain full control over the subset of data used during training.
Secondly, many machine learning tasks only require accurate predictions in a \emph{target region} of the input domain, whereas available observations may originate from a broader region with a potentially different distribution. This distribution shift may lead to a misattribution between the aleatoric and epistemic uncertainty within the target region. As a result, incorporating additional observations can have adverse effects:
\begin{enumerate*}[label=\textbf{(\roman*)}]
\item the aleatoric noise level in the target region may increase significantly when certain observations are included, and
\item may cause some samples to \emph{counteract} variance reduction precisely in the target region.
\end{enumerate*}
\looseness = -1
Both effects arise because observations that improve the global model fit may be uninformative or even detrimental to predictions in the target region.
This is particularly interesting, since in many machine learning tasks, such as \emph{transductive learning} \cite{mackay1992information} and \emph{experimental design}~\cite{chaloner1995bayesian}, the target region in which predictions are made is known beforehand. This knowledge is used in many sequential learning algorithms for BO and RL to select new samples according to information-theoretic acquisition functions, under the assumption that more observed samples strictly improve performance. However, in scenarios of model mismatch, we obtain the described non-trivial model-data dependency of the aleatoric and epistemic uncertainty, which results in potentially detrimental influences of early observed samples for the entire learning process. 
\begin{figure}
    \centering
    \input{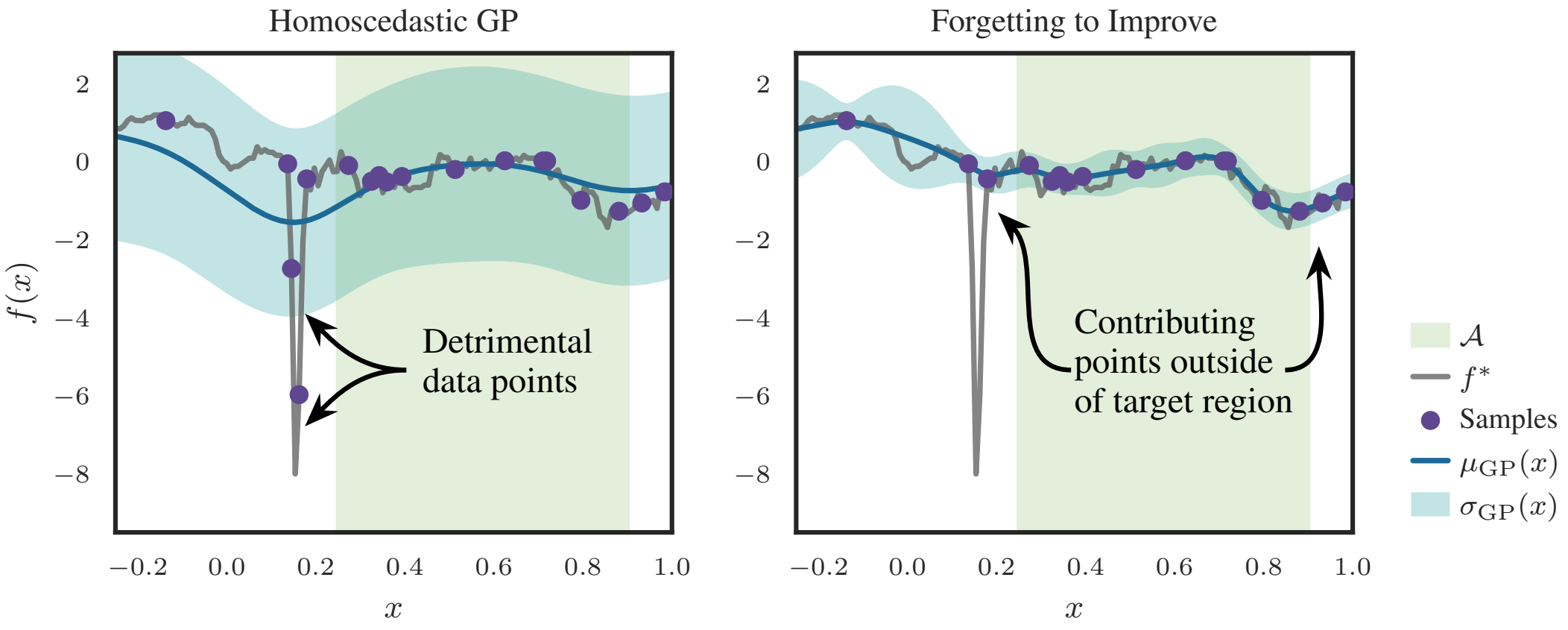}
    \caption{Influence of detrimental data structures based on the Dow Jones data set \cite{ament2024robust}. (Left) Homoscedastic GP absorbs detrimental structures in the noise, resulting in uncertain predictions in \(\set{A}\). (Right) \algname{Forgetting to Improve} (\algname{F2I}) removes detrimental data to improve predictions in \(\set{A}\).}
    \label{fig:demo_fig}
\end{figure}

To address these limitations, we propose \algname{Forgetting to Improve} (\algname{F2I}), a principled data removal algorithm that \emph{``forgets''} data points with detrimental influence on the predictions within the target region in transductive learning tasks. 
As we cannot rely on having access to labels to measure generalization in the target region, we focus instead on reducing predictive uncertainty. 
This is achieved by determining the influence of individual samples on the predictive uncertainty, consisting of the model-data dependent and location-specific epistemic and global aleatoric uncertainties. Samples with detrimental influence are removed to improve predictions and confidence in transductive learning, as visualized for a real-world stock market data set in \Cref{fig:demo_fig}. Especially sequential learning tasks, such as BO and RL, can benefit significantly from this removal paradigm when detrimental samples have been evaluated in the beginning of the optimization process. 

\paragraph{Our contributions} include theoretical grounding and practical evidence of \algname{Forgetting to Improve}:
\begin{itemize}[leftmargin=0.5cm, parsep=0.3em]
    \item We propose a principled measure to quantify the contribution of individual samples to the epistemic and aleatoric components of the predictive uncertainty. 
    \item We introduce the greedy data removal algorithm \algname{F2I} to ``forget'' detrimental data points in transductive learning tasks. In addition, we derive a near-optimality confidence for \algname{F2I} and provide a rigorous analysis by establishing approximation bounds on the predictive uncertainty with respect to the optimal data subset.
    \item We extend \algname{F2I} to \algname{Forgetful BO} and \algname{Forgetful RL}, formulating BO and RL as sequential transductive learning tasks with dynamical target regions.
    \item We conduct an extensive empirical evaluation of \algname{F2I} and \algname{Forgetful BO} on several different transductive and sequential learning tasks.
\end{itemize}

\section{Related Literature}
Prior works on probabilistic models and their use in uncertainty-aware learning algorithms fall broadly into two categories:  
\begin{enumerate*}[label=\textbf{(\roman*)}]
    \item principled methods that focus on choosing the best next unseen sample to improve predictions and
    \item algorithms aiming to improve predictions given a fixed set of samples. 
\end{enumerate*}
 
\paragraph{Targeted Data Acquisition}
\looseness=-1
A canonical line of work focuses on \emph{``where to sample next''}. 
While active learning aims to optimally select informative sampling locations, transductive active learning has additional knowledge about the desired target region, where predictions will be made. 
Early work \cite{mackay1992information} introduced information-theoretic criteria for selecting samples that are most informative for a fixed target region. 
This perspective has recently been revisited empirically for different settings \cite{kothawade2021similar,smith2023prediction,wang2021beyond}, and information-theoretically extended in \cite{seo2000gaussian,yu2006active}. 
Recently, optimality guarantees were derived in \cite{hubotter2024transductive} for these decision rules in transductive learning and BO. 
Closely related ideas appear in optimal sensor placement and Bayesian experimental design, for which acquisition functions have been designed to maximize information gain or reduce posterior uncertainty \cite{huan2024optimal,krause2008near,seo2000gaussian}.
However, these approaches require strict assumptions on well-behaved objectives and focus exclusively on selecting new samples and leveraging all collected data. 
Meaning that once a sample is acquired, it is never revisited, even if it later proves detrimental under model mismatch for the specific task. 
In contrast, our work addresses the complementary and largely unexplored question \emph{``which samples should be removed''}. 

\paragraph{Robustness via Correction and Reweighting}
A large body of work addresses robustness in probabilistic modeling with data preprocessing, alternative likelihoods, or noise modeling. 
Data curation and preprocessing methods \cite{chu2016data} aim to remove corrupted samples before training. 
These methods are typically model-agnostic, assume access to the full dataset beforehand, may introduce unintended biases, and have limited applicability in sequential tasks.
In contrast, robust GP formulations modify the likelihood or noise model to mitigate the influence of detrimental data structures. This includes heavy-tailed likelihoods, such as Student-t \cite{jylanki2011robust}, Laplace \cite{ranjan2016robust}, and Huber losses \cite{algikar2023robust}. 
These have been proven to be effective against global noise misspecification. However, noise can also be locally dependent, which is addressed in a different line of work by introducing noise models to determine input-dependent noise distributions using heteroscedastic \cite{goldberg1997regression,kersting2007most,lazaro2011variational,wang2012gaussian} or twinned GPs \cite{naish2007robust}.
 Although these approaches allow local noise variations, the underlying noise distribution has to be smooth with respect to the input. Additionally, these methods do not take into account task-specific prediction regions and allow modeling structures of interest with the noise model.
More closely related methods remove or down-weight data points. Trimmed likelihood approaches select subsets of data that maximize the marginal likelihood \cite{andrade2023robust,awasthi2022trimmed}, however require a prespecified fraction of detrimental samples \cite{awasthi2022trimmed}, or converge to a locally stationary point without optimality guarantees \cite{andrade2023robust}. Therefore, these methods may unnecessarily remove points for well-behaved objectives or in the target region of our predictions. 
Other strategies rely on residual-based heuristic pruning \cite{li2021robust} or structured noise models that assume clustered outliers \cite{altamirano2023robust}. 
Latent variable and input warping approaches aim to absorb the model mismatch by reshaping the input space \cite{bodin2020modulating,snoek2014input}. 
The latent GP model separates inputs with detrimental structure by introducing an additional dimension and performing computationally expensive posterior sampling with MCMC over the separation, while input warping is only capable of transforming the entire input space globally, ignoring local structures.
Related work also considers modifications in the output space, by assuming biased observations of the form \(y(x) = f(x) + \delta(x) + \epsilon\), where \(\epsilon \sim \mathcal{N}(0,\sigma^2)\) denotes random noise and \(\delta(x)\) captures the sample specific bias \cite{ament2024robust,park2022robust}. 
While both methods introduce learnable data-point-specific biases, \cite{park2022robust} optimizes the negative marginal log likelihood in a single step, which generally does not lead to sparsity in the biases, in contrast to \cite{ament2024robust}.  
Consequently, \cite{bodin2020modulating,ament2024robust} are most closely related to our work, as they induce effects similar to the removal of detrimental samples by distancing them in the input space or adding a large bias in the output space. 
Unlike latent-space or bias-based methods, which implicitly reduce the influence of detrimental samples through additional modeling assumptions, our approach explicitly selects a task-optimal subset of data with theoretical guarantees. 
Furthermore, our per-sample influence is related to classical influence-functions and data-valuation approaches \cite{koh2017understanding,ghorbani2019data}, that quantify single sample attributions on a global estimator or loss. Crucially, our measure enables principled task-specific removals rather than global reweighting. 

\section{Problem Setting}
\looseness=-1
We consider in this work a standard Gaussian process regression setting. Let \(\set{X}\) be the input space and \(f:\set{X} \to \mathbb{R}\) be an unknown function. We model \(f\) as a GP \(f \sim \mathrm{GP}(0,k(\cdot,\cdot))\), with an a priori chosen kernel \(k\). We are given a dataset \(\mathcal{S} \coloneq \{x_i,y_i\}^n_{i=1} \subseteq \set{X} \times \mathbb{R}\), with observations \(y_i = f(x_i) + \epsilon_i\), and \(\epsilon_i \sim \mathcal{N}(0,\sigma_w^2)\). In practice the noise variance \(\sigma_w^2\) depends on the measurement noise and model-data mismatch. The strength of this dependence may be controlled by placing a prior \(p(\sigma_w^2)\) on the noise variance, and solving for the \emph{type II maximum a posteriori} (MAP II) estimate \cite{murphy2023probabilistic}
\begin{equation}\label{eq:MAPII}
    \sigma_w^2(\set{S}) = \arg \max_{\sigma_w^2} \set{L}(\sigma_w^2,\set{S}) \coloneqq \log \prod_{i = 1}^n p(y_i|x_i,\sigma_w^2) + \log p(\sigma_w^2).
\end{equation}
The predictive distribution of a GP trained on the subset \(\set{S}'\subseteq \set{S}\) is Gaussian at any \(x\) with variance 
\begin{equation}\label{eq:variance}
    \sigma_{\set{S}'}^2(x) = k(x,x) - k_{x,\set{S}'}(K + \sigma_w^2(\set{S}')I)^{-1}k_{\set{S}',x} + \sigma_w^2(\set{S}')I ,
\end{equation}
where \(K = k(\set{S}',\set{S}')\) is the Gram matrix over \(\set{S}'\), \(k_{x,\set{S}'} = [k(x,x_i)]_{x_i\in\set{S}'}\), and we explicitly denote the dependence of the noise estimate on the subset by \(\sigma_w^2(\set{S}')\). 

Our goal is to find a subset \(\set{S}'\subseteq \set{S}\) that improves the GP predictive in a target region \(\set{A} \subseteq \set{X}\) without having access to ground truth labels in the entire region. Without labels, it is hard to guide the subset selection to improve the predictive mean. Instead, the \emph{predictive uncertainty} becomes a natural objective, as it characterizes the information content of the posterior distribution in \(\set A\).  Therefore, we aim to \emph{find the subset \(\set{S}'\subseteq \set{S}\) that minimizes the predictive uncertainty in \(\set{A}\)}.
\begin{equation}\label{eq:obj}
\min_{\mathcal{S}'\subseteq\mathcal{S}} \mathcal{J}(\mathcal{S}'), \qquad \mathcal{J}(\set{S}') = \sum_{x \in \set{A}}\sigma_{\set{S}'}^2(x) .
\end{equation}
This closely relates to established principles in \emph{Bayesian Optimization} and \emph{Reinforcement Learning}, where uncertainty guides exploration and balances exploitation (see  \Cref{sec:bo,sec:rl}). 
\section{Forgetting to Improve -- Principled Data Removal}\label{sec:forgetting_to_improve}

We propose \algname{Forgetting to Improve} (\algname{F2I}), a principled data-removal algorithm to find the subset \(\set{S}'\) in \Cref{eq:obj} by greedily removing data points based on our predictive uncertainty influence measure. The influence measure naturally decomposes into the sum of per-example aleatoric and epistemic terms. In this section, we derive these terms, present the full algorithm, and discuss its optimality.

\paragraph{Tracing Epistemic Uncertainty Back to Data}
\looseness = -1
A natural strategy for data selection, deeply rooted in related works of active learning, is to select data with the \emph{maximum gain in information} for the given task. For transductive learning, this reduces to choosing the sample that \emph{minimizes the epistemic uncertainty} in the target region \(\set{A}\) as much as possible.
The subset selection in \Cref{eq:obj} is combinatorial and computationally infeasible for large data sets. Leave-one-out (LOO) methods instead greedily remove the sample with the highest influence, which for example $i$ is defined by $\sigma^2_{\set{S} \setminus \{x_i,y_i\}} - \sigma^2_{\mathcal{S}}$. However, this requires a model refit for each individual removal candidate from the current subset. To solve this, we first generalize the discrete LOO case via a continuous removal direction \(h\) where \(k(x,\set{S},h) = k_{x,\set{S}} - h \odot k_{x,\set{S}}\) in \Cref{prop:removalDirection}. Thus, the predictive uncertainty $\sigma^2_{\mathcal{S}}(h)$ depends on $h$, where $h$ being the zero vector or the standard basis $e_i$ corresponds to $\sigma^2_{\mathcal{S}}$ and $\sigma^2_{\set{S} \setminus \{x_i,y_i\}}$, respectively. Now, to get an efficient approximation to the LOO influence, we do a first-order Taylor expansion of $\sigma^2_{\mathcal{S}}(h)$ around the full dataset \(\set{S}\) ($h=0$), and evaluate at $h=e_i$: $\sigma^2_{\set{S} \setminus \{x_i,y_i\}} - \sigma^2_{\mathcal{S}}  \approx  e_i  \cdot \nabla_{h} \sigma^2_{\mathcal{S}}\rvert_{h=0}$. Thus, we only require one gradient computation rather than several LOO evaluations per removal.

\begin{restatable}{proposition}{propEpistemicInfluence}\label{prop:epistemic_influence}
    Suppose we have a single test point \(\set{A} = \{x\}\) and the corresponding GP prediction with mean \(\mu_{\set{S}}(x)\) and variance \(\sigma_{\set{S}}^2(x)\), for a given set of samples \(\set{S} = \{x_i,y_i\}_{i=1}^n\). The epistemic influence \(u_i(x)\) of an individual sample \(x_i\in\set{S}\) on the predictive uncertainty is defined as the partial derivative of \(\sigma_{\set{S}}^2(x)\) w.r.t. the removal in \(h_i\) for sample \(x_i\in \set{S}\) under fixed aleatoric noise \(\sigma_w^2(\set{S})\)
    \begin{align*}
        &u_i(x) = 2k_{x,x_i} (A^{-1}k_{\set{S},x})_i - 2(A^{-1}k_{\set{S},x})_i(k_{x_i,\set{S}} A^{-1}k_{\set{S},x}), \\
        &\text{where } k_{x,\set{S}} = k(x,\set{S}), \text{ and } A = K+\sigma_w^2(\set{S})I, \text{ with } K=k(\set{S},\set{S}).
    \end{align*}
\end{restatable}
Furthermore, in \Cref{prop:derivative_epistemic_influence} we can interpret our derived influence measure \(u_i\) as the directional derivative of the uncertainty difference for sample removal \(x_i\) in the kernel direction of test points \(x\), unifying both the dependence of task and model in the dependence on \(x\) and \(k(\cdot)\).
\begin{restatable}{proposition}{corrDerivative}\label{prop:derivative_epistemic_influence}
    Suppose the uncertainty difference at \(\set{A} = \{x\}\), when removing sample \(i\) is given by 
    \begin{equation*}
        \Delta_i(x) = \sigma^2_{\set{S}\setminus\{x_i,y_i\}}(x) - \sigma^2_{\set{S}}(x) = \frac{(A^{-1}k_{\set{S},x})_i^2}{A^{-1}_{i,i}}.
    \end{equation*}
    Then, our uncertainty measure \(u\) in \Cref{prop:epistemic_influence} captures the half directional derivative along the kernel direction \(k_{x,\set{S}}\), and \(k_{x,x_i}\) for an individual component of \(u\):
    \begin{equation*}
        k_{x,x_i}\frac{\partial \Delta_i}{\partial k_{x,x_i}} = 2k_{x,x_i}(A^{-1} k_{\set{S},x})_i - 2(A^{-1}k_{\set{S},x})_i(k_{x_i,\set{S}} A^{-1}k_{\set{S},x}) = u_i(x).
    \end{equation*}
\end{restatable}
We refer to \Cref{sec:appendix_epistemic1} for the formal proofs of \Cref{prop:epistemic_influence,prop:derivative_epistemic_influence}.
\looseness=-1
Please note that the absolute uncertainty reduction \(\Delta_i(x)\) is nonnegative due to the quadratic term in the numerator and the positive elements of \(A\), such that also satisfies \(u_i \geq 0\). It acts as a safeguard for the removal algorithm, such that it does not remove samples that are load bearing in explaining structure and reducing epistemic uncertainty within the target region. 

\paragraph{How Data drives Aleatoric Uncertainty under Model Mismatch}
\looseness=-1
The derived epistemic influence measure \(u_i\) captures uncertainty differences, explainable by the kernel \(k\). 
The detrimental structures in the data that cannot be captured by the kernel are modeled using independent white noise \(\sigma_w^2(\set{S})\) in a homoscedastic GP setting. 
This creates a strong dependence of the \emph{training data} on the \emph{model-data mismatch} absorbed by the noise \(\sigma_w^2(\set{S})\). 
We place a Gamma prior \(p(\sigma_w^2|a,b)\) on the noise variance and derive in \Cref{prop:aleatoric_derivative} the derivative of the optimal noise MAP II estimate in the removal direction \(h_i\), that influences both the reproducing kernel Hilbert space (RKHS) and the output space (\Cref{prop:removalDirection}).

\begin{restatable}{proposition}{propAleatoricDerivative}\label{prop:aleatoric_derivative}
Let $A(\sigma_w(\set{S}))=K(\mathcal S,\mathcal S)+\sigma_w^2(\set{S}) I$ and suppose that \(\sigma_w^2(\set{S})\)  is the optimal noise variance of the log MAP II (\Cref{eq:MAPII}) \(\tfrac{\partial}{\partial \sigma_w^2} \set{L}(\sigma_w^2,\set{S},a,b) = \set{D}(\sigma_w^2,\set{S},a,b) =  0\) 
for the samples in \(\set{S}\). Then we determine the derivative of the optimal noise floor \(\sigma_w^2(\set{S})\) into the removal direction \(h_i\) using implicit differentiation \(\tfrac{\partial \sigma_w^2}{\partial h_i}
= -\tfrac{\partial}{\partial h_i} \mathcal{D}(\sigma_w^2,\set{S},a,b)/\tfrac{\partial}{\partial \sigma_w^2} \mathcal{D}(\sigma_w^2,\set{S},a,b)\), which results in
\begin{equation*}
\frac{\partial \sigma_w^2(\set{S})}{\partial h_i} = \frac{ - (A^{-1}y)_i(k_{x_i,\set{S}}(A^{-2}y)) - (A^{-2}y)_i(k_{x_i,\set{S}} (A^{-1}y)) + k_{x_i,\set{S}}(A^{-2})_{\cdot, i} + y_i(A^{-2}y)_i}{-y^\top A^{-3} y + \tfrac12\tr(A^{-2}) - \frac{a-1}{(\sigma_w^2)^2}}.
\end{equation*}
\end{restatable}
Given this sensitivity, we now derive the aleatoric influence measure \(v_i(x)\), given by the partial derivative of the aleatoric contribution of the predictive uncertainty with respect to \(h_i\). This measures how individual removals influence the global noise floor \(\sigma_w^2(\set{S})\) and change the predictive uncertainty.

\begin{restatable}{proposition}{propAleatoricInfluence}\label{prop:aleatoric_influence}
Suppose we have a single test point \(\set{A} = \{x\}\) and the corresponding GP prediction with mean \(\mu_{\set{S}}(x)\) and variance \(\sigma_{\set{S}}^2(x)\), for a given set of samples \(\set{S} = \{x_i,y_i\}_{i=1}^n\). The aleatoric influence \(v_i(x)\) of an individual sample \(x_i\in\set{S}\) on the predictive uncertainty is defined as the partial derivative of all aleatoric noise terms \(\sigma_w^2(\set{S})\) in \(\sigma_{\set{S}}^2(x)\) w.r.t. the removal in \(h_i\) for sample \(x_i\in \set{S}\). 
\begin{align*}
    &v_i(x) = (1 + k_{x,\set{S}}A^{-2}k_{\set{S},x}) \frac{\partial \sigma_w^2(\set{S})}{\partial h_i},
\end{align*}
where the partial derivative of the noise parameter is taken from \Cref{prop:aleatoric_derivative}.
\end{restatable}

We provide the mathematical derivations of \Cref{prop:aleatoric_influence,prop:aleatoric_derivative} in \Cref{sec:appendix_aleatoric}. 

\paragraph{\algnamef{Forgetting to Improve }}
\looseness=-1
\Cref{prop:epistemic_influence,prop:aleatoric_influence} establish how removing samples from $\mathcal{S}$ influences the epistemic and aleatoric components of the predictive uncertainty in a single-point target region \(\set{A}=\{x\}\). 
Building on this, our greedy algorithm \algname{Forgetting to Improve} (\Cref{algo:joint}), iteratively discards samples whose removal reduces the predictive uncertainty in a multi-point target region \(\set{A}\), and terminates once no further uncertainty reduction is indicated. To do this, we extend \Cref{prop:epistemic_influence,prop:aleatoric_influence} in two ways. First, we combine \(u(x)\) and \(v(x)\) to approximate the total change in predictive uncertainty when removing a sample. Here, \(u(x)\) captures the reduction due to information gain in the kernel, while \(v(x)\) reflects changes in the estimated aleatoric noise. In \Cref{sec:noise-ablation}, we show that incorporating the epistemic term \(u\) prevents detrimental removals caused by relying solely on the aleatoric term \(v\). Second, we compute the influence on multi-point target regions by summing the individual influences of the test points \(x \in \mathcal{A}\). Finally, as the aleatoric influence relies on the local type II MAP noise variance estimate around the current data subset, we recompute the estimate after each removal.
\vspace{-0.3cm}
\begin{algorithm}
    \caption{\algname{Forgetting to Improve} (\algname{F2I})}
    \begin{algorithmic}
        \Require Target region \(\set{A}\), all samples \(\set{S}\), and the GP model with kernel \(k\).
        \State Initialize \(t=0\), and \(\set{S}^{(0)} \gets \set{S}\). 
        \State Compute influence measures \(u^{(0)}(\set{A})\), and \(v^{(0)}(\set{A})\) \Comment{\Cref{prop:epistemic_influence,prop:aleatoric_influence}}
        \While{\(\min(u^{(t)}(\set{A}) + v^{(t)}(\set{A})) < 0\)} \Comment{Samples that reduce uncertainty}
            \State Determine worst sample: \(i = \arg \min_j (u_j^{(t)}(\set{A}) + v_j^{(t)}(\set{A}))\) 
            \State Update set of samples \(\set{S}^{(t+1)} \gets \set{S}^{(t)}\setminus x_i\) and GP noise MAP II
            \State Increase \(t\gets t+1\) and compute influences: \(u^{(t)}(\set{A}), v^{(t)}(\set{A})\)\Comment{\Cref{prop:epistemic_influence,prop:aleatoric_influence}}
        \EndWhile\\
        \Return{\(\set{S}' = \set{S}^{(t)}\)}
    \end{algorithmic}\label{algo:joint}
\end{algorithm}
\vspace{-0.3cm}
This greedy algorithm is based on the derivatives of the aleatoric and epistemic uncertainty of \Cref{prop:epistemic_influence,prop:aleatoric_influence} and can hence be interpreted as a ``\emph{reverse}'' orthogonal matching pursuit algorithm \cite{tropp2004greed} for the predictive uncertainty in \(\set{A}\). 
To analyze the realized discrete removals of \algname{F2I}, we consider the normalized set-function \(\bar{\set{J}}(T) = \set{J}(\set{S}) - \set{J}(\set{S}\setminus T)\). Rather than requiring this generally non-monotone set-function to be approximately submodular \cite{das2018approximate,elenberg2018restricted,krause2014submodular}, we bound its optimal value using restricted strong concavity and impose a gradient-dominance condition along the executed removal sequence satisfiying the \emph{Polyak-\L{}ojasiewicz} inequality \cite[sec.~3.2]{karimi2016linear}. Let $m_R,M_R>0$ denote the restricted strong concavity
and restricted smoothness constants, as formalized in \Cref{def:rsc} in \Cref{sec:appendix_submodular}.
\begin{assumption}
\label{assumption:gradient-dominance}
Let $T'=\set{S}\setminus\set{S}'$ denote the observations removed by
Algorithm~1, and suppose $q=|T'|\geq 1$. Define the removal sets
$T_k=\set{S}\setminus\set{S}^{(k)}$ and the initial gradient
\smash{$g^{(0)}=-u^{(0)}(\mathcal A)-v^{(0)}(\mathcal A)$}.
Let \smash{$g_{(1)}^{(0)}\geq\cdots\geq g_{(|\set{S}|)}^{(0)}$} denote its
ordered components, and define for every $k=0,\ldots,q-1$,
\begin{equation*}
\|u^{(k)}(\mathcal A)+v^{(k)}(\mathcal A)\|_\infty^2
\geq\frac{2m_R}{q}
\left[U_q-\bar{\mathcal J}(T_k)\right]_+, \quad U_q=\min\left\{\mathcal J(\set{S}),\;
\sum_{i=1}^{q}g_{(i)}^{(0)}-\frac{m_Rq}{2}\right\}.
\end{equation*}
\end{assumption}
This gives a conditional near-optimality bound in \Cref{thrm:forgetting_optimality}, under the checkable \Cref{assumption:gradient-dominance}.
\begin{restatable}{theorem}{thrmforgettingoptimality}
    \label{thrm:forgetting_optimality}
    Let $\bar{\mathcal J}(T)=\mathcal J(\set{S})-\mathcal J(\set{S}\setminus T)$
and let $T'=\set{S}\setminus\set{S}'$ be the $q\geq1$ samples removed by
Algorithm~1. Suppose the inner MAP II estimate for $\sigma_w^2$
is well defined and differentiable at all considered retained datasets.
Assume that $-\mathcal J$ satisfies the $m_R$-restricted strong
concavity inequality of Definition~1 for $L=\emptyset$ and all
$U\subseteq\set{S}$ with $|U|=q$, and that
\Cref{assumption:gradient-dominance} holds.
Then
\begin{equation*}
\bar{\mathcal J}(T')\geq
\bigl(1-e^{-\gamma}\bigr)
\max_{\substack{T^*\subseteq\set{S}\\|T^*|=|T'|}}
\bar{\mathcal J}(T^*),
\qquad \gamma=\rho\frac{m_R}{M_R},\vspace{-0.3cm}
\end{equation*}
for the restricted smoothness $M_R>0$ constant in \(\rho=\min_{0\leq k<q}
\frac{2M_R\bigl[\mathcal J(\set{S}^{(k)})
-\mathcal J(\set{S}^{(k+1)})\bigr]}
{\|u^{(k)}(\mathcal A)+v^{(k)}(\mathcal A)\|_\infty^2}>0\).
\end{restatable}

\paragraph{Proof Sketch}
A detailed outline of the proof is given in \Cref{sec:appendix_submodular}.
First, we define $\bar{\set{J}}$ and show that $- u_i(\set{A}) - v_i(\set{A})$ is the gradient of $-\set{J}$ in \Cref{lemma:rkhs-derivative}. Under \Cref{assumption:gradient-dominance}, \(U_q\) upper-bounds the optimal gain of any removal set of cardinality $q$. The realized-step coefficient
$\rho$ and \Cref{assumption:gradient-dominance} then guarantee that each removal reduces the remaining gap to $U_q$. Iterating over $q$ yields \Cref{thrm:forgetting_optimality}.

\section{Forgetful Bayesian Optimization}\label{sec:bo}
We extend \algname{F2I} to Bayesian optimization by viewing BO as a sequence of transductive learning problems, each defined by a dynamically shrinking region of interest. The key idea is to remove samples that distort posterior uncertainty exactly where the acquisition function relies on it. This allows us to improve optimization performance without modifying the acquisition function itself. Therefore, we consider an unknown objective function \(f(x):\set{X}\to \mathbb R\), with the goal of solving the global optimization problem \(x^* = \arg \max_{x\in \set{X}} f(x)\) under a limited evaluation budget. 

\paragraph{Model Mismatch in Bayesian Optimization}
The model-mismatch effects studied in \Cref{sec:forgetting_to_improve} are particularly problematic in Bayesian optimization. Since BO relies on uncertainty-aware acquisition functions, such as the GP Upper Confidence Bound (GP-UCB) \cite{srinivas2009gaussian}, logarithmic Expected Improvement (logEI) \cite{ament2023unexpected}, Expected Improvement (EI) and Probability of Improvement \cite{jones1998efficient,movckus1974bayesian}, misleading posterior uncertainty can directly harm exploration and exploitation decisions. Prior work addresses this issue through alternative likelihoods \cite{martinez2017robust}, latent or warped input spaces \cite{bodin2020modulating,snoek2014input}, tree-based structure \cite{jenatton2017bayesian}, or preference-based feedback \cite{gonzalez2017preferential}. In contrast to these approaches, which modify the model or acquisition function, we address the model mismatch by explicitly revising the set of observations used to form the posterior uncertainty, in each iteration.

\paragraph{Target Regions in Bayesian Optimization}
In order to address this problem, we propose to view BO as sequential transductive learning \cite{hubotter2024transductive}.
At each iteration, only a subset of the search space can plausibly contain the maximum. Improving predictive uncertainty outside this subset is therefore unnecessary and can even be harmful under model mismatch. We formalize this intuition by defining a region of interest and applying \algname{F2I} for this region. The GP-UCB algorithm \cite{srinivas2009gaussian} provides a natural way to define such regions of interest. At each iteration \(t\), future samples are selected by maximizing the upper confidence bound within regions where improvement is possible; specifically, where the upper confidence bound exceeds the current maximum lower confidence bound.
Improving the model predictions on \(\set{A}_t\) with \algname{F2I} results in better uncertainty bounds, which is traded for a potentially increased epistemic uncertainty outside of \(\set{A}_t\) due to sample removals. 
Instead of keeping \(\set{A}_t\) fixed throughout the removal process, we leverage the improved confidence bounds for the samples \smash{\(\set{S}_t^{(k)}\)} after the \(k\)-th removal to update \smash{\(\set{A}_t^{(k)}\)} accordingly. 
However, removing samples may increase uncertainty outside \smash{\(\set{A}_t^{(k)}\)}, potentially creating artificial improvement regions. To prevent this effect, we enforce that the region of interest can only shrink over forgetting iterations and propose the following intersecting update for the target region
\begin{equation}\label{eq:regionOfInterestBORec}
    \set{A}_t^{(k)} \gets \{x_i \in \set{X} : \underbrace{\mu_{\set{S}_t^{(k)}}(x_i) + \beta_t\sigma_{\set{S}_t^{(k)}}(x_i)}_{\mathrm{UCB}(x_i)} > \max_{x_m\in\set{X}} \underbrace{\mu_{\set{S}_t^{(k)}}(x_m) - \beta_t\sigma_{\set{S}^{(k)}}(x_m)}_{\mathrm{LCB}(x_m)}\} \cap \set{A}_t^{(k-1)}.
\end{equation}
Accordingly chosen constants \(\beta_t\) ensure that the model is well calibrated \cite{srinivas2009gaussian} and therefore \(\set{A}_t\) contains all points that could outperform the current best estimate. We combine the data removal algorithm \algname{F2I} (\Cref{algo:joint}) with the recursive target region update in a standard BO pipeline. The full algorithm \algname{Forgetful BO} is detailed in \Cref{algo:forgetful_bo} in \Cref{sec:appendix_bo}. 

\section{Forgetful Reinforcement Learning}\label{sec:rl}
\looseness=-1
Building on \algname{F2I}, we investigate by introducing \algname{Forgetful RL} how the data removal paradigm mitigates model-data mismatch in model-based reinforcement learning.
 We consider a discounted Markov Decision Process (MDP) \((\mathcal{X}, \mathcal{A}, p, r, \gamma, \rho_0)\) with state space \(\set{X}\), action space \(\set{A}\), transition probabilities \(p(x' | x,a)\), rewards \(r(x,a)\), discounting \(\gamma \in (0,1)\), and initial state distribution \(\rho_0\).
 In MBRL \cite{deisenroth2011pilco,janner2019trust}, a world model is learned to simulate trajectories, enabling in each episode \(n=1,\dots, N\) the optimization of the policy \(\pi_n\) to maximize the sum of rewards \(R(\pi_n) = \mathbb{E}_{\pi_n}\left[\sum_{t=0}^\infty \gamma^t r(x_t,a_t)\right]\) in this learned environment, using the corresponding state-action value function \cite{sutton1998reinforcement}
 \begin{align}
    Q^{\pi_n}(x,a) &= \mathbb{E}_{\pi_n}\left[\sum_{t=0}^{\infty} \gamma^t r(x_t,a_t) \mid x_0 = x, a_0 = a \right], \quad  V^{\pi_n}(x) = \E_{a \sim \pi_n(x)} [Q^{\pi_n}(x,a)].
\end{align}
 Since we only have limited data at episode \(n\), the convergence and quality of the resulting policy critically depend on both the accuracy and calibrated uncertainty of this probabilistic model. Since the planning trajectories are sampled from the predictive distribution of the world model, low uncertainty in the trajectory estimates is desired. Motivated by this, we transfer the idea of a dynamically determined target region, which is defined as the set of state-action pairs \((x,a)\) with potential policy improvement and from visitation state distributions of equal or better policies. In each iteration, we apply the data removal paradigm of \algname{F2I} for the target region of improvement and reduce the predictive uncertainty of the model for the state-action pairs in the target region 
\begin{align}\label{eq:rl_region_of_interest}
    \set{A}_n = \{(x,a) \in \set{X}\times\set{A}: \bar{Q}^{\pi_n}(x,a) \geq \min_{x_m \sim \rho_0} \underline{V}^{\pi_n}(x_m) \land \bar{Q}^{\pi_n}(x,a) \geq \underline{V}^{\pi_n}(x)\},
\end{align}
where \(\bar{Q}^{\pi_n}\) is an upper confidence bound on all possible state-action value functions, as well as \(\underline{V}^{\pi_n}\) being a lower confidence bound under the currently available data. 
We describe our GP-based \algname{Forgetful RL} implementation, based on the MBPO implementation of \cite{Pineda2021MBRL} in detail in \Cref{sec:appendix_rl}.

\section{Experiments}
\begin{wrapfigure}{r}{0.41\textwidth}
\vspace{-1em}
    \includegraphics[width=\linewidth]{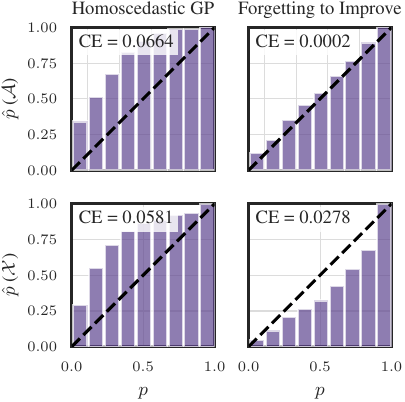}
    \caption{Empirical confidence \(\hat p\) compared to nominal \(p\). Calibration curves for (left) homoscedastic GP and improvement for (right) \algname{F2I} on (top) the target region \(\set{A}\) and (bottom) the entire domain \(\set{X}\) for the example in \Cref{fig:demo_fig}.}
    \label{fig:confidence_plot}
\end{wrapfigure}
\looseness=-1
We evaluate the empirical performance of \algname{F2I}, \algname{Forgetful BO} and \algname{Forgetful RL}, using our \href{https://github.com/ManuelWendl/ForgettingToImprove}{implementation}, based on GPyTorch \cite{gardner2018gpytorch}, BoTorch \cite{balandat2020botorch}, and MBRL-Lib \cite{Pineda2021MBRL}. Our results focus on 
\begin{enumerate*}[label=\textbf{(\roman*)}]
    \item transductive regression tasks, improving predictions in a priori known target regions; 
    \item BO, by removing detrimental data structures in the sequential learning tasks; and
    \item MBRL under global kernel misspecification of the GP world model. 
\end{enumerate*}

\begin{figure}
    \centering
    \includegraphics[width=0.8\linewidth]{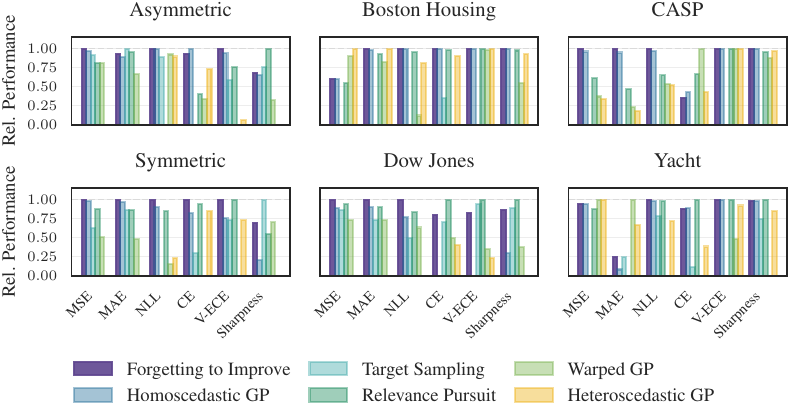}
    \caption{Benchmark comparison (mean over 100 training sample draws). \algname{F2I} achieves competitive predictive sharpness while providing strong overall accuracy and calibration across the evaluations.}    \label{fig:transductive_results}
\end{figure}

\paragraph{Transductive Learning}
\looseness=-1
For many real-world applications, the desired prediction location is known a priori, which we refer to as target region \(\set{A}\), while the training data \(\set{S}\) is distributed across the entire domain \(\set{X}\). In the following \emph{transductive learning experiments}, our aim is to improve the predictions in \(\set{A}\). Therefore, we evaluate the prediction performance in \(\set{A}\) based on the mean squared error (MSE), mean absolute error (MAE), test negative log-likelihood (NLL) \cite{deshpande2022you}, and \emph{sharpness}, which is defined as the expected predictive variance in \(\set{A}\). Additionally, we evaluate calibration, based on previous work that extends calibration measures from classification to regression  \cite{kuleshov2018accurate,levi2022evaluating,gawlikowski2023survey}, to quantify the coverage of predicted confidence intervals (CE) \cite{kuleshov2018accurate}, as well as the local calibration comparing the observed residuals (ECE) \cite{levi2022evaluating}. 
Based on these metrics, we compare \algname{F2I} with a homoscedastic and a heteroscedastic GP \cite{kersting2007most}, domain \emph{warping GP} \cite{snoek2014input} and \emph{relevance pursuit} \cite{ament2024robust} on four transductive learning tasks. These include two synthetic test functions, described in \Cref{sec:appendix_experiments_transductive}, and two real-world tasks using the Boston Housing dataset \cite{harrison1978hedonic} and the intra-day data from the Dow Jones Industrial Average index \cite{ament2024robust}.
We report in \Cref{fig:transductive_results} the mean relative performance across 100 different randomly sampled training points using the hyperparameters listed in \Cref{sec:appendix_experiments_transductive}. The relative performance is computed with respect to the best and worst performing method. The numerical results, including standard deviations, are provided in \Cref{sec:appendix_experiments_transductive}.
As we minimize predictive uncertainty with our objective in \Cref{eq:obj}, \algname{F2I} achieves competitive predictive sharpness in \(\set{A}\). We additionally observe that the calibration measures (CE, ECE) benefit significantly from the tighter confidence bounds. \Cref{fig:confidence_plot} shows this improvement for the task in \Cref{fig:demo_fig}. The top row shows the calibration improvement for unseen test samples in the target region \(\set{A}\), while the right column shows the tradeoff: underconfident predictions on the full domain \(\set{X}\). Remarkably, we also observe a consistent improvement in the accuracy metrics, which we do not explicitly optimize.

\begin{figure}
    \centering
    \includegraphics[width=0.86\linewidth]{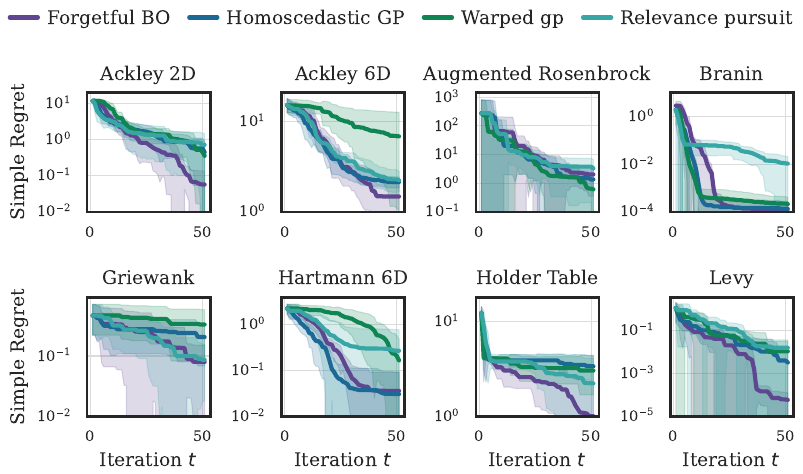}
    \caption{Simple regret across different test functions. \algname{Forgetful BO} consistently achieves low simple regret, while competing baselines exhibit strong variability across tasks.}
    \label{fig:bo_results}
\end{figure}

\paragraph{Bayesian Optimization}
\looseness=-1
We next investigate how these task-oriented prediction improvements translate to the sequential learning process of \algname{Forgetful BO}. The BO performance is measured using \emph{simple regret}, the difference between the best evaluation up to the current iteration \(t\) and the optimal achievable function value \(r(t) = f(x^*) - \max_{1\leq t' \leq t}f(x_t)\). Accumulating simple regret over all \(T\) iterations yields the cumulative regret, which is provided in \Cref{sec:appendix_additional_experiments}. We compare \algname{Forgetful BO} to a homoscedastic GP model, the \emph{warped} GP model \cite{snoek2014input} and the \emph{relevance pursuit} model \cite{ament2024robust}, based on the BoTorch implementations \cite{balandat2020botorch}, as well as additional baselines in \Cref{sec:appendix_experiments_bo}, using the Matérn \(5/2\) kernel and a LogNormal\((0,1)\) for the prior lengthscale. In \Cref{fig:bo_results}, we use the GP-UCB acquisition function \cite{srinivas2009gaussian}, while we also report results for the log expected improvement acquisition function \cite{ament2023unexpected} in \Cref{sec:appendix_experiments_bo}. We provide statistical results for \(10\) runs, with random initializations and a limited evaluation budget of \(T = 50\). We report in \Cref{fig:bo_results} the mean simple regret and standard deviation for \(8\) test-functions \cite{simulationlib}, using the default domains, provided together with the dimensionality in \Cref{sec:appendix_experiments_bo}. 
\looseness = -1
By forgetting detrimental data during the optimization task, \algname{Forgetful BO} achieves a consistently low simple regret, while the performance of competing baselines strongly varies between test-functions. To demonstrate the advantage of removals, we visualize this behavior for a two dimensional test function in \Cref{fig:removal_visualization} of \Cref{sec:appendix_bo}. In \Cref{sec:appendix_experiments_bo_timimg,sec:appendix_experiments_bo}, we additionally demonstrate superior performance of \algname{Forgetful BO} for noisy tasks with sparse outliers, and the computational efficiency of \algname{Forgetful BO}.

\begin{figure}[h]
    \centering
    \includegraphics[width=0.87\linewidth]{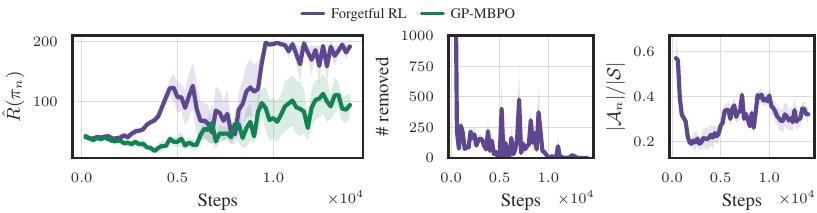}
     \includegraphics[width=0.87\linewidth]{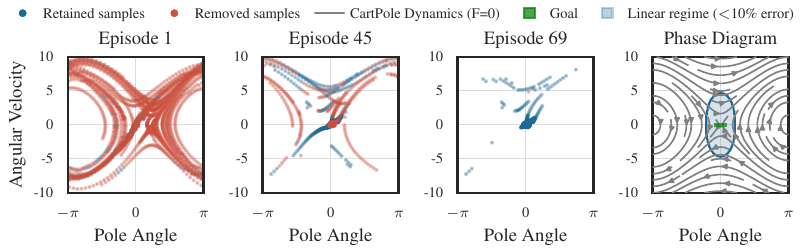}
    \caption{(Top) \algname{Forgetful RL} succeeds to learn \emph{Cartpole Balance} with a linear GP model, while standard methods suffer from retained nonlinear observations. (Bottom) Visualization of removed and retained samples for \algname{Forgetful RL}. The method keeps task-relevant data in the linear regime.}
    \label{fig:rl_linear_kernel}
\end{figure}

\looseness=-1
\paragraph{Reinforcement Learning} 
\looseness=-1
We demonstrate the full potential of data removal under model-data mismatch, by learning the overall {\em non-linear} \emph{Cartpole Balance} task using a {\em linear} GP for model-based reinforcement learning. From classical control theory, it is well known that the task relevant part of the dynamics around the upright pole position can be described with a linearized model \cite{aastrom2021feedback}. Since RL learns from trial and error and may fail in early episodes, the system may not remain in this approximately linear domain. We show in \Cref{fig:rl_linear_kernel} (top) that \algname{Forgetful RL} converges using a linear kernel (left) by constantly removing detrimental nonlinear experiences (middle) for predictions in the stabilizing target region \(\set{A}_n\) (right). In contrast MBPO \cite{janner2019trust}, which retains all samples, does not solve the task. We visualize the behavior of \algname{Forgetful RL} to delete nonlinear experiences in \Cref{fig:rl_linear_kernel} (bottom). The retained data points lie within the dynamics part relevant to stabilizing at the upward pole position. \algname{Forgetful RL} automatically removes samples from the attractive regime of the undesired stable poles, compared with the dynamics phase diagram in the right-hand subplot.

\section{Conclusion} \looseness=-1
This work revisits the fundamental assumption on the irreducibility of aleatoric uncertainty in GP regression, from which the well known paradigm of incorporating all available data to improve the predictive uncertainty originates. For task-specific predictions, we demonstrate that this assumption breaks under model mismatch. To address this, we introduce \algname{Forgetting to Improve}, a principled data-removal algorithm that explicitly minimizes predictive uncertainty in the \emph{task-specific target region} by identifying and removing samples that are detrimental to these areas under model mismatch. We further extend this paradigm to sequential decision making, reformulating BO and RL as iterative transductive learning problems. Our improvements on regression, BO and RL tasks highlight \emph{data removal} as a complementary and largely underexplored direction for improving uncertainty-aware learning.


\newpage
\section*{Acknowledgement}
\looseness=-1
We would like to thank Amanda Prorok, Jasmine Bayrooti, Massimiliano Tamborski and Yarden As for insightful discussions during this project. We thank the anonymous reviewers for their valuable comments and suggestions. This work was supported as a part of NCCR Automation, a National Centre of Competence in Research, funded by the Swiss National Science Foundation (grant number 51NF40\_225155). This work was partially supported by the Wallenberg AI, Autonomous Systems and Software Program (WASP) founded by the Knut and Alice Wallenberg Foundation.

\bibliographystyle{plain}
\bibliography{bibliography}

@inproceedings{koh2017understanding,
  title={Understanding black-box predictions via influence functions},
  author={Koh, Pang Wei and Liang, Percy},
  booktitle={International conference on machine learning},
  year={2017},
}

@inproceedings{
hubotter2024transductive,
title={Transductive Active Learning: Theory and Applications},
author={Jonas H{\"u}botter and Bhavya Sukhija and Lenart Treven and Yarden As and Andreas Krause},
booktitle={The Thirty-eighth Annual Conference on Neural Information Processing Systems},
year={2024},
}

@inproceedings{yu2006active,
    author = {Yu, Kai and Bi, Jinbo and Tresp, Volker},
    title = {Active learning via transductive experimental design},
    year = {2006},
    booktitle = {Proceedings of the 23rd International Conference on Machine Learning},
    }

@article{ginsbourger2025fast,
  title={Fast calculation of gaussian process multiple-fold cross-validation residuals and their covariances},
  author={Ginsbourger, David and Sch{\"a}rer, C{\'e}dric},
  journal={Journal of Computational and Graphical Statistics},
  year={2025},
}

@article{quellette1981schur,
title = {Schur complements and statistics},
journal = {Linear Algebra and its Applications},
year = {1981},
author = {Diane Valérie Ouellette},
}

@book{rasmussen2005gaussian,
    author = {Rasmussen, Carl Edward and Williams, Christopher K. I.},
    title = {Gaussian Processes for Machine Learning},
    publisher = {The MIT Press},
    year = {2005},
}

@book{krantz2002implicit,
  title={The implicit function theorem: history, theory, and applications},
  author={Krantz, Steven George and Parks, Harold R},
  year={2002},
  publisher={Springer Science \& Business Media}
}

@inproceedings{kuleshov2018accurate,
  title={Accurate uncertainties for deep learning using calibrated regression},
  author={Kuleshov, Volodymyr and Fenner, Nathan and Ermon, Stefano},
  booktitle={International conference on machine learning},
  year={2018},
}

@article{levi2022evaluating,
  title={Evaluating and calibrating uncertainty prediction in regression tasks},
  author={Levi, Dan and Gispan, Liran and Giladi, Niv and Fetaya, Ethan},
  journal={Sensors},
  year={2022},
}

@article{gawlikowski2023survey,
  title={A survey of uncertainty in deep neural networks},
  author={Gawlikowski, Jakob and Tassi, Cedrique Rovile Njieutcheu and Ali, Mohsin and Lee, Jongseok and Humt, Matthias and Feng, Jianxiang and Kruspe, Anna and Triebel, Rudolph and Jung, Peter and Roscher, Ribana and others},
  journal={Artificial Intelligence Review},
  year={2023},
}

@article{srinivas2009gaussian,
  title={Gaussian process optimization in the bandit setting: No regret and experimental design},
  author={Srinivas, Niranjan and Krause, Andreas and Kakade, Sham M and Seeger, Matthias},
  journal={arXiv preprint arXiv:0912.3995},
  year={2009}
}

@inproceedings{chu2016data,
  title={Data cleaning: Overview and emerging challenges},
  author={Chu, Xu and Ilyas, Ihab F and Krishnan, Sanjay and Wang, Jiannan},
  booktitle={Proceedings of the 2016 international conference on management of data},
  year={2016}
}

@article{jylanki2011robust,
  title={Robust Gaussian Process Regression with a Student-t Likelihood.},
  author={Jyl{\"a}nki, Pasi and Vanhatalo, Jarno and Vehtari, Aki},
  journal={Journal of Machine Learning Research},
  volume={12},
  year={2011}
}

@article{ranjan2016robust,
  title={Robust Gaussian process modeling using EM algorithm},
  author={Ranjan, Rishik and Huang, Biao and Fatehi, Alireza},
  journal={Journal of Process Control},
  volume={42},
  year={2016},
}

@article{algikar2023robust,
  title={Robust Gaussian process regression with Huber likelihood},
  author={Algikar, Pooja and Mili, Lamine},
  journal={arXiv preprint arXiv:2301.07858},
  year={2023}
}

@article{naish2007robust,
  title={Robust regression with twinned Gaussian processes},
  author={Naish-Guzman, Andrew and Holden, Sean},
  journal={Advances in neural information processing systems},
  volume={20},
  year={2007}
}

@article{wang2012gaussian,
  title={Gaussian process regression with heteroscedastic or non-Gaussian residuals},
  author={Wang, Chunyi and Neal, Radford M},
  journal={arXiv preprint arXiv:1212.6246},
  year={2012}
}

@article{goldberg1997regression,
  title={Regression with input-dependent noise: A Gaussian process treatment},
  author={Goldberg, Paul and Williams, Christopher and Bishop, Christopher},
  journal={Advances in neural information processing systems},
  volume={10},
  year={1997}
}

@inproceedings{lazaro2011variational,
  title={Variational Heteroscedastic Gaussian Process Regression.},
  author={L{\'a}zaro-Gredilla, Miguel and Titsias, Michalis K},
  booktitle={International conference on machine learning},
  year={2011}
}

@article{awasthi2022trimmed,
  title={Trimmed maximum likelihood estimation for robust generalized linear model},
  author={Awasthi, Pranjal and Das, Abhimanyu and Kong, Weihao and Sen, Rajat},
  journal={Advances in Neural Information Processing Systems},
  volume={35},
  year={2022}
}

@inproceedings{andrade2023robust,
  title={Robust Gaussian process regression with the trimmed marginal likelihood},
  author={Andrade, Daniel and Takeda, Akiko},
  booktitle={Uncertainty in Artificial Intelligence},
  year={2023},
}

@article{li2021robust,
  title={Robust Gaussian process regression based on iterative trimming, Astron. Comput., 36, 100483},
  author={Li, ZZ and Li, L and Shao, Z},
  journal={arXiv preprint arXiv:2011.11057},
  year={2021}
}

@article{park2022robust,
  title={Robust Gaussian process regression with a bias model},
  author={Park, Chiwoo and Borth, David J and Wilson, Nicholas S and Hunter, Chad N and Friedersdorf, Fritz J},
  journal={Pattern Recognition},
  volume={124},
  year={2022},
}

@article{altamirano2023robust,
  title={Robust and conjugate Gaussian process regression},
  author={Altamirano, Matias and Briol, Fran{\c{c}}ois-Xavier and Knoblauch, Jeremias},
  journal={arXiv preprint arXiv:2311.00463},
  year={2023}
}

@inproceedings{bodin2020modulating,
  title={Modulating surrogates for Bayesian optimization},
  author={Bodin, Erik and Kaiser, Markus and Kazlauskaite, Ieva and Dai, Zhenwen and Campbell, Neill and Ek, Carl Henrik},
  booktitle={International Conference on Machine Learning},
  year={2020},
}

@inproceedings{kersting2007most,
  title={Most likely heteroscedastic Gaussian process regression},
  author={Kersting, Kristian and Plagemann, Christian and Pfaff, Patrick and Burgard, Wolfram},
  booktitle={Proceedings of the 24th international conference on Machine learning},
  year={2007}
}

@article{ament2024robust,
  title={Robust Gaussian processes via relevance pursuit},
  author={Ament, Sebastian and Santorella, Elizabeth and Eriksson, David and Letham, Ben and Balandat, Maximilian and Bakshy, Eytan},
  journal={Advances in Neural Information Processing Systems},
  volume={37},
  year={2024}
}

@article{mackay1992information,
  title={Information-based objective functions for active data selection},
  author={MacKay, David JC},
  journal={Neural computation},
  volume={4},
  year={1992},
}

@inproceedings{smith2023prediction,
  title={Prediction-oriented bayesian active learning},
  author={Smith, Freddie Bickford and Kirsch, Andreas and Farquhar, Sebastian and Gal, Yarin and Foster, Adam and Rainforth, Tom},
  booktitle={International conference on artificial intelligence and statistics},
  year={2023},
}

@article{kothawade2021similar,
  title={Similar: Submodular information measures based active learning in realistic scenarios},
  author={Kothawade, Suraj and Beck, Nathan and Killamsetty, Krishnateja and Iyer, Rishabh},
  journal={Advances in Neural Information Processing Systems},
  volume={34},
  year={2021}
}

@inproceedings{wang2021beyond,
  title={Beyond Marginal Uncertainty: How Accurately can Bayesian Regression Models Estimate Posterior Predictive Correlations?},
  author={Wang, Chaoqi and Sun, Shengyang and Grosse, Roger},
  booktitle={International Conference on Artificial Intelligence and Statistics},
  year={2021},
}

@inproceedings{snoek2014input,
  title={Input warping for Bayesian optimization of non-stationary functions},
  author={Snoek, Jasper and Swersky, Kevin and Zemel, Rich and Adams, Ryan},
  booktitle={International conference on machine learning},
  year={2014},
}

@inproceedings{seo2000gaussian,
  title={Gaussian process regression: Active data selection and test point rejection},
  author={Seo, Sambu and Wallat, Marko and Graepel, Thore and Obermayer, Klaus},
  booktitle={Proceedings of the IEEE-INNS-ENNS international joint conference on neural networks. Neural computing: New challenges and perspectives for the New Millennium},
  volume={3},
  year={2000},
}

@article{krause2008near,
  title={Near-optimal sensor placements in Gaussian processes: Theory, efficient algorithms and empirical studies.},
  author={Krause, Andreas and Singh, Ajit and Guestrin, Carlos},
  journal={Journal of Machine Learning Research},
  volume={9},
  year={2008}
}

@article{huan2024optimal,
  title={Optimal experimental design: Formulations and computations},
  author={Huan, Xun and Jagalur, Jayanth and Marzouk, Youssef},
  journal={Acta Numerica},
  volume={33},
  year={2024},
}

@book{garnett2023bayesian,
  title={Bayesian optimization},
  author={Garnett, Roman},
  year={2023},
  publisher={Cambridge University Press}
}

@inproceedings{chowdhury2017kernelized,
  title={On kernelized multi-armed bandits},
  author={Chowdhury, Sayak Ray and Gopalan, Aditya},
  booktitle={International Conference on Machine Learning},
  year={2017},
}

@article{kakade2020information,
  title={Information theoretic regret bounds for online nonlinear control},
  author={Kakade, Sham and Krishnamurthy, Akshay and Lowrey, Kendall and Ohnishi, Motoya and Sun, Wen},
  journal={Advances in Neural Information Processing Systems},
  volume={33},
  year={2020}
}

@inproceedings{deisenroth2011pilco,
  title={PILCO: A model-based and data-efficient approach to policy search},
  author={Deisenroth, Marc and Rasmussen, Carl E},
  booktitle={Proceedings of the 28th International Conference on machine learning (ICML-11)},
  year={2011}
}

@article{chaloner1995bayesian,
  title={Bayesian experimental design: A review},
  author={Chaloner, Kathryn and Verdinelli, Isabella},
  journal={Statistical science},
  year={1995},
}

@article{tropp2004greed,
  title={Greed is good: Algorithmic results for sparse approximation},
  author={Tropp, Joel A},
  journal={IEEE Transactions on Information theory},
  volume={50},
  year={2004},
}

@article{elenberg2018restricted,
  title={Restricted strong convexity implies weak submodularity},
  author={Elenberg, Ethan R and Khanna, Rajiv and Dimakis, Alexandros G and Negahban, Sahand},
  journal={The Annals of Statistics},
  volume={46},
  year={2018},
}

@article{das2018approximate,
  title={Approximate submodularity and its applications: Subset selection, sparse approximation and dictionary selection},
  author={Das, Abhimanyu and Kempe, David},
  journal={Journal of Machine Learning Research},
  volume={19},
  year={2018}
}

@article{krause2014submodular,
  title={Submodular function maximization.},
  author={Krause, Andreas and Golovin, Daniel},
  journal={Tractability},
  volume={3},
  year={2014}
}

@inproceedings{jenatton2017bayesian,
  title={Bayesian optimization with tree-structured dependencies},
  author={Jenatton, Rodolphe and Archambeau, Cedric and Gonz{\'a}lez, Javier and Seeger, Matthias},
  booktitle={International Conference on Machine Learning},
  year={2017},
}

@inproceedings{gonzalez2017preferential,
  title={Preferential bayesian optimization},
  author={Gonz{\'a}lez, Javier and Dai, Zhenwen and Damianou, Andreas and Lawrence, Neil D},
  booktitle={International Conference on Machine Learning},
  year={2017},
}

@article{martinez2017robust,
  title={Robust Bayesian optimization with Student-t likelihood},
  author={Martinez-Cantin, Ruben and McCourt, Michael and Tee, Kevin},
  journal={arXiv preprint arXiv:1707.05729},
  year={2017}
}

@article{ament2023unexpected,
  title={Unexpected improvements to expected improvement for bayesian optimization},
  author={Ament, Sebastian and Daulton, Samuel and Eriksson, David and Balandat, Maximilian and Bakshy, Eytan},
  journal={Advances in Neural Information Processing Systems},
  volume={36},
  year={2023}
}

@article{jones1998efficient,
  title={Efficient global optimization of expensive black-box functions},
  author={Jones, Donald R and Schonlau, Matthias and Welch, William J},
  journal={Journal of Global optimization},
  volume={13},
  year={1998},
}

@inproceedings{movckus1974bayesian,
  title={On Bayesian methods for seeking the extremum},
  author={Mo{\v{c}}kus, Jonas},
  booktitle={IFIP Technical Conference on Optimization Techniques},
  year={1974},
}

@article{deshpande2022you,
  title={Are you using test log-likelihood correctly?},
  author={Deshpande, Sameer K and Ghosh, Soumya and Nguyen, Tin D and Broderick, Tamara},
  journal={arXiv preprint arXiv:2212.00219},
  year={2022}
}

@article{harrison1978hedonic,
  title={Hedonic housing prices and the demand for clean air},
  author={Harrison Jr, David and Rubinfeld, Daniel L},
  journal={Journal of environmental economics and management},
  volume={5},
  year={1978},
}

@inproceedings{balandat2020botorch,
  title = {{BoTorch: A Framework for Efficient Monte-Carlo Bayesian Optimization}},
  author = {Balandat, Maximilian and Karrer, Brian and Jiang, Daniel R. and Daulton, Samuel and Letham, Benjamin and Wilson, Andrew Gordon and Bakshy, Eytan},
  booktitle = {Advances in Neural Information Processing Systems 33},
  year = 2020,
  url = {http://arxiv.org/abs/1910.06403}
}

@article{gardner2018gpytorch,
  title={Gpytorch: Blackbox matrix-matrix gaussian process inference with gpu acceleration},
  author={Gardner, Jacob and Pleiss, Geoff and Weinberger, Kilian Q and Bindel, David and Wilson, Andrew G},
  journal={Advances in neural information processing systems},
  volume={31},
  year={2018}
}

@MISC{simulationlib, 
 author = {Surjanovic, S. and Bingham, D.}, 
 title = {Virtual Library of Simulation Experiments: Test Functions and Datasets}, 
 howpublished = {Retrieved January 11, 2026, from \url{http://www.sfu.ca/~ssurjano}} 
}

@inproceedings{ghorbani2019data,
  title={Data shapley: Equitable valuation of data for machine learning},
  author={Ghorbani, Amirata and Zou, James},
  booktitle={International conference on machine learning},
  year={2019},
}

@book{sutton1998reinforcement,
  title={Reinforcement learning: An introduction},
  author={Sutton, Richard S and Barto, Andrew G and others},
  volume={1},
  year={1998},
  publisher={MIT press Cambridge}
}

@article{janner2019trust,
  title={When to trust your model: Model-based policy optimization},
  author={Janner, Michael and Fu, Justin and Zhang, Marvin and Levine, Sergey},
  journal={Advances in neural information processing systems},
  volume={32},
  year={2019}
}

@Article{Pineda2021MBRL,
  author  = {Luis Pineda and Brandon Amos and Amy Zhang and Nathan O. Lambert and Roberto Calandra},
  journal = {Arxiv},
  title   = {MBRL-Lib: A Modular Library for Model-based Reinforcement Learning},
  year    = {2021},
  url     = {https://arxiv.org/abs/2104.10159},
}

@inproceedings{haarnoja2018soft,
  title={Soft actor-critic: Off-policy maximum entropy deep reinforcement learning with a stochastic actor},
  author={Haarnoja, Tuomas and Zhou, Aurick and Abbeel, Pieter and Levine, Sergey},
  booktitle={International conference on machine learning},
  year={2018},
}

@article{friedman2010regularization,
  title={Regularization paths for generalized linear models via coordinate descent},
  author={Friedman, Jerome H and Hastie, Trevor and Tibshirani, Rob},
  journal={Journal of statistical software},
  volume={33},
  year={2010}
}

@article{kuhn2018automatic,
  title={Automatic exploration of machine learning experiments on openml},
  author={K{\"u}hn, Daniel and Probst, Philipp and Thomas, Janek and Bischl, Bernd},
  journal={arXiv preprint arXiv:1806.10961},
  year={2018}
}

@article{bischl2017openml,
  title={OpenML benchmarking suites and the OpenML100},
  author={Bischl, Bernd and Casalicchio, Giuseppe and Feurer, Matthias and Hutter, Frank and Lang, Michel and Mantovani, Rafael G and Van Rijn, Jan N and Vanschoren, Joaquin},
  journal={stat},
  volume={1050},
  year={2017}
}

@inproceedings{schur2023lifelong,
  title={Lifelong bandit optimization: no prior and no regret},
  author={Schur, Felix and Kassraie, Parnian and Rothfuss, Jonas and Krause, Andreas},
  booktitle={Uncertainty in Artificial Intelligence},
  year={2023},
}

@article{perrone2018scalable,
  title={Scalable hyperparameter transfer learning},
  author={Perrone, Valerio and Jenatton, Rodolphe and Seeger, Matthias W and Archambeau, C{\'e}dric},
  journal={Advances in neural information processing systems},
  volume={31},
  year={2018}
}

@article{rothfuss2021meta,
  title={Meta-learning reliable priors in the function space},
  author={Rothfuss, Jonas and Heyn, Dominique and Krause, Andreas and others},
  journal={Advances in Neural Information Processing Systems},
  volume={34},
  year={2021}
}

@inproceedings{smith2025rethinking,
  title={Rethinking Aleatoric and Epistemic Uncertainty},
  author={Smith, Freddie Bickford and Kossen, Jannik and Trollope, Eleanor and Van Der Wilk, Mark and Foster, Adam and Rainforth, Tom},
  booktitle={International Conference on Machine Learning},
  year={2025},
}

@book{murphy2023probabilistic,
  title={Probabilistic machine learning: Advanced topics},
  author={Murphy, Kevin P},
  year={2023},
  publisher={MIT press}
}

@article{box-1976-scien-statis,
	title        = {Science and Statistics},
	author       = {George E. P. Box},
	year         = 1976,
	journal      = {Journal of the American Statistical Association},
	volume       = 71,
}

@book{nesterov2013introductory,
  title={Introductory lectures on convex optimization: A basic course},
  author={Nesterov, Yurii},
  volume={87},
  year={2013},
  publisher={Springer Science \& Business Media}
}

@book{aastrom2021feedback,
  title={Feedback systems: an introduction for scientists and engineers},
  author={{\AA}str{\"o}m, Karl Johan and Murray, Richard},
  year={2021},
  publisher={Princeton university press}
}

@inproceedings{o2018uncertainty,
  title={The uncertainty bellman equation and exploration},
  author={O’Donoghue, Brendan and Osband, Ian and Munos, Remi and Mnih, Volodymyr},
  booktitle={International conference on machine learning},
  year={2018}
}

@article{wendl2026safe,
  title={Safe Exploration via Policy Priors},
  author={Wendl, Manuel and As, Yarden and Prajapat, Manish and Pollak, Anton and Coros, Stelian and Krause, Andreas},
  journal={The Fourteenth International Conference on Learning Representations},
  year={2026}
}

@article{vignola2026sampling,
  title={Sampling-Based Safe Reinforcement Learning},
  author={Vignola, Luca and Lee, Bruce D and Prajapat, Manish and Wendl, Manuel and Zeilinger, Melanie and Krause, Andreas and As, Yarden},
  journal={arXiv preprint arXiv:2605.19469},
  year={2026}
}

@inproceedings{karimi2016linear,
  title={Linear convergence of gradient and proximal-gradient methods under the polyak-{\l}ojasiewicz condition},
  author={Karimi, Hamed and Nutini, Julie and Schmidt, Mark},
  booktitle={Joint European conference on machine learning and knowledge discovery in databases},
  pages={795--811},
  year={2016},
  organization={Springer}
}


\newpage
\appendix
\startcontents[appendices]
\section*{Appendix Contents}
\printcontents[appendices]{}{1}{\setcounter{tocdepth}{2}}
\newpage
\section{Epistemic and Aleatoric influence proofs}
In this appendix we provide the detailed proofs for (i) the removal direction for leave-one-out in \Cref{sec:appendix_removalDirection}, (ii) the derivation of the epistemic and (ii) the aleatoric influence in \Cref{sec:appendix_epistemic1,sec:appendix_aleatoric}.

\subsection{LOO -- Removal Direction}\label{sec:appendix_removalDirection}

\begin{restatable}{proposition}{propRemovalDirec}\label{prop:removalDirection}
Let \(k\) be a positive-definite kernel with RKHS \(\mathcal H_k\) and feature map \(\phi(x)=k(x,\cdot)\). For a dataset \(\mathcal S=\{(x_j,y_j)\}_{j=1}^n,\)
define the kernel vector and targets by \(k_{\mathcal S}(x)\coloneqq [k(x,x_1),\dots,k(x,x_n)]^\top,\qquad
y\coloneqq [y_1,\dots,y_n]^\top.\)
Fix an index \(i\), and define the continuous removal direction \(h_i\in[0,1]\) by scaling the \(i\)-th basis function \(\phi(x_i)\) by \(1-h_i\). Equivalently,
\begin{equation*}
    k_{\mathcal S}(x;h_i)\coloneqq k_{\mathcal S}(x)-h_i\,k(x,x_i)e_i, \qquad y(h_i)\coloneqq y-h_i\,y_i e_i,
\end{equation*}
where \(e_i\) is the \(i\)-th standard basis vector. Then, for fixed aleatoric noise \(\sigma_w^2\), setting \(h_i=1\) recovers the leave-one-out predictor:
\begin{equation*}
\mu_{\mathcal S}(x;1)=\mu_{\set{S} \setminus \{x_i,y_i\}}(x),
\qquad
\sigma^2_{\mathcal S}(x;1)=\sigma^2_{\set{S} \setminus \{x_i,y_i\}}(x).
\end{equation*}
\end{restatable}

\begin{proof}
Write the regularized kernel matrix as
\begin{equation}
A(h_i)\coloneqq K(h_i)+\sigma_w^2 I,
\end{equation}
and the corresponding predictor in kernel form as
\begin{align}
\mu_{\mathcal S}(x;h_i) & =k_{\mathcal S}(x;h_i)^\top A(h_i)^{-1}y(h_i),\\
\sigma^2_{\mathcal S}(x;h_i) & =k(x,x)-k_{\mathcal S}(x;h_i)^\top A(h_i)^{-1}k_{\mathcal S}(x;h_i)+\sigma_w^2.
\end{align}

At \(h_i=1\), the \(i\)-th basis function is removed, so the \(i\)-th row and column of \(K(1)\) vanish. After permuting the \(i\)-th coordinate to the last position, \(A(1)\) has the block form
\begin{equation}
A(1)=
\begin{bmatrix}
K_{\set{S} \setminus \{x_i,y_i\}}+\sigma_w^2 I & 0\\
0^\top & \sigma_w^2
\end{bmatrix},
\qquad
A(1)^{-1}=
\begin{bmatrix}
(K_{\set{S} \setminus \{x_i,y_i\}}+\sigma_w^2 I)^{-1} & 0\\
0^\top & \sigma_w^{-2}
\end{bmatrix},
\end{equation}
\begin{equation}
k_{\mathcal S}(x;1)=
\begin{bmatrix}
k_{\set{S} \setminus \{x_i,y_i\}}(x)\\
0
\end{bmatrix},
\qquad
y(1)=
\begin{bmatrix}
y_{\set{S} \setminus \{x_i,y_i\}}\\
0
\end{bmatrix}.
\end{equation}
Substituting these expressions into the formulas above gives
\begin{align}
\mu_{\mathcal S}(x;1)
= &
k_{\set{S} \setminus \{x_i,y_i\}}(x)^\top
(K_{\set{S} \setminus \{x_i,y_i\}}+\sigma_w^2 I)^{-1}
y_{\set{S} \setminus \{x_i,y_i\}}
=
\mu_{\set{S} \setminus \{x_i,y_i\}}(x),\\
\sigma^2_{\mathcal S}(x;1)
= &
k(x,x)-k_{\set{S} \setminus \{x_i,y_i\}}(x)^\top
(K_{\set{S} \setminus \{x_i,y_i\}}+\sigma_w^2 I)^{-1}
k_{\set{S} \setminus \{x_i,y_i\}}(x)+\sigma_w^2
=
\sigma^2_{\set{S} \setminus \{x_i,y_i\}}(x).
\end{align}

To complete the proof for the output space, we examine the weight vector $v(h_i) = A(h_i)^{-1} y(h_i)$. At the limit $h_i = 1$, the target vector is $y(1) = [y_1, \dots, 0, \dots, y_n]^\top$. Using the previously derived block-inverse structure of $A(1)^{-1}$:
\begin{equation}
    v(1) = \begin{bmatrix} (K_{\set{S}\setminus \{x_i,y_i\}} + \sigma_w^2 I)^{-1} & 0 \\ 0^\top & \frac{1}{\sigma_w^2} \end{bmatrix} \begin{bmatrix} y_{\set{S}\setminus \{x_i,y_i\}} \\ 0 \end{bmatrix} = \begin{bmatrix} (K_{\set{S}\setminus \{x_i,y_i\}} + \sigma_w^2 I)^{-1} y_{\set{S}\setminus \{x_i,y_i\}} \\ 0 \end{bmatrix}
\end{equation}
Let $v_{\set{S}\setminus \{x_i,y_i\}}$ be the weights obtained by training a GP on the subset $\set{S}\setminus \{x_i, y_i\} $. It shows that:
\begin{equation}
    v(1) = 
    \begin{cases} 
    (v_{\set{S}\setminus \{x_i,y_i\}})_j & j \neq i \\
    0 & j = i
    \end{cases}
\end{equation}
The predictive mean at $x$ under the removal direction is $\mu_{\set{S}}(x, 1) = k(x, \set{S}, 1)^\top v(1)$. Substituting the zeroed $i$-th components:
\begin{equation}
    \mu_{\set{S}}(x, 1) = \sum_{j \neq i} k(x, x_j) v_j(1) + k(x, x_i) \cdot 0 = \sum_{j \neq i} k(x, x_j) (v_{\set{S}\setminus \{x_i,y_i\}})_j
\end{equation}
This is identically the LOO predictive mean $\mu_{\set{S}\setminus \{x_i,y_i\}}(x)$. Since both the mean and the variance recover the LOO statistics at $h_i=1$ under fixed $\sigma_w^2$, the proposition holds.
\end{proof}

\subsection{Epistemic Influence} \label{sec:appendix_epistemic1}
\propEpistemicInfluence*
\begin{proof}
   Let the predictive uncertainty of the GP for a set of samples \(\set{S}\) be defined as in \Cref{eq:variance}, and we consider the aleatoric noise estimate \(\sigma_w^2(\set{S})\) to be fixed. 
   \begin{equation}
       \sigma_{\set{S}}^2(x) = k(x,x) - k_{x,\set{S}}(K + \sigma_w^2(\set{S})I)^{-1}k_{\set{S},x} + \sigma_w^2(\set{S})I
   \end{equation}
   Furthermore, we consider the RKHS and output removal directions in \(h_i\) introduced in \Cref{prop:removalDirection}:
  \begin{equation}
   \sigma_{\set{S}}^2(x) = k(x,x) - k(x,\set{S},h_i)(K(h_i) + \sigma_w^2(\set{S})I)^{-1}k(\set{S},x,h_i) + \sigma_w^2(\set{S})I,
   \end{equation}
   where we define the Gram matrix under the removal is defined using the  as 
   \begin{equation}\label{eq:GramDirection}
       K(h_i) = k(\set{S},\set{S},h_i) = [(1-h_j)K_{j,k}(1-h_k)]_{j,k} = \begin{cases}
       (1-h_i)k(x_j,x_k) & \text{if } j = i \lor k= i\\
       (1-h_i)^2k(x_j,x_k) & \text{if } i = j = k\\
       k(x_j,x_k) & \text{otherwise.}
       \end{cases}
   \end{equation}
   Hence, after deriving the continuous surrogate, we differentiate w.r.t. \(h_i\). Thereby, we observe that all noise terms \(\sigma_w^2(\set{S})\) drop out, as we assume the aleatoric component to be fixed: 
   \begin{align}
       \frac{\partial  \sigma_{\set{S}}^2(x)}{\partial h_i} =& \frac{\partial}{\partial h_i} k(x,x) - \frac{\partial}{\partial h_i} k(x,\set{S},h_i)(K(h_i) + \sigma_w^2(\set{S})I)^{-1}k(\set{S},x,h_i) + \frac{\partial}{\partial h_i} \sigma_w^2(\set{S}) I\\
       =& - \frac{\partial}{\partial h_i} k(x,\set{S},h_i)(K(h_i) + \sigma_w^2(\set{S})I)^{-1}k(\set{S},x,h_i) \\
       =& -\frac{\partial k(x,\set{S},h_i)}{\partial h_i} (K(h_i) + \sigma_w^2(\set{S})I)^{-1}k(\set{S},x,h_i) \\ &- k(x,\set{S},h_i)\frac{\partial (K(h_i) + \sigma_w^2(\set{S})I)^{-1}}{\partial h_i} k(\set{S},x,h_i) \nonumber \\ &- k(x,\set{S},h_i)(K(h_i) + \sigma_w^2(\set{S})I)^{-1}\frac{\partial k(\set{S},x,h_i)}{\partial h_i}\nonumber\\
       =& - 2 \frac{\partial k(x,\set{S},h_i)}{\partial h_i}(K(h_i) + \sigma_w^2(\set{S})I)^{-1}k(\set{S},x,h_i) \\
       &+ k(x,\set{S},h_i)A(h_i)^{-1}\frac{\partial K(h_i)}{\partial h_i} A(h_i)^{-1} k(\set{S},x,h_i)\nonumber\\
       =& 2k(x,x_i)e_i^\top A(h_i)^{-1}k(\set{S},x,h_i) \\
       & + k(x,\set{S},h_i)A(h_i)^{-1}(e_i k(x_i,\set{S}) + k(\set{S},x_i) e_i^\top) A(h_i)^{-1} k(\set{S},x,h_i).\nonumber 
\end{align}
We can further simplify this expression by differentiating the Gram matrix \(K(h_i)\) of \Cref{eq:GramDirection} and introducing the short notation \(\alpha = A(h_i)^{-1}k(\set{S},x,h_i) \). We hence obtain
\begin{align}
       \frac{\partial  \sigma_{\set{S}}^2(x)}{\partial h_i} =& 2k(x,x_i) \alpha_i + \alpha^\top \left[ - e_i k(x_i, \set{S}, h_i) - k(\set{S}, x_i, h_i) e_i^\top \right] \alpha \\ &= 2k(x,x_i) \alpha_i - \left[ (\alpha^\top e_i) (k(x_i, \set{S}, h) \alpha) + (\alpha^\top k(\set{S}, x_i, h)) (e_i^\top \alpha) \right] \\ &= 2k(x,x_i) \alpha_i - 2 \alpha_i (k(x_i, \set{S}, h) \alpha)
   \end{align}
Since we evaluate the partial derivative at \(h_i = 0\), the term simplifies to 
\begin{align}
    \left. \frac{\partial  \sigma_{\set{S}}^2(x)}{\partial h_i} \right|_{h_i=0} =&
    2k(x,x_i) \alpha_i - 2 \alpha_i (k(x_i, \set{S}) \alpha) \\
    =& 2 k(x,x_i) (A^{-1}k(\set{S},x))_i - 2 (A^{-1}k(\set{S},x))_i (k(x_i,\set{S})A^{-1}k(\set{S},x)),
\end{align}
which completes the proof, when writing the kernel vectors in index notation.
\end{proof}

\corrDerivative*
\begin{proof}
\label{proof:delta-and-derivative}
Let \( \mathcal S=\{x_1,\dots,x_n\}\) be the training set, fix an index \(i\) and a single test point \(x_t\). 
Recall
\begin{equation}
A = K(\mathcal S,\mathcal S) + \sigma_w^2(\set{S}) I,
\qquad k := k_{\mathcal S,x_t} = k(\mathcal S,x_t)\in\mathbb R^n,
\end{equation}
and partition \(A\) and \(k\) conformably isolating index \(i\):
\begin{equation}
A =
\begin{bmatrix}
A_{i,i} & A_{i,-i}\\
A_{-i,i} & A_{-i,-i}
\end{bmatrix},\qquad
k = \begin{bmatrix} k_i \\ k_{-i} \end{bmatrix}.
\end{equation}
Define the Schur complement \cite{quellette1981schur}
\begin{equation}
S_{i,i} := A_{i,i} - A_{i,-i} A_{-i,-i}^{-1} A_{-i,i}.
\end{equation}
\paragraph{Epistemic uncertainty difference for leave one-out:}
Using the block inverse identities in \cite[Prop.~2]{ginsbourger2025fast},
the \(ii\)-block of \(A^{-1}\) equals \(A^{-1}_{i,i}=S_{i,i}^{-1}\) and the off-blocks satisfy
\begin{gather}
A^{-1}_{i,-i} = -S_{i,i}^{-1} A_{i,-i} A_{-i,-i}^{-1},\qquad A^{-1}_{-i,i}=-A_{-i,-i}^{-1} A_{-i,i} S_{i,i}^{-1},\\
A^{-1}_{-i,-i} = A^{-1}_{-i,-i} + A^{-1}_{-i,-i}A_{-i,i}S_{i,i}^{-1}A_{i,-i}A_{-i,-i}^{-1}
\end{gather}
The GP predictive variance with the full set \(\mathcal S\) is
\begin{equation}
\sigma^2_{\mathcal S}(x_t) = k(x_t,x_t) - k^\top A^{-1} k,
\end{equation}
and the variance when training on \(\set{S} \setminus \{x_i,y_i\}\) (i.e. removing index \(i\))
is
\begin{equation}
\sigma^2_{\set{S} \setminus \{x_i,y_i\}}(x_t) = k(x_t,x_t) - k_{-i}^\top A_{-i,-i}^{-1} k_{-i}.
\end{equation}
Therefore the variance drop due to sample \(i\) is
\begin{equation}
\Delta_i(x_t) = \sigma^2_{\set{S} \setminus \{x_i,y_i\}}(x_t) - \sigma^2_{\mathcal S}(x_t)
= k^\top A^{-1} k - k_{-i}^\top A_{-i,-i}^{-1} k_{-i}.
\end{equation}
Expanding \(k^\top A^{-1} k\) using the block-inverse factorization yields the Schur-form:
\begin{align}
    k^\top A^{-1} k =& \begin{bmatrix}
        k_i &k_{-i}^\top
    \end{bmatrix}\begin{bmatrix}
        S^{-1}_{i,i} & A^{-1}_{i,-i}\\
         A^{-1}_{-i,i} & A^{-1}_{-i,-i}
    \end{bmatrix}
    \begin{bmatrix}
        k_i \\k_{-i}
    \end{bmatrix}\\
    =& S^{-1}_{i,i}k_i^2 + 2k_iA_{i,-i}^{-1}k_{-i} + k_{-i}^\top P_{-i,-i}^{-1}k_{-i}\\
    =& S^{-1}_{i,i}k_i^2 + 2k_i(-S_{i,i}^{-1}A_{i,-i}A_{-i,-i}^{-1})k_{-i} \\&+ k_{-i}^\top(A_{-i,-i}^{-1} + A_{-i,-i}^{-1}A_{-i,i}S_{i,i}^{-1}A_{i,-i}A_{-i,-i}^{-1})k_{-i}\nonumber\\
    =& S^{-1}_{i,i}k_i^2 - 2k_iS_{i,i}^{-1}A_{i,-i}A_{-i,-i}^{-1}k_{-i} \\&+ k_{-i}^\top A_{-i,-i}^{-1}k_{-i} +  k_{-i}^\top A_{-i,-i}^{-1}A_{-i,i}S_{i,i}^{-1}A_{i,-i}A_{-i,-i}^{-1}k_{-i}\nonumber\\
    =& k_{-i}^\top A_{-i,-i}^{-1}k_{-i} + S^{-1}_{i,i}(k_i^2 - 2k_iA_{i,-i}A_{-i,-i}^{-1}k_{-i} \\&+  (A_{i,-i}A_{-i,-i}^{-1}k_{-i})^\top(A_{i,-i}A_{-i,-i}^{-1}k_{-i}))\nonumber\\
    =& k_{-i}^\top A_{-i,-i}^{-1}k_{-i} + S^{-1}_{i,i}(k_i - A_{i,-i}A_{-i,-i}^{-1}k_{-i})^2.\label{eq:kA-1k}
\end{align}
Further, the \(i\)-th component of \(A^{-1} k\) is given by
\begin{equation}\label{eq:Pk_i}
(A^{-1}k)_i = A^{-1}_{ii}k_i + A^{-1}_{i,-i}\,k_{-i} = S_{i,i}^{-1}\big(k_i - A_{i,-i} A_{-i,-i}^{-1} k_{-i}\big).
\end{equation}
Finally, we express the variance drop using \Cref{eq:kA-1k}:
\begin{equation}
\Delta_i(x_t)
= S_{i,i}^{-1}\big(k_i - A_{i,-i} A_{-i,-i}^{-1} k_{-i}\big)^2
\labelrel{=}{sup_eq_Pk_i} S_{i,i}^{-1}\big(S_{i,i} (A^{-1} k)_i\big)^2
\labelrel{=}{use_S_inv_A} \frac{(A^{-1} k_{\set{S},x_t})_i^2}{A^{-1}_{i,i}}.
\end{equation}
We obtain step \ref{sup_eq_Pk_i} by substituting \Cref{eq:Pk_i} and \ref{use_S_inv_A} from \(S^{-1}_{i,i} = A^{-1}_{i,i}\).

\paragraph{Directional-derivative:}
Let us abbreviate $\alpha = A^{-1}k$ and $v = A^{-1}_{ii} = \mathbf{e}_i^\top A^{-1} \mathbf{e}_i$. We compute the exact directional derivative of $\Delta_i = \alpha_i^2 / v$ with respect to a change in the $i$-th kernel direction. By following the quotient rule:
\begin{equation}
    \partial_{k_i} \Delta_i = \frac{2\alpha_i}{v} \partial_{k_i}\alpha_i - \frac{\alpha_i^2}{v^2} \partial_{k_i}v.
\end{equation}
Using the identity $\partial_{k_i}(A^{-1}) = -A^{-1} (\partial_{k_i}A) A^{-1}$, the differentials for $\alpha_i = \mathbf{e}_i^\top A^{-1} k$ and $v$ are:
\begin{align}
    \partial_{k_i}\alpha_i &= \mathbf{e}_i^\top (-A^{-1} (\partial_{k_i}A) A^{-1} k + A^{-1} \partial_{k_i}k) = - \mathbf{e}_i^\top A^{-1} (\partial_{k_i}A) \alpha + v \partial_{k_i}k_, \\
    \partial_{k_i}v &= \mathbf{e}_i^\top (-A^{-1} (\partial_{k_i}A) A^{-1}) \mathbf{e}_i = - v \mathbf{e}_i^\top (\partial_{k_i}A) A^{-1}_{:,i} - (A^{-1}_{i,:} \partial_{k_i}A) \mathbf{e}_i v.
\end{align}
For a directional change along the $i$-th component, the variations are $k_i\partial_{k_i}k = \mathbf{e}_i k_i$ and $k_i\partial_{k_i}A = \mathbf{e}_i K_{i, \mathcal{S}} + K_{\mathcal{S}, i} \mathbf{e}_i^\top$. Substituting these into the differentials:
\begin{align}
    \partial_{k_i}\alpha_i &= - (v (K_{i, \mathcal{S}} \alpha) + (A^{-1}_{i, \mathcal{S}} K_{\mathcal{S}, i}) \alpha_i) + v k_i, \\
    \partial_{k_i}v &= - (v (K_{i, \mathcal{S}} A^{-1}_{:,i}) + (A^{-1}_{i, \mathcal{S}} K_{\mathcal{S}, i}) v) = -2v (K_{i, \mathcal{S}} A^{-1}_{:,i}).
\end{align}
Now, substitute $\partial_{k_i} \alpha_i$ and $\partial_{k_i} v$ back into the total differential $d\Delta_i$:
\begin{align}
    \partial_{k_i}\Delta_i &= \frac{2\alpha_i}{v} \left( -v(K_{i, \mathcal{S}}\alpha) - \alpha_i(K_{i, \mathcal{S}}A^{-1}_{:,i}) + v k_i \right) - \frac{\alpha_i^2}{v^2} \left( -2v(K_{i, \mathcal{S}}A^{-1}_{:,i}) \right) \\
    &= -2\alpha_i (K_{i, \mathcal{S}}\alpha) - \frac{2\alpha_i^2}{v} (K_{i, \mathcal{S}}A^{-1}_{:,i}) + 2\alpha_i k_i + \frac{2\alpha_i^2}{v} (K_{i, \mathcal{S}}A^{-1}_{:,i}) \\
    &= 2\alpha_i k_i - 2\alpha_i (K_{i, \mathcal{S}}\alpha).
\end{align}
Hence, we recover the epistemic influence $u_i(x)$:
\begin{equation}
    u_i(x) = 2k_{x,x_i}(A^{-1}k)_i - 2(A^{-1}k)_i(K_{i,\mathcal{S}} A^{-1}k).
\end{equation}
\end{proof}

\subsection{Aleatoric Influence}\label{sec:appendix_aleatoric}
\propAleatoricDerivative*
\begin{proof}
We abbreviate \(A=A(\sigma_w^2(\set{S}))\) and \(\sigma_w^2 = \sigma_w^2(\set{S})\) for shorter notation. By definition of the maximal log marginal likelihood the pair \((\sigma_w^2(\set{S}),y)\) satisfies
\begin{align}
\mathcal{D}(\sigma_w^2,\set{S},a,b) &= \frac{\partial (\log p(y|\set{S},\sigma_w^2) + \log p(\sigma_w^2(\set{S})))}{\partial \sigma_w^2} = 0 \\
 &= \tfrac{1}{2}y^\top A^{-1} \frac{\partial A^{-1}}{\sigma_w^2}A^{-1} y - \frac{1}{2}\tr\left(A^{-1}\frac{\partial A}{\partial \sigma_w^2} \right) + \frac{\partial p(\sigma_w^2)}{\partial \sigma_w^2}= 0\\
 &=\frac{1}{2}(y^\top A^{-2}y -\tr(A^{-1})) + \frac{a-1}{\sigma_w^2} - b = 0.
\end{align}
We apply the implicit function theorem \cite[Theorem~1.3.1]{krantz2002implicit}: differentiate \(\mathcal{D}(\sigma_w^2,\set{S},a,b)=0\) with respect to \(y_i\) while treating \(\sigma_w^2=\sigma_w^2(y)\) as an implicit function of \(y\). We obtain
\begin{equation}
\frac{\partial \mathcal{D}(\sigma_w^2,\set{S},a,b)}{\partial \sigma_w^2}\frac{\partial \sigma_w^2}{\partial h_i} + \frac{\partial \mathcal{D}(\sigma_w^2,\set{S},a,b)}{\partial h_i} = 0
\qquad\Longrightarrow\qquad
\frac{\partial \sigma_w^2}{\partial h_i}
= -\frac{\frac{\partial \mathcal{D}(\sigma_w^2,\set{S},a,b)}{\partial h_i}}{\frac{\partial \mathcal{D}(\sigma_w^2,\set{S},a,b)}{\partial \sigma_w^2}}.
\end{equation}

Since \(\mathcal{D}(\sigma_w^2,\set{S},a,b)=y^\top(h_i) A(h_i)^{-2} y(h_i) - \tr(A(h_i)^{-1}) + \frac{a-1}{\sigma_w^2} - b\), we get for the partial derivative of \(\mathcal{D}(\sigma_w^2,\set{S},a,b)\) with respect to \(h_i\), that the enumerator is given by
\begin{align}
\frac{\partial \mathcal{D}(\sigma_w^2,\set{S},a,b)}{\partial h_i}
= & \frac{\partial}{\partial h_i}((y^\top(h_i) A(h_i)^{-2} y(h_i)) - \tr(A(h_i)^{-1}))\\
= & y^\top(h_i) \frac{\partial A(h_i)^{-2}}{\partial h_i} y(h_i) - \frac{\partial}{\partial h_i} \tr(A(h_i)^{-1}) + 2 \frac{\partial y(h_i)}{h_i}A(h_i)^{-2}\\
= & 2((A^{-1}y)_i(k_{x_i,\set{S}}(A^{-2}y)) + (A^{-2}y)_i(k_{x_i,\set{S}} (A^{-1}y))) \\ &- 2k_{x_i,\set{S}}(A^{-2})_{\cdot, i} - 2y_i(A^{-2}y)_i.\nonumber
\end{align}

Hence, we can use the results of  to compute the derivative of \(\mathcal{D}(\sigma_w^2,\set{S},a,b)\) with respect to \(\sigma_w^2\) as:
\begin{align}
\frac{\partial \mathcal{D}(\sigma_w^2,\set{S},a,b)}{\partial \sigma_w^2}
&= \frac{\partial}{\partial\sigma_w^2}\left((y^\top A(\sigma_w^2)^{-2} y) - \tr(A(\sigma_w^2)^{-1}) + \frac{2(a-1)}{\sigma_w^2}\right) \\
&= y^\top\big(-2A^{-3}\big)y - \tr\left(-A^{-2}\right) + \frac{-2(a-1)}{(\sigma_w^2)^2}\\
&= -2y^\top A^{-3} y + \tr(A^{-2}) - \frac{2(a-1)}{(\sigma_w^2)^2}
\end{align}

Finally, we combine the derived terms in the implicit differentiation, evaluated at \(h_i = 0\) 
\begin{align}
\frac{\partial \sigma_w^2}{\partial h_i} &= -\frac{ 2((A^{-1}y)_i(k_{x_i,\set{S}}(A^{-2}y)) + (A^{-2}y)_i(k_{x_i,\set{S}} (A^{-1}y))) - 2k_{x_i,\set{S}}(A^{-2})_{\cdot, i} - 2y_i(A^{-2}y)_i}{-2y^\top A^{-3} y + \tr(A^{-2}) - \frac{2(a-1)}{(\sigma_w^2)^2}}\\ 
&= \frac{- (A^{-1}y)_i(k_{x_i,\set{S}}(A^{-2}y)) - (A^{-2}y)_i(k_{x_i,\set{S}} (A^{-1}y)) + k_{x_i,\set{S}}(A^{-2})_{\cdot, i} + y_i(A^{-2}y)_i}{-y^\top A^{-3} y + \tfrac12\tr(A^{-2}) - \frac{a-1}{(\sigma_w^2)^2}}.
\end{align}
This completes the proof for the derivative of the aleatoric MAP w.r.t. the removal direction \(h_i\).
\end{proof}

\propAleatoricInfluence*
\begin{proof}
    Let the predictive uncertainty of the GP for a set of samples \(\set{S}\) be defined as in \Cref{eq:variance}, then we compute the derivative with respect to all aleatoric noise terms of \(\sigma_{\set{S}}^2\), assuming that the kernel contributions and Gram matrix are fixed
    \begin{equation}
        \sigma_{\set{S}}^2(x) = k(x,x) - k_{x,\set{S}}(K + \sigma_w^2(\set{S})I)^{-1}k_{\set{S},x} + \sigma_w^2(\set{S})I.
    \end{equation}
    We further consider the RKHS and output space direction \(h_i\) and differentiate all aleatoric terms, assuming that the kernel and Gram matrix are fixed. 
    \begin{align}
        \frac{\partial \sigma_{\set{S}}^2(x)}{\partial h_i} & = \frac{\partial}{\partial h_i} k_{x,\set{S}}(K + \sigma_w^2(\set{S})I)^{-1}k_{\set{S},x} + \frac{\partial}{\partial h_i} \sigma_w^2(\set{S})\\
        & = k_{x,\set{S}} A^{-1} \frac{\partial \sigma_w^2(\set{S})I}{\partial h_i} A^{-1} k_{\set{S},x} + \frac{\partial \sigma_w^2(\set{S})}{\partial h_i}\\
        & = (1 + k_{x,\set{S}} A^{-2}k_{\set{S},x}) \frac{\partial \sigma_w^2(\set{S})}{\partial h_i},
    \end{align}
    which completes the proof for the aleatoric influence measure \(v_i(x)\).
\end{proof}

\section{Forgetful Bayesian Optimization}\label{sec:appendix_bo}

\begin{algorithm}[h]
\caption{\algname{Forgetful BO}}
\label{algo:forgetful_bo}
\begin{algorithmic}
\Require Objective $f$, domain $\set{X}$, acquisition function $\alpha$, initial dataset size $n_0$, budget $T$

\State Initialize dataset $\set{S}_0 = \{(x_i, y_i)\}_{i=1}^{n_0}$ with $y_i = f(x_i)$
\State Fit initial GP model $(\mu_{\set{S}_0}(\cdot),\sigma_{\set{S}_0}(\cdot))$ on $\set{S}_0$
\State Set $x^* = \arg\max_{x_i \in \set{S}_0} y_i$, $y^* = \max_{x_i \in \set{S}_0} y_i$

\For{$t = 1, \dots, T$}
\State Optimize acquisition function: $x_t = \arg\max_{x \in \set{X}} \alpha(x |(\mu_{\textcolor{blue}{\set{S}'_{t-1}}}(\cdot),\sigma_{\textcolor{blue}{\set{S}'_{t-1}}}(\cdot)))$
    \State Evaluate objective: $y_t = f(x_t)$ and update $\mathcal{S}_t = \mathcal{S}_{t-1} \cup \{(x_t, y_t)\}$
    \State \((x^*,y^*) \gets \max((x_t,y_t),(x^*,y^*)\)
    \State Initialize \(\tau=0\), \(\set{S}_t^{(0)} = \set{S}_t\) and fit noise MAP II
    \State \textcolor{blue}{Determine \(\set{A}_t^{(0)}\) and compute \(u^{(0)}(\set{A}_t^{(0)})\), \(v^{(0)}(\set{A}_t^{(0)})\) \Comment{\Cref{eq:regionOfInterestBORec,prop:epistemic_influence,prop:aleatoric_influence}}
    \While{\(\min(u^{(\tau)}(\set{A}_t^{(\tau)})+v^{(\tau)}(\set{A}_t^{(\tau)})) < 0\)}
        \State Determine worst sample: \(i = \arg\min_{j} (u_j^{(\tau)}(\set{A}_t^{(\tau)})+v_j^{(\tau)}(\set{A}_t^{(\tau)}))\)
        \State Update \(\set{S}_t^{(\tau+1)} = \set{S}_t^{(\tau)} \setminus x_i\) and refit GP noise MAP II
        \State Update \(\tau \gets \tau+ 1\),  \(\set{A}_t^{(\tau)}\), \(u^{(\tau)}(\set{A}_t^{(\tau)})\), \(v^{(\tau)}(\set{A}_t^{(\tau)})\) \Comment{\Cref{eq:regionOfInterestBORec,prop:epistemic_influence,prop:aleatoric_influence}}
    \EndWhile
    \State Set \(\set{S}_t' = \set{S}_t^{(\tau)}\)}
\EndFor\\
\Return $x^*$, $y^*$
\end{algorithmic}
\end{algorithm}

Starting from an initial sample set \(\set{S}_0\) of size \(n_0\), a Gaussian process model is fit to \(\set{S}_0\) and iteratively updated as new evaluations are acquired by maximizing an acquisition function \(\alpha\). Before each function evaluation, except for the first, the algorithm identifies a region of interest \(\set{A}\) using \Cref{eq:regionOfInterestBORec} and assesses the contribution of each data point in \(\set{S}\) to the predictive uncertainty within this region \(\set{A}\). Samples with negative influence are removed iteratively as in \Cref{algo:joint}, and the GP noise MLE estimate is refit on the reduced dataset \(\set{S}'\) while recursively updating the region of interest according to \Cref{eq:regionOfInterestBORec}. This forgetting step continues until all remaining samples provide a positive contribution to uncertainty reduction in the target region \(\set{A}\). Using this improved model, we next maximize the acquisition function \(\alpha\) and determine the next probing location \(x_t\), which is evaluated and added to the overall data set \(\set{S}\). From this data-set, we start the next iteration by fitting the GP with the updated \(\set{S}\) and repeat this process until reaching the evaluation budget \(T\).

\begin{figure}[h]
    \centering
    \includegraphics[width=1\linewidth]{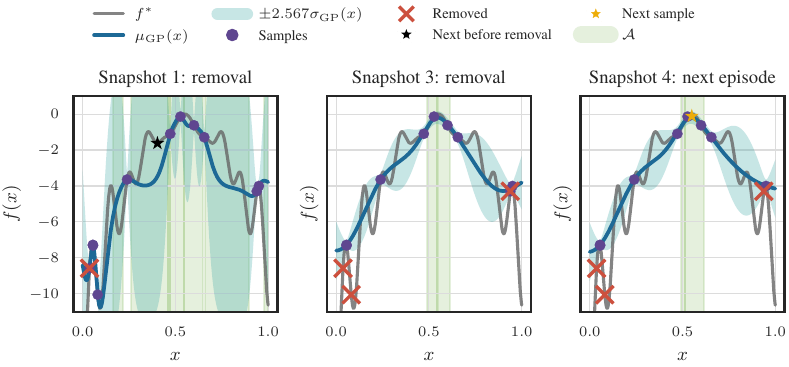}
    \caption{\algname{Forgetful BO}: Visualization of sequential removal mechanism integrated in BO pipeline.}
    \label{fig:removal_visualization}
\end{figure}

In \Cref{fig:removal_visualization}, we visualize the application of our iterative removal scheme \algname{F2I} in our \algname{Forgetful BO} pipeline. In \emph{Snapshot 1}, we visualize the default predictive distribution based on all samples and indicate with the black star the next chosen query point according to the GP-UCB acquisition function. The first red cross in \emph{Snapshot 1} indicates that \algname{Forgetful BO} is removing the right-most sample. As we can see from \emph{Snapshot 3}, this removal causes the uncertainty bands to inflate at the location of the removal, which are outside of the previous target area \(\set{A}\), for which we have optimized the samples. Now it is important to consider the intersection rule of the target region update in \Cref{eq:regionOfInterestBORec}, to not include previously excluded regions in the target region. Further, we see in \emph{Snapshot 3} that the detrimental data structure, composed of the three data points in the bottom left corner, is removed, resulting in the improved posterior predictive distribution visualized in \emph{Snapshot 4}. The recursive target region in \Cref{eq:regionOfInterestBORec} has substantially decreased the size of the target region further. Within the target region, we now apply the GP-UCB acquisition function to query the next point at the location indicated with the yellow star, which coincides with the global optimum of our test function. This demonstrates visually how our removal scheme in \algname{Forgetful BO} improves the predictive distribution of the model, and thereby results in better acquisition. 
\section{Forgetful Reinforcement Learning}\label{sec:appendix_rl}

\begin{algorithm}[htbp]
\caption{\algname{Forgetful RL}}
\label{algo:forgetful_rl}
\begin{algorithmic}
\Require Initial real experience $\mathcal{S}_0 = \{(x_i,a_i,x_{i+1},r_i)\}_{i=1}^{n_0}$, 
GP prior, SAC buffer $\mathcal{D}_{\mathrm{SAC}} \leftarrow \{\}$
\State Fit initial GP model $(\mu_{\mathcal{S}_0}(\cdot),\sigma_{\mathcal{S}_0}(\cdot))$ on $\mathcal{S}_0$ (type II MAP)
\State Initialize policy $\pi_0$, value $Q^{\pi_0}(x,a)$
\For{$n = 1, \dots, N$}
    \State Fit GP on current dataset $\mathcal{S}_{n-1}$
    \textcolor{blue}{
    \State Compute target region $\mathcal{A}_n$ and $u(\mathcal{A}_n), v(\mathcal{A}_n)$ \Comment{\Cref{eq:regionOfInterestBORec,prop:epistemic_influence,prop:aleatoric_influence}}
    \If{there exist samples $j\in\mathcal{A}_n$ with $(u_j+v_j) < 0$}
        \State $R \gets \{ j \in \mathcal{S}_{n-1} \mid (u_j + v_j) < 0 \}$ \Comment{indices of negatively influential samples}
        \State Remove samples: $\mathcal{S}_{n-1} \gets \mathcal{S}_{n-1} \setminus \{(x_j,a_j,x_{j+1},r_j)\}_{j\in R}$
        \State Refit GP on the pruned dataset $\mathcal{S}_{n-1}$ to obtain $(\mu_{\mathcal{S}_{n-1}}(\cdot),\sigma_{\mathcal{S}_{n-1}}(\cdot))$
    \EndIf
    \State 
    }
    
    \For{$m=1,\dots,M$} \Comment{Generate $M$ synthetic rollouts of length $H$}
        \State Sample start state $\tilde{x}_0 \sim \rho_0$
        \For{$t=0,\dots,H-1$}
            \State Sample action $\tilde{a}_t \sim \pi_{n-1}(\cdot\mid \tilde{x}_t)$
            \State Predict via GP: $\tilde{x}_{t+1} \sim p_{\mathrm{GP}}(\cdot\mid \tilde{x}_t,\tilde{a}_t)$, \; $\tilde{r}_t \leftarrow r_{\mathrm{GP}}(\tilde{x}_t,\tilde{a}_t)$
            \State Store synthetic transition $(\tilde{x}_t,\tilde{a}_t,\tilde{x}_{t+1},\tilde{r}_t)$ in $\mathcal{D}_{\mathrm{SAC}}$
        \EndFor
    \EndFor
    \State 
    \State Learn \(\pi_n = \arg \max_{\pi \in \Pi} \mathbb{E}_{\pi,p_{\mathrm{GP}}} \left[\sum_{h=0}^{H-1} r(s_t,a_t)\right]\) with SAC on \(\set{D}_{\mathrm{SAC}}\).
    \State 
    \For{$t=0,\dots,H-1$}
        \State Execute action $a_t \sim \pi_{n}(s_t)$ in the environment
        \State Append collected transition to the real buffer: $\mathcal{S}_{n} \gets \mathcal{S}_{n-1} \cup \{s_t,a_t,s_{t+1},r_t\}$
    \EndFor
\EndFor
\end{algorithmic}
\end{algorithm}

Our MBRL implementation builds upon Model-Based Policy Optimization (MBPO) \cite{janner2019trust} and employs GPs to model both the dynamics and reward function. In each episode \(n\), the model is trained on the set of real-word transitions stored in the replay buffer \(\set{S}_n\). The learned model is then used to simulate trajectories for improving the current policy \(\pi_n\). Policy optimization is performed using a Soft Actor-Critic (SAC) \cite{haarnoja2018soft} agent with replay buffer \(\set{D}_{SAC}\) and corresponding state-action value function \(Q^{\pi_n}(x,a)\). We in practice approximate the upper and lower confidence bound values in \Cref{eq:rl_region_of_interest} using the respective highest and lowest ensemble member of value functions. For a more principled approach to obtain these optimistic and pessimistic Q-functions, one can implement an uncertainty Bellman equation \cite{o2018uncertainty} or an approximation by for example using an optimistic and pessimistic uncertainty bonus, based on the model uncertainty \cite{wendl2026safe}, or sampling based \cite{vignola2026sampling}.

To enhance model reliability, we introduce an additional removal step prior to generating the synthetic rollouts. Specifically, we determine the target region for potential improvements in \Cref{eq:rl_region_of_interest}. For this target region, we next determine the influence measures \(u\) and \(v\) for all samples in \(\set{S}_n\) and remove all samples with negative total influence. After pruning these samples, we refit the GP by recomputing its MAP estimate for the reduced dataset. This procedure aims to reduce predictive uncertainty and mitigate the effect of detrimental data points before model-based rollouts are performed.

Finally, this model is used to roll out the current policy \(\pi_n\), generating synthetic transitions that are stored in the buffer \(\mathcal{D}_{SAC}\). The current policy is then further optimized using both real and synthetic experiences through the learned state-action value function \(Q^{\pi_n}(x,a)\).

\section{Limitations and Future Work}\label{sec:limitations}

Although the proposed removal procedure is substantially more efficient than exhaustive search or extensive leave-one-out evaluations, it still incurs computational overhead compared to a standard homoscedastic GP because each greedy step requires updating the model and evaluating the removal influence under the current posterior. Compared to other robust Gaussian process Bayesian Optimization baselines, we have comparable or lower computational costs \Cref{fig:runtimes}. 

The accompanying near-optimality guarantee is based on a restricted curvature condition that serves as a theoretical certificate rather than an algorithmic constraint; in practice, these conditions may be conservative, and the method can remain effective even when they are not verified explicitly. Finally, the current formulation is tailored to Gaussian process models, and it would be interesting to investigate analogous removal influences for parametric models.
\section{Experimental Setup}
In this section, we provide a detailed description of the experimental setup of the transductive and Bayesian optimization benchmarks in \Cref{sec:appendix_experiments_transductive,sec:appendix_experiments_bo}. For transductive learning, we additionally provide the numeric results table of \Cref{fig:transductive_results}.

\subsection{Transductive Learning}
\label{sec:appendix_experiments_transductive}
We provide in \Cref{tab:transuctive_results}, the absolute values with standard deviations for the results in \Cref{fig:transductive_results}.

\begin{table*}[t]
\centering
\tiny
\setlength{\tabcolsep}{4pt}
\begin{tabular}{c l c c c c c c}
\toprule
 &  & \textbf{Homoscedastic GP} & \textbf{Target Sampling} & \textbf{Relevance Pursuit} & \textbf{Warped GP} & \textbf{Heteroscedastic GP} & \textbf{F2I} \\
\midrule

\multirow{6}{*}{\rotatebox{90}{\textbf{Dow Jones}}}
 & MSE
 & \(0.089 \pm 0.093\)
 & \(0.100 \pm 0.021\)
 & \(0.074 \pm 0.060\)
 & \(0.165 \pm 0.259\)
 & \(0.352 \pm 0.220\)
 & \(\mathbf{0.064} \pm 0.051\) \\

 & MAE
 & \(0.209 \pm 0.089\)
 & \(0.262 \pm 0.022\)
 & \(0.207 \pm 0.071\)
 & \(0.295 \pm 0.185\)
 & \(0.499 \pm 0.184\)
 & \(\mathbf{0.191} \pm 0.065\) \\

 & NLL
 & \(0.188 \pm 0.371\)
 & \(0.455 \pm 0.378\)
 & \(0.118 \pm 0.412\)
 & \(0.578 \pm 0.553\)
 & \(0.946 \pm 0.510\)
 & \(\mathbf{-0.006} \pm 0.244\) \\

 & CE
 & \(0.040 \pm 0.038\)
 & \(0.019 \pm 0.018\)
 & \(\mathbf{0.010} \pm 0.021\)
 & \(0.045 \pm 0.039\)
 & \(0.028 \pm 0.034\)
 & \(0.011 \pm 0.014\) \\

 & ECE
 & \(0.027 \pm 0.022\)
 & \(0.011 \pm 0.003\)
 & \(\mathbf{0.010} \pm 0.004\)
 & \(0.030 \pm 0.012\)
 & \(0.023 \pm 0.010\)
 & \(\mathbf{0.010} \pm 0.004\) \\

 & Sharpness
 & \(0.455 \pm 0.288\)
 & \(0.263 \pm 0.067\)
 & \(\mathbf{0.227} \pm 0.037\)
 & \(0.537 \pm 0.123\)
 & \(0.553 \pm 0.243\)
 & \(0.237 \pm 0.061\) \\

\midrule
\multirow{6}{*}{\rotatebox{90}{\textbf{Yacht}}}
 & MSE
 & \(0.003 \pm 0.002\)
 & \(0.018 \pm 0.089\)
 & \(0.004 \pm 0.005\)
 & \(\mathbf{0.002} \pm 0.002\)
 & \(\mathbf{0.002} \pm 0.005\)
 & \(0.003 \pm 0.002\) \\

 & MAE
 & \(0.040 \pm 0.017\)
 & \(0.038 \pm 0.130\)
 & \(0.041 \pm 0.027\)
 & \(\mathbf{0.029} \pm 0.012\)
 & \(0.033 \pm 0.027\)
 & \(0.038 \pm 0.016\) \\

 & NLL
 & \(-1.138 \pm 0.314\)
 & \(-0.729 \pm 0.699\)
 & \(-1.152 \pm 0.384\)
 & \(0.958 \pm 0.008\)
 & \(-0.591 \pm 0.601\)
 & \(\mathbf{-1.189} \pm 0.296\) \\

 & CE
 & \(0.116 \pm 0.044\)
 & \(0.233 \pm 0.015\)
 & \(\mathbf{0.100} \pm 0.060\)
 & \(0.251 \pm 0.006\)
 & \(0.193 \pm 0.052\)
 & \(0.118 \pm 0.045\) \\

 & ECE
 & \(0.011 \pm 0.003\)
 & \(0.188 \pm 0.845\)
 & \(\mathbf{0.009} \pm 0.004\)
 & \(0.101 \pm 0.001\)
 & \(0.023 \pm 0.013\)
 & \(0.010 \pm 0.003\) \\

 & Sharpness
 & \(0.150 \pm 0.039\)
 & \(0.364 \pm 0.947\)
 & \(\mathbf{0.131} \pm 0.050\)
 & \(1.039 \pm 0.008\)
 & \(0.266 \pm 0.142\)
 & \(0.143 \pm 0.034\) \\

\midrule
\multirow{6}{*}{\rotatebox{90}{\textbf{CASP}}}
 & MSE
 & \(0.340 \pm 0.467\)
 & \(0.893 \pm 0.245\)
 & \(0.543 \pm 0.413\)
 & \(0.679 \pm 0.367\)
 & \(0.700 \pm 0.289\)
 & \(\mathbf{0.318} \pm 0.478\) \\

 & MAE
 & \(0.434 \pm 0.389\)
 & \(0.912 \pm 0.245\)
 & \(0.675 \pm 0.295\)
 & \(0.794 \pm 0.220\)
 & \(0.819 \pm 0.170\)
 & \(\mathbf{0.408} \pm 0.389\) \\

 & NLL
 & \(0.504 \pm 0.922\)
 & \(2.225 \pm 1.180\)
 & \(1.068 \pm 0.509\)
 & \(1.273 \pm 0.167\)
 & \(1.296 \pm 0.346\)
 & \(\mathbf{0.458} \pm 0.945\) \\

 & CE
 & \(0.136 \pm 0.062\)
 & \(0.174 \pm 0.008\)
 & \(0.116 \pm 0.050\)
 & \(\mathbf{0.087} \pm 0.023\)
 & \(0.136 \pm 0.046\)
 & \(0.143 \pm 0.062\) \\

 & ECE
 & \(0.022 \pm 0.016\)
 & \(3.739 \pm 1.025\)
 & \(0.019 \pm 0.015\)
 & \(0.025 \pm 0.015\)
 & \(\mathbf{0.014} \pm 0.012\)
 & \(0.024 \pm 0.015\) \\

 & Sharpness
 & \(0.590 \pm 0.270\)
 & \(4.653 \pm 1.266\)
 & \(0.740 \pm 0.125\)
 & \(1.043 \pm 0.030\)
 & \(0.692 \pm 0.098\)
 & \(\mathbf{0.561} \pm 0.261\) \\

\midrule
\multirow{6}{*}{\rotatebox{90}{\textbf{Asymmetric}}}
 & MSE
 & \(0.239 \pm 0.176\)
 & \(0.251 \pm 0.223\)
 & \(0.275 \pm 0.193\)
 & \(0.275 \pm 0.201\)
 & \(0.474 \pm 0.245\)
 & \(\mathbf{0.230} \pm 0.169\) \\

 & MAE
 & \(0.332 \pm 0.175\)
 & \(\mathbf{0.305} \pm 0.170\)
 & \(0.316 \pm 0.156\)
 & \(0.390 \pm 0.157\)
 & \(0.561 \pm 0.169\)
 & \(0.323 \pm 0.169\) \\

 & NLL
 & \(0.171 \pm 0.886\)
 & \(1.042 \pm 3.261\)
 & \(8.546 \pm 23.641\)
 & \(0.789 \pm 0.412\)
 & \(1.017 \pm 0.533\)
 & \(\mathbf{0.143} \pm 0.877\) \\

 & CE
 & \(\mathbf{0.023} \pm 0.024\)
 & \(0.038 \pm 0.031\)
 & \(0.032 \pm 0.031\)
 & \(0.033 \pm 0.028\)
 & \(0.027 \pm 0.023\)
 & \(0.024 \pm 0.026\) \\

 & ECE
 & \(0.016 \pm 0.007\)
 & \(0.022 \pm 0.010\)
 & \(0.019 \pm 0.010\)
 & \(0.032 \pm 0.033\)
 & \(0.031 \pm 0.031\)
 & \(\mathbf{0.015} \pm 0.007\) \\

 & Sharpness
 & \(0.405 \pm 0.157\)
 & \(0.363 \pm 0.154\)
 & \(\mathbf{0.262} \pm 0.183\)
 & \(0.544 \pm 0.045\)
 & \(0.681 \pm 0.360\)
 & \(0.394 \pm 0.148\) \\

\midrule
\multirow{6}{*}{\rotatebox{90}{\textbf{Symmetric}}}
 & MSE
 & \(0.241 \pm 0.119\)
 & \(0.335 \pm 0.273\)
 & \(0.268 \pm 0.129\)
 & \(0.366 \pm 0.166\)
 & \(0.502 \pm 0.062\)
 & \(\mathbf{0.236} \pm 0.129\) \\

 & MAE
 & \(0.351 \pm 0.129\)
 & \(0.379 \pm 0.202\)
 & \(0.379 \pm 0.140\)
 & \(0.481 \pm 0.155\)
 & \(0.613 \pm 0.033\)
 & \(\mathbf{0.341} \pm 0.137\) \\

 & NLL
 & \(0.273 \pm 0.612\)
 & \(1.543 \pm 4.810\)
 & \(0.348 \pm 0.526\)
 & \(1.330 \pm 0.802\)
 & \(1.218 \pm 0.443\)
 & \(\mathbf{0.141} \pm 0.619\) \\

 & CE
 & \(0.023 \pm 0.027\)
 & \(0.041 \pm 0.041\)
 & \(0.019 \pm 0.031\)
 & \(0.051 \pm 0.040\)
 & \(0.022 \pm 0.018\)
 & \(\mathbf{0.017} \pm 0.026\) \\

 & ECE
 & \(0.024 \pm 0.012\)
 & \(0.025 \pm 0.013\)
 & \(\mathbf{0.016} \pm 0.006\)
 & \(0.049 \pm 0.076\)
 & \(0.025 \pm 0.006\)
 & \(\mathbf{0.016} \pm 0.007\) \\

 & Sharpness
 & \(0.568 \pm 0.201\)
 & \(\mathbf{0.382} \pm 0.135\)
 & \(0.487 \pm 0.125\)
 & \(0.450 \pm 0.122\)
 & \(0.616 \pm 0.124\)
 & \(0.453 \pm 0.135\) \\

\midrule
\multirow{6}{*}{\rotatebox{90}{\textbf{Boston Housing}}}
 & MSE
 & \(0.060 \pm 0.048\)
 & \(0.072 \pm 0.029\)
 & \(0.061 \pm 0.049\)
 & \(0.054 \pm 0.023\)
 & \(\mathbf{0.052} \pm 0.025\)
 & \(0.060 \pm 0.049\) \\

 & MAE
 & \(0.179 \pm 0.064\)
 & \(0.219 \pm 0.039\)
 & \(0.181 \pm 0.066\)
 & \(0.185 \pm 0.042\)
 & \(\mathbf{0.178} \pm 0.045\)
 & \(\mathbf{0.178} \pm 0.064\) \\

 & NLL
 & \(\mathbf{-0.076} \pm 0.521\)
 & \(1.073 \pm 1.154\)
 & \(-0.029 \pm 0.426\)
 & \(0.932 \pm 0.207\)
 & \(0.140 \pm 0.396\)
 & \(-0.066 \pm 0.532\) \\

 & CE
 & \(0.019 \pm 0.016\)
 & \(0.102 \pm 0.114\)
 & \(0.020 \pm 0.019\)
 & \(0.147 \pm 0.034\)
 & \(0.031 \pm 0.024\)
 & \(\mathbf{0.018} \pm 0.016\) \\

 & ECE
 & \(\mathbf{0.004} \pm 0.002\)
 & \(1.725 \pm 2.295\)
 & \(0.005 \pm 0.003\)
 & \(0.027 \pm 0.005\)
 & \(0.007 \pm 0.003\)
 & \(\mathbf{0.004} \pm 0.002\) \\

 & Sharpness
 & \(0.226 \pm 0.059\)
 & \(1.940 \pm 2.295\)
 & \(0.257 \pm 0.089\)
 & \(0.999 \pm 0.142\)
 & \(0.328 \pm 0.093\)
 & \(\mathbf{0.214} \pm 0.056\) \\

\bottomrule
\end{tabular}
\caption{Benchmark comparison showing mean and standard deviation across 100 different training sample draws. Best (lowest mean) values per row are highlighted in bold. Sharpness measures the average prediction uncertainty (lower is better, indicating sharper/more confident predictions).}
\label{tab:transuctive_results}
\end{table*}

\begin{figure}
    \centering
    \includegraphics[width=.95\linewidth]{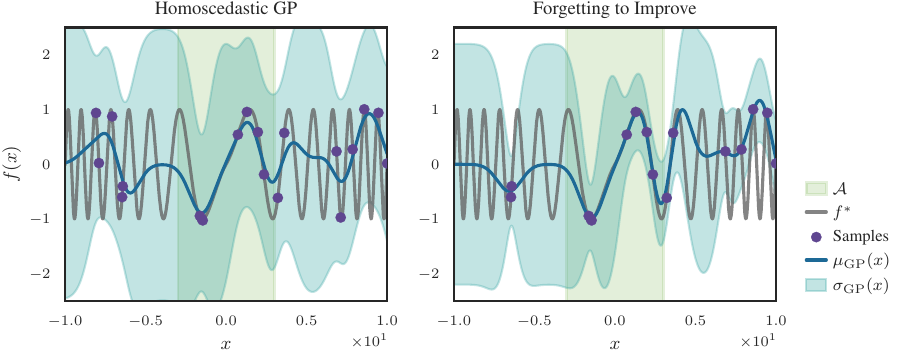}
    \caption{Influence of detrimental data structures based on \emph{Symmetric LS} test functions. (Left) A standard homoscedastic GP absorbs the detrimental lengthscale structures in the noise. (Right) We visualize the result of applying \algname{F2I}.}
    \label{fig:demo_symmetric}
\end{figure}

\begin{figure}
    \centering
    \includegraphics[width=.95\linewidth]{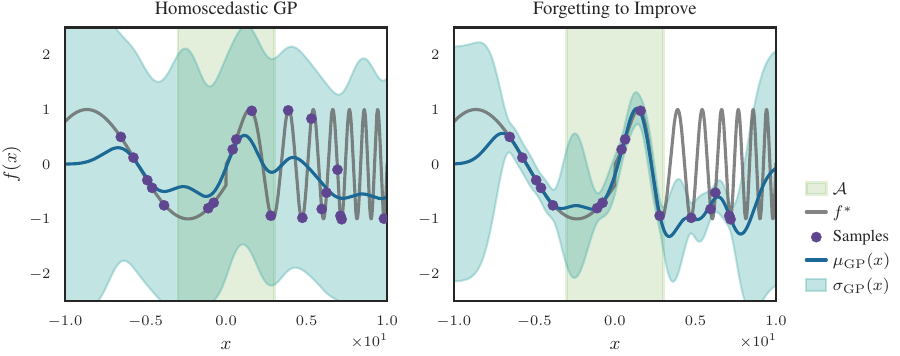}
    \caption{Influence of detrimental data structures based on \emph{Asymmetric LS} test functions. (Left) A standard homoscedastic GP absorbs the detrimental lengthscale structures in the noise. (Right) We visualize the result of applying \algname{F2I}.}
    \label{fig:demo_asymmetric}
\end{figure}

\begin{figure}[ht]
    \centering
    \begin{subfigure}[t]{0.45\linewidth}
        \centering
        \includegraphics[width=\linewidth]{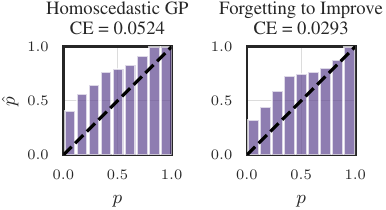}
        \caption{Symmetric lengthscale increase}
    \end{subfigure}
    \hfill
    \begin{subfigure}[t]{0.45\linewidth}
        \centering
        \includegraphics[width=\linewidth]{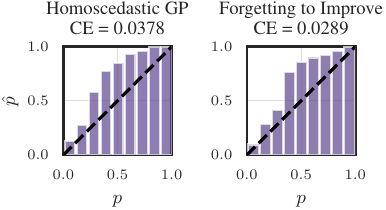}
        \caption{Asymmetric lengthscale increase}
    \end{subfigure}
    \vspace{1em}
    \caption{Calibration comparison of (left) homoscedastic GP and (right) \algname{F2I} for (a) symmetric and (b) asymmetric lengthscale increases test functions.}
    \label{fig:demo_callibration}
\end{figure}

We evaluate \algname{F2I} in the transductive learning setting, in addition to the real world data sets, on two synthetic functions with a detrimental wavelength structure. Therefore, we consider a varying lengthscale across the domain, where our target region-aware algorithm \algname{F2I} significantly improves in prediction accuracy and calibration. We therefore consider the test-functions \emph{Symmetric Lengthscale (LS)} \(f_{symmetric}(x)\) and \emph{Asymmetric Lengthscale (LS)} \(f_{asymmetric}(x)\), which are formally defined by 
\begin{align}
    f_{symmetric}(x) &= \sin(0.5 + 0.39 \cdot |x|),\\  f_{assymetric}(x) & = \begin{cases}\sin(0.5 + 0.39x) & \text{if } x \geq 0\\\sin(0.5 + 0.39\tfrac{1}{x})& \text{otherwise.} \end{cases}
\end{align}
We provide visualizations comparing the homoscedastic GP and \algname{F2I} for both synthetic test functions in \Cref{fig:demo_symmetric,fig:demo_asymmetric}. Further, we also provide the calibration metrics for the two examples in \Cref{fig:demo_callibration}. The two designed test functions do not have specific outliers, but are rather detrimental in their overall design for the chosen kernel with a single lengthscale. We observe, that \emph{relevance pursuit} is outperformed by \algname{F2I}, which is also acknowledged in the limitations section of \cite{ament2024robust} by not considering the location of removed samples in \emph{relevance pursuite}.

In addition, for all benchmarks in the transductive learning setting, we report the domain and task specifications in \Cref{tab:transductive_hyperparams}. We evaluated all methods for 100 random seeds for an RBF kernel with a log prior \(log(0,1)\) on the lengthscale and a Gamma prior with parameters \(a = 2\) and \(b=0.15\) for the aleatoric noise variance. All transductive learning experiments are run on an Intel Core Ultra 9 185H CPU with a NVIDIA RTX 3000 Ada GPU. 

\begin{table}
\centering
\small              
\setlength{\tabcolsep}{4pt}   
\caption{Hyperparameters and settings for the transductive learning benchmarks.}
\label{tab:transductive_hyperparams}
\renewcommand{\arraystretch}{1.3}
\begin{tabular}{lcccccc}
\hline
\textbf{Setting} 
& \textbf{Dow Jones} 
& \textbf{Boston Housing} 
& \textbf{Symmetric LS} 
& \textbf{Asymmetric LS} 
& \textbf{Yacht}
& \textbf{CASP} \\
\hline

$\set{X}$ limits 
& $[-1.5,\,1.5]$ 
& $[-15,\,15]^{13}$ 
& $[-10,\,10]$ 
& $[-10,\,10]$
& $[-5,\,5]^6$
& $[-5,\,5]^9$\\

$\set{A}$ limits 
& $[0.75,\,1.25]$ 
& $[-0.7,\,0.4]^{13}$ 
& $[-3,\,3]$ 
& $[-3,\,3]$ 
& $\substack{[[-0.05, 0.16],\\
[-2.2, 1.6],\\
[-0.5, 1.7],\\
[-0.7, 0.6],\\
[-1.1, 1.7],\\
[-1.7, -1]]}$
&
$\substack{[[-1.5, -1.4],\\
[-1.34, -1.31],\\
[-0.51, -0.48],\\
[-1.27, -1.24],\\
[-1.44, -1.41],\\
[-1.44, -1.41],\\
[-1.12, -1.10],\\
[-1.04, -1.02],\\
[1.65, 1.68]]}$
\\

$|\set{X}|$ 
& $405$ 
& $506$ 
& $1000$ 
& $1000$ 
& $308$
& $45000$\\

$|\set{S}|$ 
& $50$ 
& $40$ 
& $20$ 
& $20$ 
& $40$
& $200$\\
\hline
\end{tabular}
\end{table}

\subsection{Bayesian Optimization} \label{sec:appendix_experiments_bo}
For our BO benchmarks, we provide the standard domain limits, the dimensionality and the global optimum in \Cref{tab:bo_domains}. During the optimization process, we rescale all objective functions into the hypercube \([0,1]^n\), with dimensionality \(n\). Further, we start all experiments from \(5\) initial random observations and proceed with a sampling budget of \(50\) for the remaining optimization. We perform all BO experiments on an Intel Core Ultra 9 185H CPU with a NVIDIA RTX 3000 Ada GPU.

\begin{table}[htbp]
\centering
\caption{Bayesian Optimization Test Functions: Domain Limits and Optimal Values}
\label{tab:bo_domains}
\begin{tabular}{lcccc}
\toprule
\textbf{Function} & \textbf{Dim.} & \textbf{Domain} & \textbf{Optimum} & \textbf{Optimizer} \\
\midrule
Ackley & 2 & $[-10, 30]^2$ & $0.0$ & $(0, 0)$ \\
Ackley & 6 & $[-10, 30]^6$ & $0.0$ & $(0, \ldots, 0)$ \\
A. Rosenbrock & 4 & $[-5,10]^2 \times [1,0]^2$ & $0.0$ & $$(1,\dots, 1)$$\\
Branin & 2 & $[-5, 0] \times [10, 15]$ & $-0.3979$ & $(-3.14, 12.27)$ \\
Griewank & 2 & $[-50, 20]^2$ & $0.0$ & $(0, 0)$ \\
Holder Table & 2 & $[-10, 10]^2$ & $19.21$ & $(8.06, 9.67)$ \\
Hartmann & 6 & $[0, 1]^6$ & $-3.322$ & $(0.2, 0.15, 0.48, 0.28, 0.31, 0.66)$ \\
Levy & 2 & $[-10, 10]^2$ & $0.0$ & $(1, 1)$ \\
\bottomrule
\end{tabular}
\end{table}

\section{Additional Experiments}
Further, we provide in this section different additional experiments to demonstrate the empirical performance and behavior of \algname{F2I}, \algname{Forgetful BO} and \algname{Forgetful RL}. We first showcase the influence of the epistemic term of \algname{F2I} in \Cref{sec:noise-ablation}, and provide additional Bayesian optimization and reinforcement learning experiments in \Cref{sec:appendix_additional_experiments,sec:appendix_experiments_rl}. 

\subsection{Aleatoric Noise Ablation} \label{sec:noise-ablation}
\begin{figure}
    \centering
    \includegraphics[width=1\linewidth]{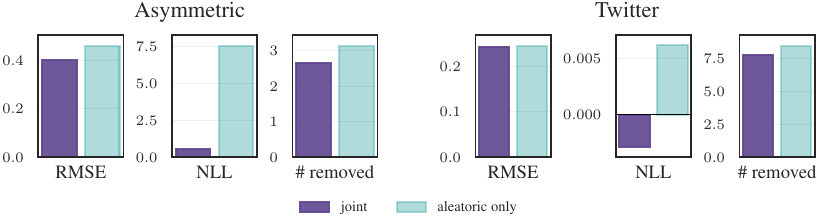}
    \caption{Comparing joint \((u+v)\) and aleatoric-only \((v)\) removal criteria on Asymmetric and Twitter. Incorporating \(u\) prevents detrimental removals, yielding substantially lower NLL with comparable or improved RMSE.}
    \label{fig:noise-ablation}
\end{figure}

We ablate the epistemic term \(u\) by performing removals based solely on the aleatoric term \(v\), rather than \(u+v\). Since \(u\geq 0\), incorporating it can only prevent potentially detrimental removals without inducing additional ones. As shown in \Cref{fig:noise-ablation}, the joint criterion removes slightly fewer points while achieving substantially lower NLL on both benchmarks, with improved RMSE on Asymmetric and comparable RMSE on Twitter.

\subsection{Bayesian Optimization Ablations}\label{sec:appendix_additional_experiments}

\begin{figure}[h]
    \centering
    \includegraphics[width=1\linewidth]{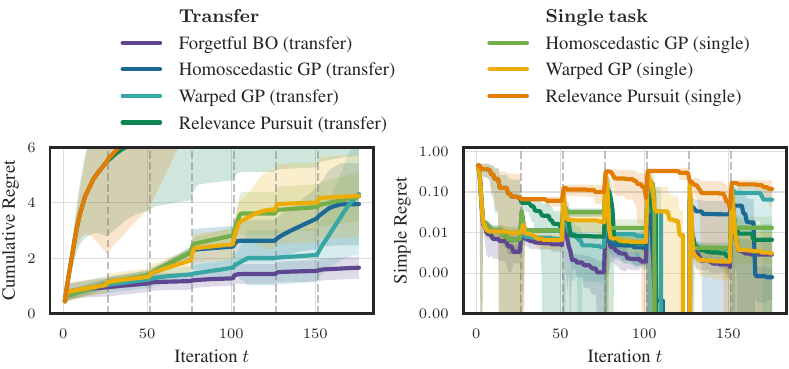}
    \caption{Transfer learning AutoML benchmark for seven sequential GLMNET hyper-parameter training tasks. Comparison of (left) cumulative regret and (right) simple regret }
    \label{fig:auto_ml}
\end{figure}

As a challenging real-world benchmark for Bayesian optimization, we consider hyperparameter tuning for the GLMNET training algorithm \cite{friedman2010regularization} across multiple datasets. Following \cite{schur2023lifelong,perrone2018scalable,rothfuss2021meta}, we replace the expensive evaluation step with a table lookup constructed from a large set of precomputed evaluations \cite{kuhn2018automatic}. Using the first seven classification datasets from OpenML \cite{bischl2017openml}, we study in \Cref{fig:auto_ml} whether \algname{Forgetful BO} can improve transfer learning of hyperparameters across tasks. We evaluate the first seven tasks sequentially, using 25 samples per task. To highlight the effect of forgetting potentially detrimental information from previous datasets, and thus improving transfer across tasks, we compare our method against all baselines in two settings: one that retains all observations from previous tasks, and one that uses only data from the current task.
Most importantly, all baselines benefit from retaining data from previous tasks, since this allows them to transfer knowledge about well-performing configurations across datasets. At the same time, we observe that \algname{Forgetful BO} outperforms all baselines by selectively discarding information that is harmful for subsequent tasks. Both the cumulative regret in the left panel and the simple regret in the right panel confirm the value of forgetting in this setting.

\begin{figure}[h]
    \centering
    \includegraphics[width=1\linewidth]{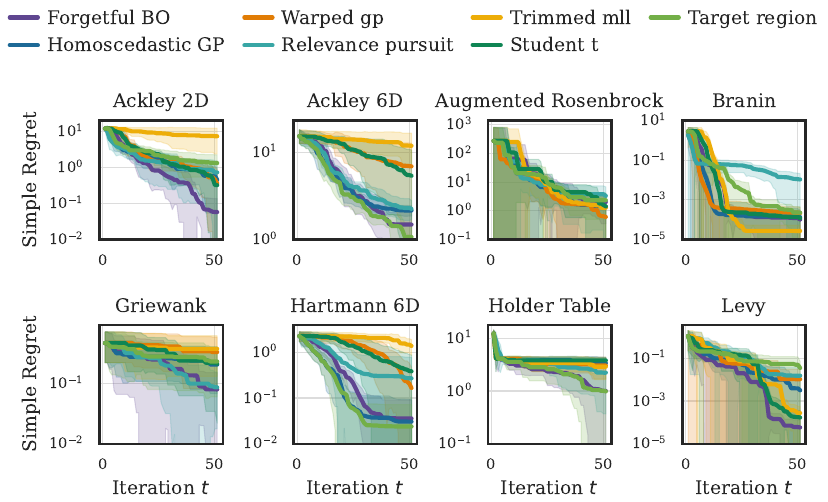}
    \caption{Simple regret of \algname{Forgetful BO} and competing baselines for test functions.}
    \label{fig:bo_other_baselines}
\end{figure}

\begin{figure}[h]
    \centering
    \includegraphics[width=0.95\linewidth]{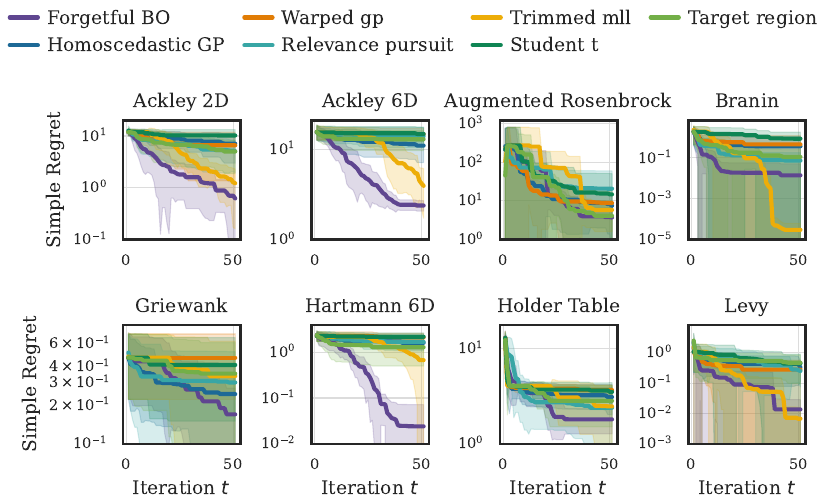}
    \caption{Simple regret of \algname{Forgetful BO} and competing baselines for \emph{perturbed} test functions.}
    \label{fig:bo_noisy_results}
\end{figure}

\begin{figure}[h]
    \centering
    \includegraphics[width=0.95\linewidth]{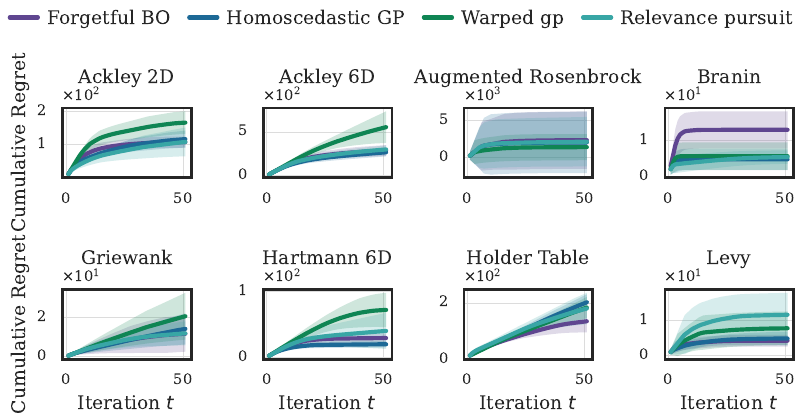}
    \caption{Cumulative regret of \algname{Forgetful BO} and competing baselines for different test functions.}
    \label{fig:bo_cumulative}
\end{figure}

\begin{figure}[h]
    \centering
    \includegraphics[width=0.95\linewidth]{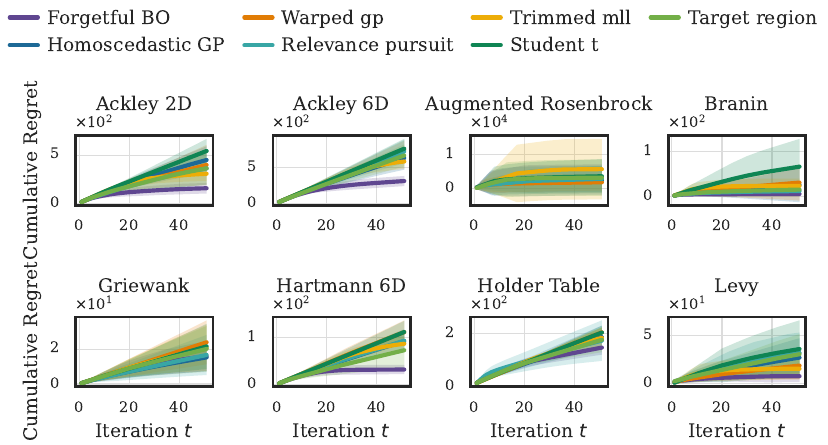}
    \caption{Cumulative regret of \algname{Forgetful BO} and competing baselines for different test functions with constant corruptions.}
    \label{fig:bo_noisy_cumulative}
\end{figure}

\begin{figure}[h]
    \centering
    \includegraphics[width=1\linewidth]{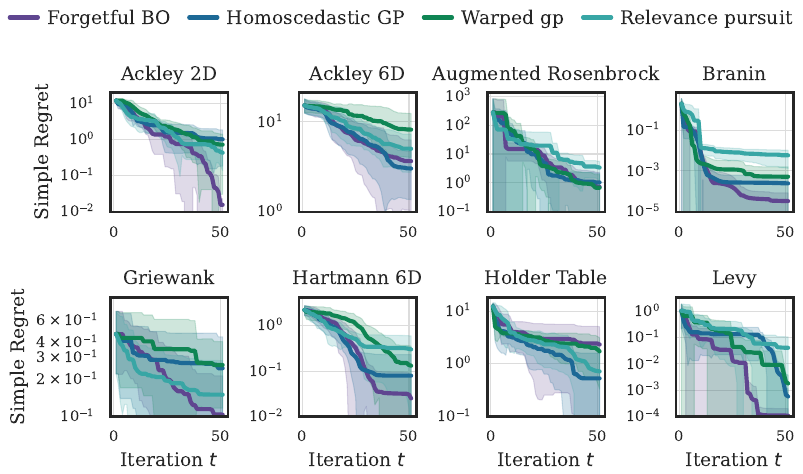}
    \caption{Simple regret of \algname{Forgetful BO} and competing baselines for different test functions using log expected improvement acquisition function.}
    \label{fig:bo_lei}
\end{figure}

\begin{figure}[h]
    \centering
    \includegraphics[width=1\linewidth]{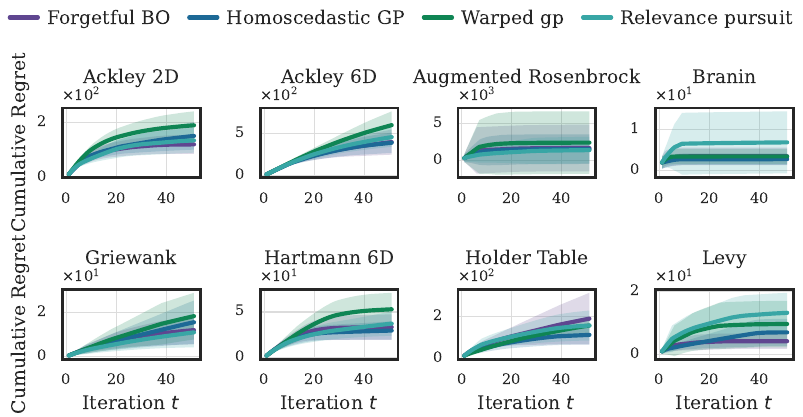}
    \caption{Cumulative regret of \algname{Forgetful BO} and competing baselines for different test functions using log expected improvement acquisition function.}
    \label{fig:bo_lei_cumulative}
\end{figure}

\begin{figure}[h]
    \centering
    \includegraphics[width=1\linewidth]{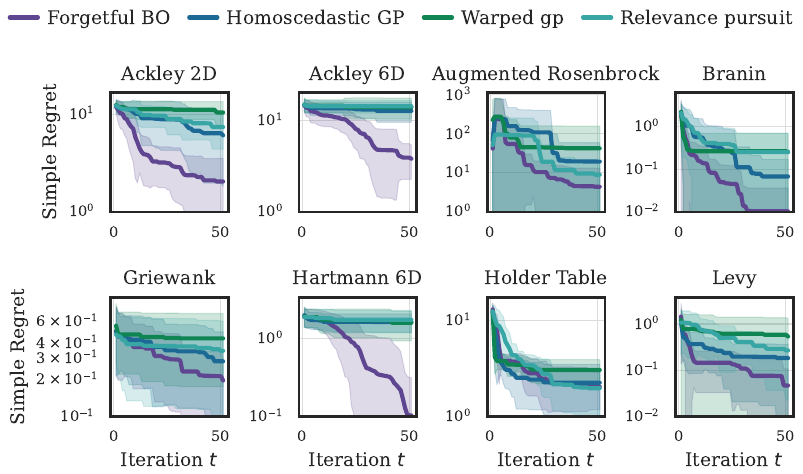}
    \caption{Simple regret of \algname{Forgetful BO} and competing baselines for perturbed test functions using log expected improvement acquisition function.}
    \label{fig:bo_lei_noisy}
\end{figure}

\begin{figure}[h]
    \centering
    \includegraphics[width=1\linewidth]{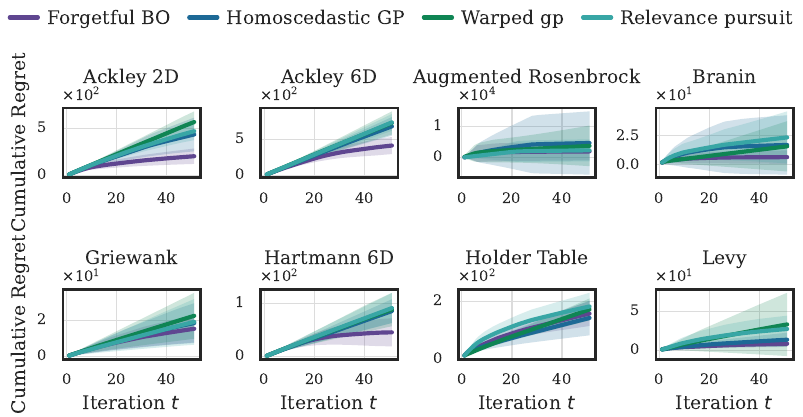}
    \caption{Cumulative regret of \algname{Forgetful BO} and competing baselines for perturbed test functions using log expected improvement acquisition function.}
    \label{fig:bo_lei_noisy_cumulative}
\end{figure}

For \Cref{fig:bo_results}, we add in \Cref{fig:bo_other_baselines} the two additional baselines for robust Bayesian Optimization, Student-t likelihood \cite{martinez2017robust} and trimmed marginal log likelihood \cite{andrade2023robust}, are compared to \algname{Forgetful BO}. We observe that both trimmed MLL and Student-t likelihood are outperformed by \algname{Forgetful BO}.
We extend our evaluation in \Cref{fig:bo_noisy_results} to the outlier setting of \cite{ament2024robust}, where the test functions are perturbed using constant outliers with probability \(p = 0.1\). This experiment demonstrates that \algname{Forgetfull BO} also improves optimization in the perturbed setting and outperforms all competing baselines. Also note that \emph{relevance search} performs better relative to baselines in the perturbed setting, compared to the unperturbed, as mentioned in \cite{ament2024robust}, the algorithm might remove important samples related to the optimum by mistaking them for outliers, due to the dependence on the missing target region. 
Additionally, we provide the cumulative regret results for the BO experiments from \Cref{fig:bo_results,fig:bo_noisy_results}.
\Cref{fig:bo_cumulative} demonstrates that \algname{Forgetful BO} is not only reaching good simple regret, but also has a fast diminishing regret, so that the accumulated value over all iterations \(t=1\dots T\) is lower or on par compared to all competing baselines. 
We also report the cumulative regret for the perturbed setting of \Cref{fig:bo_noisy_results} in \Cref{fig:bo_noisy_cumulative}.
In addition, \algname{Forgetful BO} can seamlessly integrate different acquisition functions than \emph{GP-UCB} \cite{srinivas2009gaussian}. TO demonstrate this, we evaluate \algname{Forgetful BO} on all benchmarks using the \emph{log expected improvement} (LEI) acquisition function \cite{ament2023unexpected}. Therefore, we also use LEI for all competitive baselines and visualize the achieved regret in \Cref{fig:bo_lei}.
We note that, except for the \emph{Ackley 6D} test-function, \algname{Forgetful BO} performs better or on par with competing baselines. The cumulative regret for the LEI experiments are also provided in \Cref{fig:bo_lei_cumulative}. 
To complete this evaluation, we provide the performance of \algname{Forgetful BO} on the noisy benchmarks using the LEI acquisition function in \Cref{fig:bo_lei_noisy,fig:bo_lei_noisy_cumulative}. We observe, that also in this setting \algname{Forgetful BO} demonstrates superior or equal performance across all benchmarks and compared to all baselines. 

\subsection{Bayesian Optimization Computational Complexity} \label{sec:appendix_experiments_bo_timimg}
\begin{figure}[h]
    \centering
    \includegraphics[width=0.9\linewidth]{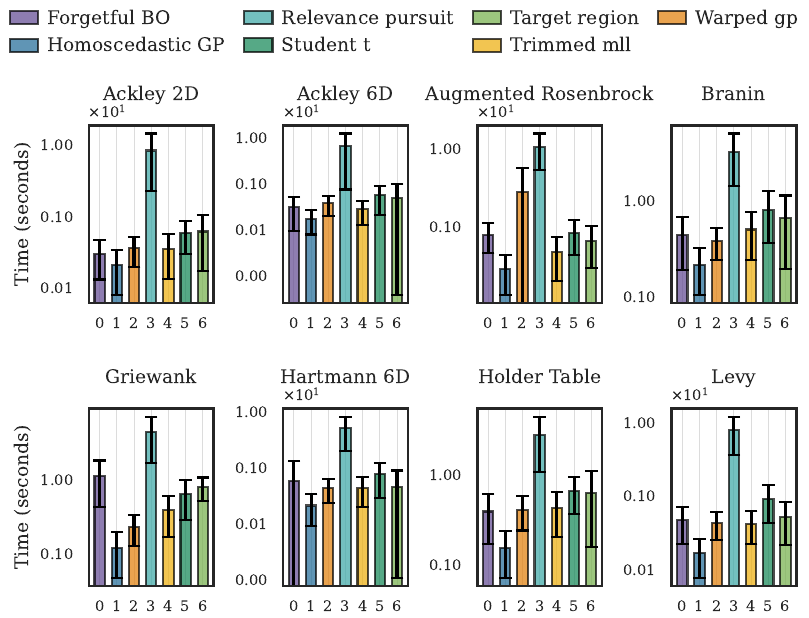}
    \caption{Acquisition time comparisons for the different BO algorithms. Comparing mean and standard deviation across all evaluated non-corrupted benchmarks and baselines.}
    \label{fig:runtimes}
\end{figure}

We demonstrate in \Cref{fig:runtimes}, that \algname{Forgetful BO}, based on the removal algorithm \algname{F2I} is computationally efficient. We report the mean and standard deviation of the acquisition time per sample. We visualize the mean across all random seeds and the entire optimization process of \Cref{fig:bo_results} in \Cref{fig:runtimes}. Note that \algname{Forgetful BO} outperforms \emph{relevance pursuite} by nearly one order of magnitude in the log-scale, while having small computational overhead compared to the \emph{homoscedastic GP} and the \emph{Warped GP}. The large variance of the \algname{Forgetful BO} runtime is caused due to the potentially higher number of removals in later stages of the optimization process. 

\subsection{Reinforcement Learning}\label{sec:appendix_experiments_rl}
\begin{figure}[h]
    \centering
    \includegraphics[width=1\linewidth]{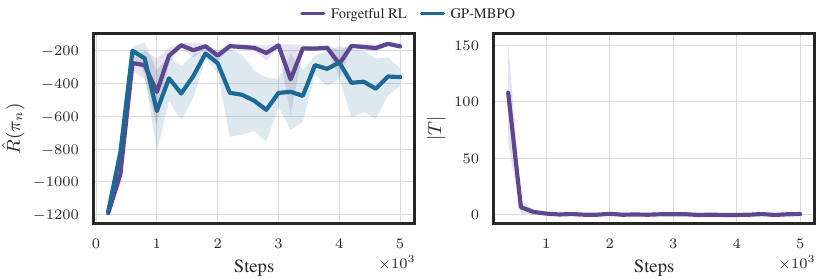}
    \caption{Comparing the performance and data removal of \algname{Forgetful RL} and vanilla GP-based MBPO for \emph{Inverted Pendulum}.}
    \label{fig:inverted_pendulum}
\end{figure}

\begin{figure}[h!]
    \centering
    \includegraphics[width=1\linewidth]{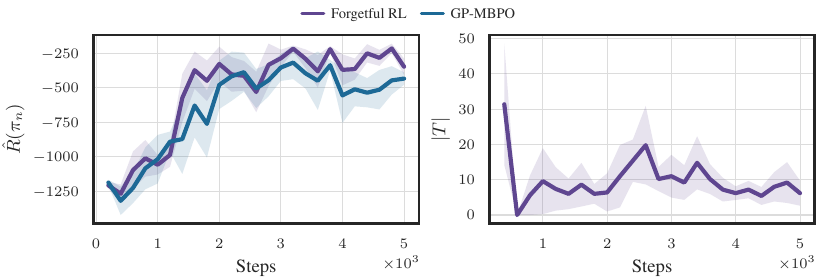}
    \caption{Comparing the performance and data removal of \algname{Forgetful RL} and vanilla GP-based MBPO for \emph{Inverted Pendulum}, beginning with an empty buffer.}
    \label{fig:inverted_pendulum_empty}
\end{figure}

\begin{figure}[h!]
    \centering
    \includegraphics[width=1\linewidth]{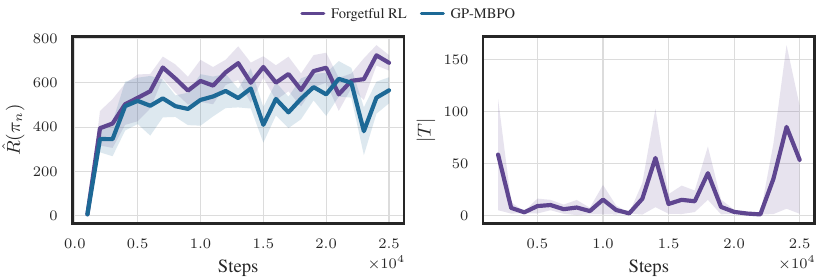}
    \caption{Comparing the performance and data removal of \algname{Forgetful RL} and vanilla GP-based MBPO for \emph{Cartpole Swingup}.}
    \label{fig:cartpole_swingup}
\end{figure}

\begin{figure}[h!]
    \centering
    \includegraphics[width=1\linewidth]{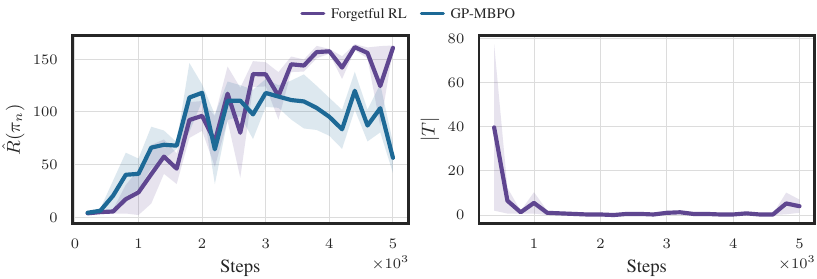}
    \caption{Comparing the performance and data removal of \algname{Forgetful RL} and vanilla GP-based MBPO for \emph{Cartpole Swingup} on a sparse reward task.}
    \label{fig:inverted_pendulum_sparse}
\end{figure}

\looseness = -1
We conduct all reinforcement learning experiments for \algname{Forgetful RL} using a single NVIDIA RTX 4090 GPU per run, with 5 CPU cores. 
Besides the linear kernel experiment in \Cref{fig:rl_linear_kernel}, we also consider tasks, which require a nonlinear model and for the pendulum, for example start in a downward pendulum position and require to swing up. This task nature requires a significantly larger target region, where the dynamics should be modeled well, and which can definitely no longer be approximated with a linear model. We first consider the \emph{inverted pendulum} task in \Cref{fig:inverted_pendulum}. We observe, that relatively little data removal occurs due to a much larger target region, but these lead to better convergence compared to the vanilla MBPO. 
We observe in \Cref{fig:inverted_pendulum}, that we initially remove most points from the replay buffer, that we obtained from the initial random samples prefilling the buffer. We ablate in \Cref{fig:inverted_pendulum_empty} the removal behavior by not filling the buffer with random experiences before the initial episode. 
We observe that we converge slower and require to remove more detrimental experiences throughput learning, which have been gathered during suboptimal episodes. 
Further, we extend the swingup task in \Cref{fig:cartpole_swingup} to \emph{Cartpole Swingup}. 
Analogous to the other \emph{swingup}, tasks we also remove fewer samples than in the \emph{Cartpole Balance} task, because large parts of the state space are visited of the optimal policy, and require detailed modeling. 
We additionally increase the task difficulty of \emph{Inverted Pendulum} by introducing a sparse reward function. In \Cref{fig:inverted_pendulum_sparse}, we observe that the task is significantly harder with a sparse reward and requires longer to converge. \algname{Forgetful RL} also improves in this problem setting the learning behavior and reaches the optimum. 

\newpage
\section{Near-Optimality}
In this section, we both derive the near optimality theory \Cref{sec:appendix_submodular}, and demonstrate empirically the used curvature and optimality diagnostics in \Cref{sec:appendix_optimality_diagnostics} dependent on the chosen noise prior.

\subsection{Proof of the Near-Optimality Guarantee}
\label{sec:appendix_submodular}
In order to establish the near-optimality proof under a curvature condition on the magnitude of the prior parameter \(a\) of the gamma prior \( p(\sigma_w^2(\set{S})) = \frac{b^a}{\Gamma(a)}(\sigma_w^2(\set{S}))^{a-1} e^{-b \sigma_w^2(\set{S})}\), we proceed as follows
\begin{enumerate}
     \item We introduce the normalized removal-gain set function $\bar{\set{J}}$, which reformulates data removal as a greedy maximization problem. As shown in \Cref{lemma:complement}, this transformation preserves the marginal contributions of individual removals.
    
    \item Secondly, we show in \Cref{lemma:rkhs-derivative} that \algname{F2I} selects the removed data points greedily according to the first-order influence gradient \(-u_i(\set{A}) - v_i(\set{A})\) is the gradient \(\tfrac{\partial}{\partial h_i} - \set{J}(\set{A}_i)\) of the negated objective \(-\set{J}\). 
    
    \item We characterize the curvature of the continuous removal objective. Specifically, \Cref{lemma:aleatoric-hessian} derives the sensitivity of the estimated noise variance, while \Cref{lemma:rkhs-hessian} establishes the Hessian
    contributions of epistemic and aleatoric uncertainty.
    Furthermore, \Cref{lemma:inner_opti} provides conditions on the Gamma-prior parameter $a$ ensuring strict concavity of the inner noise optimization problem. The resulting Hessians characterize local curvature; the discrete restricted strong concavity (RSC) condition in \Cref{def:rsc} is assumed separately
    for the near-optimality analysis
    
   \item Using RSC, we establish an upper bound $U_q$ on  the optimal improvement achievable by removing $q$ observations, without requiring knowledge of the optimal removal set. We also impose restricted gradient dominance according to \Cref{assumption:gradient-dominance} along the executed removal sequence, ensuring that the influence gradients remain sufficiently informative about the remaining improvement.

    \item Finally, \Cref{lemma:gradient-dominance-progress} combines gradient dominance with the realized-step coefficient $\rho$, which accounts for the actual
    improvement obtained from binary deletion. This yields the greedy-progress inequality and, consequently, the conditional exponential approximation guarantee in \Cref{thrm:forgetting_optimality}.
\end{enumerate}

\vspace{1em}
In the following, we list all relevant lemmas and definitions required to establish \Cref{thrm:forgetting_optimality}:
\vspace{1em}

\begin{lemma}\label{lemma:complement}
Consider the set of given samples \(\set{S}\) and the objective \(\min_{\set{S}'\subseteq \set{S}} \obj(\set{S}')\) in \Cref{eq:obj}. Let the removed set of samples be \(T = \set{S}\setminus \set{S}'\), then we defined the set function \(\comp(T)\) as
\begin{equation*}
    \max_{T \subseteq \set{S}} \comp (T) \coloneqq \obj(\set{S}) - \obj(\set{S}') =\obj(\set{S})  -\obj(\set{S}\setminus T).
\end{equation*}
This set function has the same form as \cite[Equation~4]{elenberg2018restricted}, with the submodularity ratio \(\gamma_{L,U}\), which is defined for any \(L \subseteq T\) and \(U \subseteq T\setminus L\) as 
\begin{equation*}
    \gamma_{L,U} (\comp) = \frac{\sum_{j\in U}(\comp(L \cup \{j\}) - \comp(L))}{\comp(L \cup U) - \comp(L)}.
\end{equation*}  
\end{lemma}

\begin{proof}
Consider a retained set \(\set{S}'\) and its complement set \(T=\set{S}\setminus \set{S}'\). Additionally let \(L \subseteq T\) and \(U \subseteq T \setminus L\), and \(j\in U\),
\begin{align}\label{eq:marginal_contribution_complement}
\bar{\mathcal J}(L\cup\{j\})-\bar{\mathcal J}(L) & = \obj(\set{S}) - \mathcal J(\set{S}\setminus (L \cup \{j\})) -  \obj(\set{S}) + \mathcal J(\set{S} \setminus L)\\
& = \mathcal J(\set{S} \setminus L) - \mathcal J(\set{S}\setminus (L \cup \{j\})) 
\end{align}
Thus, the forward marginal on \(\bar{\mathcal J}\) equals the removal marginal on \(- \mathcal J\). The submodularity ratio defined in \cite[Def.~2]{elenberg2018restricted} depends only on the marginal combinations. Hence, the submodularity ratio for the set-function considering the complement removal sets \(L \subseteq T\) and \(U \subseteq T \setminus L\) is given by
\begin{equation}
    \gamma_{L,U} (\comp) = \frac{\sum_{j\in U}(\comp(L \cup \{j\}) - \comp(L))}{\comp(L \cup U) - \comp(L)}.
\end{equation}
\end{proof}

\begin{lemma} \label{lemma:rkhs-derivative}
Let \(\set{S}\subseteq\set{X}\) and an individual sample \(x_i\in \set{S}\), then the negated objective is given by
\begin{equation*}
- \mathcal J (\set{S}) = -\sum_{x \in \set{A}}\sigma_{\set{S}}^2(x).
\end{equation*}
Let the RKHS removal direction associated with a sample \(i\) be
\begin{equation*}
    k(x,\set{S}, h_i)=k_{x,\set{S}}- h_i k(x,x_i)e_i,
\qquad h_i \in [0,1] \subset \mathbb R,
\end{equation*}
i.e. an infinitesimal decrease of the kernel coupling to \(x_i\).
Then the directional derivative of \(-\mathcal J\) at \(\set{S}\) in this RKHS removal direction satisfies
\begin{equation*}
    \left.\frac{\partial}{\partial h_i} - \mathcal {J}\bigl(\set{A}_i\bigr)\right|_{h_i=0} = - u_i(\set{A}) - v_i(\set{A}),
\end{equation*}
where \(u_i(\set{A})\) and \(v_i(\set{A})\) are the epistemic and aleatoric influences defined in \Cref{prop:epistemic_influence,prop:aleatoric_influence}.
\end{lemma}

\begin{proof}
We compute the gradient of \(- \obj(\set{S})\) under an infinitesimal reduction of the RKHS basis component \(k(\cdot,x_i)\).
We decompose the predictive variance formulation as follows
\begin{equation}
    - \mathcal J(\set{S})= \sum_{x  \in \set{A}} \left( - k(x,x) + k_{x,\set{S}} A^{-1}k_{\set{S},x} \right) - |\set{A}|\sigma_w^2(\set{S}),
\end{equation}
where \(|\set{A}|\) is the Lebesgue measure of the target are \(\set{A}\). We next differentiate every term that is dependent on the removal parameter \(h_i\). Since \(- \mathcal{J}\) is continuous in \(x \in \set{A}\) and we have defined our surrogate function to be continuous in \(h_i\), we interchange the integral over \(\set{A}\) and the derivative 
\begin{align}
    \frac{\partial  - \mathcal J(\set{S})}{\partial h_i} =& \sum_{x \in \set{A}}\left( - \frac{\partial}{\partial h_i}k(x,x) + \frac{\partial}{\partial h_i}k(x,\set{S},h_i) A(h_i)^{-1}k(\set{S},x,h_i)\right) - |\set{A}|\frac{\partial}{\partial h_i}\sigma_w^2(\set{S})\\
    =& \sum_{x \in \set{A}} \left( 2\frac{\partial k(x,\set{S},h_i)}{\partial h_i} A^{-1}k_{\set{S},x} - k(x,\set{S},h_i) A(h_i)^{-1}\frac{\partial A(h_i)}{\partial h_i} A(h_i)^{-1}\right. \\  & \times  k(\set{S},x,h_i)\bigg)  - |\set{A}|\frac{\partial}{\partial h_i}\sigma_w^2(\set{S}')\nonumber\\
    =& \sum_{x \in \set{A}} \left(- 2 k(x,x_i)(A^{-1} k(\set{S},x))_i -  k(x,\set{S},h_i) A(h_i)^{-1}\frac{\partial K(h_i)}{\partial h_i} A(h_i)^{-1}k(\set{S},x,h_i) \right.\\ 
    &- \left. k(x,\set{S},h_i) A(h_i)^{-1}\frac{\partial\sigma_w^2(\set{S})}{\partial h_i} A(h_i)^{-1}k(\set{S},x,h_i)\right) - |\set{A}| \frac{\partial}{\partial h_i}\sigma_w^2(\set{S})\nonumber\\
    = & \sum_{x \in \set{A}} - 2 k(x,x_i)(A^{-1} k(\set{S},x))_i -  k(x,\set{S},h_i) A(h_i)^{-1}(-e_ik(x_i,\set{S})- k(\set{S},x_i)e_i^\top) \\ &\times A(h_i)^{-1}k(\set{S},x,h_i) - \frac{\partial}{\partial h_i}\sigma_w^2(\set{S}) (1+k(x,\set{S},h_i) A(h_i)^{-2}k(\set{S},x,h_i)) \nonumber\\
    = & \sum_{x \in \set{A}}  \underbrace{- 2 k(x,x_i)(A^{-1} k(\set{S},x))_i + 2 (A^{-1}k(\set{S},x))_i (k(x_i,\set{S})A^{-1}k(\set{S},x))}_{-u_i(x)}\\
    & \underbrace{- (1+k(x,\set{S},h_i) A(h_i)^{-2}k(\set{S},x,h_i)) \frac{\partial}{\partial h_i}\sigma_w^2(\set{S})}_{- v_i(x)} \nonumber
\end{align}
This completes the proof that \(-u_i(\set{A})-v_i(\set{A})\) is the directional derivative of the negated objective \(- \mathcal J\) in the RKHS removal direction induced by the sample \(x_i\), when evaluating at \(h_i=0\).
\end{proof}

\begin{definition}[Restricted Strong Concavity, Restricted Smoothness]\label{def:rsc}
    A function \(- \mathcal J(\cdot )\) is said to be restricted strong concave (RSC) with \(m_R \in \mathbb{R}_+\) and restricted smooth (RSM) with \(M_R \in \mathbb{R}_+\) on the set \(R\), if 
    \begin{equation*}
        -\frac{m_R}{2}\|h\|_2^2 \geq \mathcal{J}(\set{S}\setminus L) - \mathcal{J}(\set{S}\setminus (L\cup U))  - \langle \nabla - \mathcal{J}(\set{S}\setminus L),h\rangle \geq -\frac{M_R}{2}\|h\|_2^2,
    \end{equation*}
    where \(L \cup U \subseteq R \), and \(h\) is the indicator vector of the removed points in \(L \cup U\) compared to \(L\).
\end{definition}
\mbox{}

\begin{lemma}\label{lemma:aleatoric-hessian}
Let $\set{S}$ be a fixed sample set and let $\sigma_w^2(\set{S})$ denote the type II MAP estimate, that satisfies \(\mathcal{D}(\sigma_w^2,\set{S},a,b) = 0\) from \Cref{prop:aleatoric_derivative}. Suppose that the type II MAP solution is interior, twice continuously differentiable in \(h\) around \(h=0\), and that \(T\neq 0\). Then the Hessian of \(\sigma_w^2(\set{S})\) w.r.t. the removal directions in \(h\), evaluated at \(h=0\), is given by
\begin{align*}
    \nabla_{h}^2 \sigma_w^2 =& -\frac{1}{T}\Big(\nabla_{h}^2 \mathcal{D}(\sigma_w^2;y) + \nabla_{\sigma_w^2,h}^2 \mathcal{D}(\sigma_w^2;y)(\nabla_h \sigma_w^2)^\top \\ & + \nabla_h \sigma_w^2(\nabla_{\sigma_w^2,h}^2 \mathcal{D}(\sigma_w^2;y))^\top+ \nabla_{\sigma_w^2} T\,(\nabla_h \sigma_w^2)(\nabla_h \sigma_w^2)^\top\Big),
\end{align*}
where we define the following sub-parts of the derivatives as
\begin{align*}
    \nabla_h \sigma_w^2  = & \frac{1}{T}\big(-\alpha^{(1)} \odot s^{(2)} - \alpha^{(2)} \odot s^{(1)} + \diag (M^{(2)}) + y \odot \alpha^{(2)}\big),\\
    \nabla_{\sigma_w^2,h}^2 \mathcal{D}(\sigma_w^2;y) = & -\big(2 \alpha^{(2)} \odot s^{(2)} + 2 \alpha^{(1)}\odot s^{(3)} + 2 \alpha^{(3)} \odot s^{(1)} - 2 \diag (M^{(3)})\big) + 2\, y \odot \alpha^{(3)}, \\
    \nabla_{h}^2 \mathcal{D}(\sigma_w^2;y) = & A^{-1}\odot\big(s^{(1)}s^{(2)\top} + s^{(2)}s^{(1)\top} - y s^{(2)\top} - s^{(2)}y^\top\big) + A^{-2}\odot (s^{(1)}-y)(s^{(1)}-y)^\top \\
    &+ M^{(1)}\odot\big(\alpha^{(1)}s^{(2)\top} + s^{(2)}\alpha^{(1)\top} + \alpha^{(2)}s^{(1)\top} + s^{(1)}\alpha^{(2)\top} - y\alpha^{(2)\top} - \alpha^{(2)}y^\top\big) \\
    &+ M^{(2)}\odot \big(\alpha^{(1)}(s^{(1)}-y)^\top + (s^{(1)}-y)\alpha^{(1)\top}\big) + M^{(1)}K\odot \big(\alpha^{(1)}\alpha^{(2)\top} + \alpha^{(2)}\alpha^{(1)\top}\big) \\
    &+ M^{(2)}K\odot \alpha^{(1)}\alpha^{(1)\top} - \big(2\,M^{(1)}\odot M^{(2)} + A^{-1}\odot M^{(2)}K + A^{-2}\odot M^{(1)}K\big)\\
    &+ K\odot\big(A^{-2} - \alpha^{(1)}\alpha^{(2)\top} - \alpha^{(2)}\alpha^{(1)\top}\big),\\
    \nabla_{\sigma_w^2} T = & 3 \alpha^{(2)\top}\alpha^{(2)} -\tr(A^{-3}) + \frac{2(a-1)}{(\sigma_w^2)^3},
\end{align*}
using the short hand notations
\begin{align*}
    &\alpha^{(p)} = A^{-p}y,\quad M^{(p)} = K A^{-p}, \quad s^{(p)} = K\alpha^{(p)}, \\ &T = -\alpha^{(1)\top}\alpha^{(2)} + \frac{1}{2}\tr(A^{-2}) - \frac{a-1}{(\sigma_w^2)^2}.
\end{align*}
\end{lemma}

\begin{proof}
We obtained the first derivative from the implicit function theorem
\begin{equation}\label{eq:implicitFunctionIdentity}
    \frac{\partial \mathcal{D}(\sigma_w^2;y)}{\partial \sigma_w^2}\frac{\partial \sigma_w^2}{\partial h_i} + \frac{\partial \mathcal{D}(\sigma_w^2;y)}{\partial h_i} = 0.
\end{equation}
By differentiating the identity in \Cref{eq:implicitFunctionIdentity} w.r.t. \(h_j\), we obtain
\begin{align}
    \frac{\partial}{\partial h_j} \left( \frac{\partial \mathcal{D}(\sigma_w^2;y)}{\partial \sigma_w^2}\frac{\partial \sigma_w^2}{\partial h_i}\right. & +\left. \frac{\partial \mathcal{D}(\sigma_w^2;y)}{\partial h_i}\right) =
    \left(\frac{\partial^2 \mathcal{D}(\sigma_w^2;y)}{\partial \sigma_w^2\partial h_j} + \frac{\partial^2 \mathcal{D}(\sigma_w^2;y)}{\partial (\sigma_w^2)^2}\frac{\partial \sigma_w^2}{\partial h_j}\right)\frac{\partial \sigma_w^2}{\partial h_i} \\ &+ \frac{\partial^2 \mathcal{D}(\sigma_w^2;y)}{\partial h_i\partial h_j}  + \frac{\partial^2 \mathcal{D}(\sigma_w^2;y)}{\partial \sigma_w^2\partial h_i}\frac{\partial \sigma_w^2}{\partial h_j} + \frac{\partial \mathcal{D}(\sigma_w^2;y)}{\partial \sigma_w^2}\frac{\partial^2 \sigma_w^2}{\partial h_i\partial h_j} = 0. \nonumber
\end{align}
Reordering the terms yields the desired second derivative from the implicit function theorem
\begin{align}
    \frac{\partial^2 \sigma_w^2}{\partial h_i\partial h_j} =  - \frac{1}{\frac{\partial \mathcal{D}(\sigma_w^2;y)}{\partial \sigma_w^2}}& \bigg(\frac{\partial^2 \mathcal{D}(\sigma_w^2;y)}{\partial \sigma_w^2\partial h_j}\frac{\partial \sigma_w^2}{\partial h_i} + \frac{\partial^2 \mathcal{D}(\sigma_w^2;y)}{\partial (\sigma_w^2)^2}\frac{\partial \sigma_w^2}{\partial h_j}\frac{\partial \sigma_w^2}{\partial h_i}  \\ & + \frac{\partial^2 \mathcal{D}(\sigma_w^2;y)}{\partial h_i\partial h_j} + \frac{\partial^2 \mathcal{D}(\sigma_w^2;y)}{\partial \sigma_w^2\partial h_i}\frac{\partial \sigma_w^2}{\partial h_j} \bigg).\nonumber
\end{align}
We now differentiate each term to obtain the full Hessian. For convenience, we introduce the abbreviation \(\alpha^{(p)} = A^{-p}y\). The derivatives \(\frac{\partial \sigma_w^2}{\partial h_i}\) and \(\frac{\partial \sigma_w^2}{\partial h_j}\) are given by the implicit function theorem in \Cref{prop:aleatoric_derivative} as
\begin{align}\label{eq:sigma_derivative}
    \frac{\partial \sigma_w^2}{\partial h_i} =&  -\frac{\frac{\partial \mathcal{D}(\sigma_w^2;y)}{\partial h_i}}{\frac{\partial \mathcal{D}(\sigma_w^2;y)}{\partial \sigma_w^2}} = \frac{-\alpha^{(1)}_i(k_{x_i,\set{S}}\alpha^{(2)})-\alpha^{(2)}_i(k_{x_i,\set{S}}\alpha^{(1)}) + k_{x_i,\set{S}}(A^{-2})_{\cdot, i}+ y_i(A^{-2}y)_i}{-\alpha^{(1)\top}\alpha^{(2)} + \frac{1}{2}\tr(A^{-2}) - \frac{a-1}{(\sigma_w^2)^2}}, \\
    \frac{\partial \sigma_w^2}{\partial h_j} =&  -\frac{\frac{\partial \mathcal{D}(\sigma_w^2;y)}{\partial h_j}}{\frac{\partial \mathcal{D}(\sigma_w^2;y)}{\partial \sigma_w^2}} = \frac{-\alpha^{(1)}_j(k_{x_j,\set{S}}\alpha^{(2)})-\alpha^{(2)}_j(k_{x_j,\set{S}}\alpha^{(1)}) + k_{x_j,\set{S}}(A^{-2})_{\cdot, j}+ y_j(A^{-2}y)_j}{-\alpha^{(1)\top}\alpha^{(2)} + \frac{1}{2}\tr(A^{-2}) - \frac{a-1}{(\sigma_w^2)^2}}. \nonumber
\end{align}

The denominator is given by
\begin{align}
    \frac{\partial \mathcal{D}(\sigma_w^2;y)}{\partial \sigma_w^2} = & - y^\top A^{-3}y + \frac{1}{2}\tr(A^{-2}) - \frac{a-1}{(\sigma_w^2)^2} \\
    = & -\alpha^{(1)\top}\alpha^{(2)} + \frac{1}{2}\tr(A^{-2}) - \frac{a-1}{(\sigma_w^2)^2}.\nonumber
\end{align}
The second derivative of this term is given by
\begin{align}
    \frac{\partial^2 \mathcal{D}(\sigma_w^2;y)}{\partial (\sigma_w^2)^2} = &  3y^\top A^{-4}y - \tr(A^{-3}) + \frac{2(a-1)}{(\sigma_w^2)^3} \\
    = & 3 \alpha^{(2)\top}\alpha^{(2)} - \tr(A^{-3}) + \frac{2(a-1)}{(\sigma_w^2)^3}.\nonumber
\end{align}

Further, the cross derivatives \(\frac{\partial^2 \mathcal{D}(\sigma_w^2;y)}{\partial \sigma_w^2\partial h_j}\) and \(\frac{\partial^2 \mathcal{D}(\sigma_w^2;y)}{\partial \sigma_w^2\partial h_i}\) follow by differentiating \(\frac{\partial\mathcal{D}}{\partial h_i} = \alpha^{(1)}_i s^{(2)}_i + \alpha^{(2)}_i s^{(1)}_i - \diag(M^{(2)})_i - y_i\alpha^{(2)}_i\) w.r.t.\ \(\sigma_w^2\), using \(\frac{\partial \alpha^{(p)}}{\partial \sigma_w^2} = -p\,\alpha^{(p+1)}\) and \(\frac{\partial s^{(p)}}{\partial \sigma_w^2} = -p\,s^{(p+1)}\):
\begin{align}\label{eq:mixed_derivatives}
    \frac{\partial^2 \mathcal{D}(\sigma_w^2;y)}{\partial \sigma_w^2\partial h_i} = & -2\alpha^{(2)}_i(k_{x_i,\set{S}}\alpha^{(2)}) - 2\alpha^{(1)}_i(k_{x_i,\set{S}}\alpha^{(3)}) - 2 \alpha^{(3)}_i(k_{x_i,\set{S}}\alpha^{(1)}) \\ &+ 2 k_{x_i,\set{S}}(A^{-3})_{\cdot,i} + 2y_i\alpha_i^{(3)},
    \nonumber\\ \frac{\partial^2 \mathcal{D}(\sigma_w^2;y)}{\partial \sigma_w^2\partial h_j} = & -2\alpha^{(2)}_j(k_{x_j,\set{S}}\alpha^{(2)}) - 2\alpha^{(1)}_j(k_{x_j,\set{S}}\alpha^{(3)}) - 2 \alpha^{(3)}_j(k_{x_j,\set{S}}\alpha^{(1)}) \\ & + 2 k_{x_j,\set{S}}(A^{-3})_{\cdot,j} + 2y_j\alpha_j^{(3)}.\nonumber
\end{align}
Finally, we derive \(\frac{\partial^2 \mathcal{D}(\sigma_w^2;y)}{\partial h_i\partial h_j}\). By the continuous removal direction in \Cref{prop:removalDirection}, we have \(A(h)=P(h)KP(h)+\sigma_w^2I\) and \(y(h)=P(h)y\), with \(P(h)=\diag(1-h)\). Thus, \(y(h)\) is affine in \(h\), whereas \(A(h)\) is quadratic. At \(h=0\), we obtain
\begin{align}
    \frac{\partial A}{\partial h_i} &= -(e_ie_i^\top K + K e_ie_i^\top), \qquad \frac{\partial y}{\partial h_i} = -y_ie_i, \\
    \frac{\partial^2 A}{\partial h_i\partial h_j} &= e_ie_i^\top K e_je_j^\top + e_je_j^\top K e_ie_i^\top, \qquad \frac{\partial^2 y}{\partial h_i\partial h_j}=0. \nonumber
\end{align}
The first derivative reads
\begin{equation}
    \frac{\partial \mathcal{D}(\sigma_w^2;y)}{\partial h_i} = -y_i\alpha^{(2)}_i - \alpha^{(1)\top}\frac{\partial A}{\partial h_i}\alpha^{(2)} + \frac{1}{2}\tr\Big(A^{-2}\frac{\partial A}{\partial h_i}\Big),
\end{equation}
and differentiating once more w.r.t. \(h_j\) gives
\begin{align}
    \frac{\partial^2 \mathcal{D}(\sigma_w^2;y)}{\partial h_i\partial h_j} = & -y_i\Big(\frac{\partial \alpha^{(2)}}{\partial h_j}\Big)_i - \Big(\frac{\partial \alpha^{(1)}}{\partial h_j}\Big)^\top\frac{\partial A}{\partial h_i}\alpha^{(2)} - \alpha^{(1)\top}\frac{\partial A}{\partial h_i}\frac{\partial \alpha^{(2)}}{\partial h_j} \\
    & - \alpha^{(1)\top}\frac{\partial^2 A}{\partial h_i\partial h_j}\alpha^{(2)}
    - \frac{1}{2}\tr\Big(\Big(A^{-1}\frac{\partial A}{\partial h_j}A^{-2} + A^{-2}\frac{\partial A}{\partial h_j}A^{-1}\Big)\frac{\partial A}{\partial h_i}\Big)\nonumber\\
    &+ \frac{1}{2}\tr\Big(A^{-2}\frac{\partial^2 A}{\partial h_i\partial h_j}\Big).\nonumber
\end{align}
with
\begin{equation}
    \frac{\partial \alpha^{(1)}}{\partial h_j} = A^{-1}\Big(\frac{\partial y}{\partial h_j} - \frac{\partial A}{\partial h_j}\alpha^{(1)}\Big), \qquad \frac{\partial \alpha^{(2)}}{\partial h_j} = A^{-1}\Big(\frac{\partial \alpha^{(1)}}{\partial h_j} - \frac{\partial A}{\partial h_j}\alpha^{(2)}\Big).
\end{equation}
At \(h = 0\), \(A^{-1}\) and \(K\) commute, such that \(M^{(p)}\) and \(M^{(p)}K\) are symmetric. Inserting the derivatives of \(A\) and \(y\) gives
\begin{align}\label{eq:double_derivative}
    \frac{\partial^2 \mathcal{D}(\sigma_w^2;y)}{\partial h_i\partial h_j} = & (A^{-1})_{ij}\big(s^{(1)}_is^{(2)}_j + s^{(2)}_is^{(1)}_j - y_is^{(2)}_j - s^{(2)}_iy_j\big) + (A^{-2})_{ij}\big(s^{(1)}_i - y_i\big)\big(s^{(1)}_j - y_j\big) \nonumber\\
    & + M^{(1)}_{ij}\big(\alpha^{(1)}_is^{(2)}_j + s^{(2)}_i\alpha^{(1)}_j + \alpha^{(2)}_is^{(1)}_j + s^{(1)}_i\alpha^{(2)}_j - y_i\alpha^{(2)}_j - \alpha^{(2)}_iy_j\big) \nonumber\\
    & + M^{(2)}_{ij}\big(\alpha^{(1)}_i(s^{(1)}_j - y_j) + (s^{(1)}_i - y_i)\alpha^{(1)}_j\big) + (M^{(1)}K)_{ij}\big(\alpha^{(1)}_i\alpha^{(2)}_j \\ 
    &
    + \alpha^{(2)}_i\alpha^{(1)}_j\big) + (M^{(2)}K)_{ij}\,\alpha^{(1)}_i\alpha^{(1)}_j \nonumber\\
    & - \big(2M^{(1)}_{ij}M^{(2)}_{ij} + (A^{-1})_{ij}(M^{(2)}K)_{ij} + (A^{-2})_{ij}(M^{(1)}K)_{ij}\big)\nonumber\\
    &+ K_{ij}\big((A^{-2})_{ij}-\alpha_i^{(1)}\alpha_j^{(2)}-\alpha_i^{(2)}\alpha_j^{(1)}\big).\nonumber
\end{align}
Next, we express all terms in matrix vector notation to obtain the full Hessian matrix. Therefore, we used the following additional definitions
\begin{align}
    M^{(p)} = K A^{-p}, \quad s^{(p)} = K\alpha^{(p)}, \quad T = \frac{\partial \mathcal{D}(\sigma_w^2;y)}{\partial \sigma_w^2}.
\end{align}
First, the gradient \(\nabla_h \sigma_w^2\) follows from \Cref{eq:sigma_derivative} as
\begin{equation}
    \nabla_h \sigma_w^2 = \frac{1}{T}\big(-\alpha^{(1)} \odot s^{(2)} - \alpha^{(2)} \odot s^{(1)} + \diag (M^{(2)}) + y\odot\alpha^{(2)}\big).
\end{equation}
Further, the mixed terms \(\nabla_{\sigma_w^2,h}^2 \mathcal{D}(\sigma_w^2;y)\) from \Cref{eq:mixed_derivatives} simplify to
\begin{align}
    \nabla_{\sigma_w^2,h}^2 \mathcal{D}(\sigma_w^2;y) = -\big(2 \alpha^{(2)} \odot s^{(2)} + 2 \alpha^{(1)}\odot s^{(3)} + 2 \alpha^{(3)} \odot s^{(1)} - 2 \diag (M^{(3)})\big) + 2\,y\odot \alpha^{(3)}.
\end{align}
And the term \(\frac{\partial^2 \mathcal{D}(\sigma_w^2;y)}{\partial h_i\partial h_j}\) of \Cref{eq:double_derivative} is given by
\begin{align}
    \nabla_{h}^2 \mathcal{D}(\sigma_w^2;y) = & A^{-1}\odot\big(s^{(1)}s^{(2)\top} + s^{(2)}s^{(1)\top} - y s^{(2)\top} - s^{(2)}y^\top\big) \\
    &+ A^{-2}\odot (s^{(1)}-y)(s^{(1)}-y)^\top + M^{(1)}\odot\big(\alpha^{(1)}s^{(2)\top} + s^{(2)}\alpha^{(1)\top} \nonumber\\
    &+ \alpha^{(2)}s^{(1)\top} + s^{(1)}\alpha^{(2)\top} - y\alpha^{(2)\top} - \alpha^{(2)}y^\top\big) + M^{(2)}\odot \big(\alpha^{(1)}(s^{(1)}-y)^\top \nonumber\\
    &+ (s^{(1)}-y)\alpha^{(1)\top}\big) + M^{(1)}K\odot \big(\alpha^{(1)}\alpha^{(2)\top} + \alpha^{(2)}\alpha^{(1)\top}\big) \nonumber\\
    &+ M^{(2)}K\odot \alpha^{(1)}\alpha^{(1)\top} - \big(2\,M^{(1)}\odot M^{(2)} + A^{-1}\odot M^{(2)}K + A^{-2}\odot M^{(1)}K\big)\nonumber\\
    &+ K\odot\big(A^{-2} - \alpha^{(1)}\alpha^{(2)\top} - \alpha^{(2)}\alpha^{(1)\top}\big).\nonumber
\end{align}
Finally, the last term as the gradient of the denominator \(\nabla_{\sigma_w^2} T\) as
\begin{equation}
    \nabla_{\sigma_w^2} T = 3 \alpha^{(2)\top}\alpha^{(2)} -\tr(A^{-3}) + \frac{2(a-1)}{(\sigma_w^2)^3}.
\end{equation}
So that we can write the final Hessian as
\begin{align}
    \nabla_{h}^2 \sigma_w^2 =& -\frac{1}{T}\Big(\nabla_{h}^2 \mathcal{D}(\sigma_w^2;y) + \nabla_{\sigma_w^2,h}^2 \mathcal{D}(\sigma_w^2;y)(\nabla_h \sigma_w^2)^\top \\ & + \nabla_h \sigma_w^2(\nabla_{\sigma_w^2,h}^2 \mathcal{D}(\sigma_w^2;y))^\top+ \nabla_{\sigma_w^2} T\,(\nabla_h \sigma_w^2)(\nabla_h \sigma_w^2)^\top\Big).\nonumber
\end{align}
\end{proof}

\begin{lemma}\label{lemma:rkhs-hessian}
Let $\set{S}\subseteq \mathcal X$ and sample $x_i\in\set{S}$. Consider the negated objective \(- \mathcal J(\set{S})\) and let the RKHS removal path of sample $x_i$ be defined as \(k(x,\set{S},h_i) = k_{x,\set{S}}-h_i\,k(x,x_i)e_i, h_i \in [0,1]\) from (\Cref{prop:removalDirection}).
Then the Hessian of $- \mathcal J$ evaluated at $h=0$ is given by
\begin{equation*}
\left.\nabla_{h}^2
- \mathcal J (\set{S})
\right|_{h=0} = H^{\mathrm{epi}}(\set{S}) + H^{\mathrm{ale}}(\set{S}),
\end{equation*}
where the epistemic and aleatoric Hessian contributions are given by
\begin{align*}
H^{\mathrm{epi}}(x) = & \;8(\sigma_w^2)^2\diag(c_x)A^{-1}\diag(c_x) - 6\sigma_w^2\diag(c_x\odot c_x),\\
H^{\mathrm{ale}}(x) = & \;\mathrm{mix}(\nabla_h\sigma_w^2)^\top + (\nabla_h\sigma_w^2)\mathrm{mix}^\top + 2(c_x\cdot c_x')(\nabla_h\sigma_w^2)(\nabla_h\sigma_w^2)^\top - (\|c_x\|^2+1)\nabla^2_h\sigma_w^2,
\end{align*}
with \(\nabla_h\sigma_w^2\) and \(\nabla_h^2 \sigma_w^2\) defined as in \Cref{lemma:aleatoric-hessian}, the mixed partial vector
\begin{equation*}
    \mathrm{mix} = -(2\,c_x'\odot(s_x - k(\set{S},x)) + 2\,s_x'\odot c_x),
\end{equation*}
and the following abbreviations
\begin{align*}
    c_x = A^{-1} k(\set{S},x),\quad c_x' = A^{-2}k(\set{S},x),\quad s_x = Kc_x,\quad s_x' = Kc_x'.
\end{align*}
For a multi-point target region, the Hessians \(H^{\mathrm{epi}}(\set{S})\) and \(H^{\mathrm{ale}}(\set{S})\) are obtained by summing these contributions over its test points.
\end{lemma}

\begin{proof}
We begin by revisiting the derivative of the negated objective function in \Cref{lemma:rkhs-derivative}
\begin{align}
    \frac{\partial -\mathcal{J}(\set{S})}{\partial h_i}  = - 2 k(x,x_i)(c_x)_i +  2 (c_x)_i(k_{x_i,\set{S}}c_x) - (\|c_x\|^2 + 1) \frac{\partial \sigma_w^2}{\partial h_i}.
\end{align}
We first derive the epistemic part, corresponding to the negative of the epistemic influence \(u_i(x)\). For fixed noise variance \(s=\sigma_w^2\), we have
\begin{equation}
    \left.\frac{\partial F}{\partial h_i}\right|_{h=0,s} = -2k(x,x_i)(c_x)_i + 2(c_x)_i(k_{x_i,\set{S}}c_x) = -2\sigma_w^2(c_x)_i^2 = -u_i(x),
\end{equation}
where we used \(Kc_x-k(\set{S},x)=-\sigma_w^2c_x\). To obtain the second derivative, we account for the quadratic dependence of the Gram matrix on the removal direction. Let \(P(h)=\diag(1-h)\). In a neighborhood of \(h=0\), \(P(h)\) is invertible, and we may rewrite
\begin{align}
    F(h,s) &= k(x,\set{S})P(h)\big(P(h)KP(h)+sI\big)^{-1}P(h)k(\set{S},x)\nonumber\\
    &= k(x,\set{S})\big(K+sP(h)^{-2}\big)^{-1}k(\set{S},x).
\end{align}
Writing \(Q(h,s)=K+sP(h)^{-2}\), its derivatives at \(h=0\) satisfy
\begin{equation}
    \frac{\partial Q}{\partial h_i}=2s\,e_ie_i^\top,\qquad \frac{\partial^2 Q}{\partial h_i\partial h_j}=6s\,\mathbf{1}_{\{i=j\}}e_ie_i^\top.
\end{equation}
By differentiating \(F(h,s)=k(x,\set{S})Q(h,s)^{-1}k(\set{S},x)\) twice, we obtain
\begin{align}
    \left.\frac{\partial^2F}{\partial h_i\partial h_j}\right|_{h=0,s} =& \;8s^2(c_x)_i(A^{-1})_{ij}(c_x)_j-6s\,\mathbf{1}_{\{i=j\}}(c_x)_i^2.
\end{align}
Therefore, in matrix form,
\begin{equation}
    H^{\mathrm{epi}}(x) = 8(\sigma_w^2)^2\diag(c_x)A^{-1}\diag(c_x) - 6\sigma_w^2\diag(c_x\odot c_x).
\end{equation}
For the aleatoric term, we write the negated objective as \(-\mathcal{J} = F(h,\sigma_w^2(h)) - \sigma_w^2(h)\), where \(F(h,s) = k(h)^\top (K(h)+sI)^{-1}k(h)\). Applying the total derivative decomposition to the second-order variation gives
\begin{equation}
    \frac{\partial^2 (-\mathcal{J})}{\partial h_i\,\partial h_j} = \frac{\partial^2 F}{\partial h_i\partial h_j}\bigg|_s + \frac{\partial^2 F}{\partial s\,\partial h_j}g_i + \frac{\partial^2 F}{\partial s\,\partial h_i}g_j + \frac{\partial^2 F}{\partial s^2}g_ig_j - \left(\|c_x\|^2+1\right)H_\sigma[i,j],
\end{equation}
where \(g = \nabla_h\sigma_w^2\), \(H_\sigma = \nabla_h^2\sigma_w^2\), and we used \(\frac{\partial F}{\partial s} = -\|c_x\|^2\). The first term yields \(H^{\mathrm{epi}}\). For the mixed partial we differentiate \(\frac{\partial F}{\partial h_j}\big|_{h=0,s} = -u_j(x)\) w.r.t.\ \(s\), using \(\frac{\partial c_x}{\partial s}=-c_x'\) and \(\frac{\partial s_{x,j}}{\partial s}=-s_{x,j}'\):
\begin{equation}
    \frac{\partial^2 F}{\partial s\,\partial h_j} = -\frac{\partial u_j}{\partial s} = -2c_{x,j}'(s_{x,j}-k(\set{S},x)_j) - 2c_{x,j}s_{x,j}' = \mathrm{mix}_j.
\end{equation}
The pure \(s\)-second derivative of \(F\) is
\begin{equation}
    \frac{\partial^2 F}{\partial s^2} = 2\,k(\set{S},x)^\top A^{-3}k(\set{S},x) = 2\,c_x\cdot c_x'.
\end{equation}
Collecting these contributions yields the aleatoric Hessian term
\begin{equation}
    H^{\mathrm{ale}} = \mathrm{mix}(\nabla_h\sigma_w^2)^\top + (\nabla_h\sigma_w^2)\mathrm{mix}^\top + 2(c_x\cdot c_x')(\nabla_h\sigma_w^2)(\nabla_h\sigma_w^2)^\top - (\|c_x\|^2+1)\nabla_h^2\sigma_w^2.
\end{equation}
\end{proof}

\begin{lemma}\label{lemma:RSCProof}
    Let \(H(\set{S}^{(k)}) = H^{\mathrm{epi}}(\set{S}^{(k)}) + H^{\mathrm{ale}}(\set{S}^{(k)})\) be the Hessian of the negated objective \(- \mathcal{J}\), as defined in \Cref{lemma:rkhs-hessian} for the retained dataset \(\set{S}^{(k)}\) in episode \(k\). Then for all possible removal subsets \(T^{(k)} \subseteq T \coloneqq \set{S}\setminus S' \subseteq R\), we denote by \(H_{R}(\set{S}^{(k)})\) the principal sub-matrix of the \(H(\set{S}^{(k)})\) on \(R\).
    Then, the curvature bounds across all possible removal directions are obtained by 
    \begin{equation*}
        m_R \coloneqq \inf_{\set{S}^{(k)}} \lambda_{\mathrm{min}}(-H_{R}(\set{S}^{(k)})), \qquad M_R \coloneqq \sup_{\set{S}^{(k)}} \lambda_{\mathrm{max}}(-H_R(\set{S}^{(k)})),
    \end{equation*}
    such that for all \(\set{S}^{(k)}\) and \(T^{(k)} \subseteq T \subseteq R\), the projected Hessians \(H_R(\set{S}^{(k)})\) satisfy
    \begin{equation*}
        -M_R I \leq H_{R}(\set{S}^{(k)}) \leq -m_R I.
    \end{equation*}
\end{lemma}
\begin{proof}
By definition of $m_R$ and $M_R$, for every retained subset $\set{S}^{(k)} \subseteq \set{S}'$ the principal submatrix $H_R(\set{S}^{(k)})$ satisfies
\begin{equation}\label{eq:all_principal_spectral}
    -M_R I \leq H_R(\set{S}^{(k)}) \leq -m_R I,
\end{equation}
which is precisely the final criterion in \Cref{lemma:RSCProof}.

It remains to show that this criterion implies restricted strong concavity and smoothness (RSC/RSM).
Let $L \subset L \cup U \subseteq R$ and let $h$ denote the indicator vector of the removed coordinates in $U$ relative to $L$.
Since $h$ is supported on $U \subseteq R$, the quadratic form satisfies $h^\top H(\set{S}^{(k)}) h = h^\top H_R(\set{S}^{(k)}) h$ for any retained dataset.
The uniform spectral bound \Cref{eq:all_principal_spectral} therefore says that $-\bar{\mathcal{J}}$, restricted to removal directions in $R$, is simultaneously $m_R$-strongly convex and $M_R$-smooth.
By the standard first-order characterization of strong convexity and smoothness \cite[Lemma~1.2.3]{nesterov2013introductory}, this is equal to
\begin{equation}
- \frac{M_R}{2} \|h\|_2^2 \;\leq\;
\bar{\mathcal J}(L \cup U) - \bar{\mathcal J}(L) - \langle \nabla \bar{\mathcal J}(L),\, h \rangle
\;\leq\;
- \frac{m_R}{2} \|h\|_2^2,
\end{equation}
which is the RSC/RSM condition for the outer objective.
\end{proof}

\begin{lemma}\label{lemma:inner_opti}
Let \(K\) be the Gram matrix and \(\lambda_i(K)\) its eigenvalues, and let \(\|y\|_2^2\) denote the output norm. Given the Gamma prior distribution \(p(\sigma_w^2(\set{S})|a,b) = \frac{b^a}{\Gamma(a)}(\sigma_w^2(\set{S}))^{a-1}e^{-b\sigma_w^2(\set{S})}\), with \(b>0\), suppose that the parameter \(a\) satisfies, for an admissible range \(\sigma_w^2\in[\xi_{\min},\xi_{\max}]\) with \(0<\xi_{\min}<\xi_{\max}\),
\begin{equation}
    a > 1 + \frac{1}{2}\sum_{i=1}^{|\set{S}|}\left(\frac{\xi_{\max}}{\lambda_i(K)+\xi_{\max}}\right)^2 - \frac{\xi_{\min}^2\|y\|_2^2}{(\lambda_{\max}(K)+\xi_{\max})^3}.
\end{equation}
Then the noise log MAP II objective \(\mathcal{L}(\sigma_w^2,\set{S},a,b)\) is strictly concave on \([\xi_{\min},\xi_{\max}]\) and consequently has a unique maximizer on this interval. If, additionally, the maximizer is interior, then \(\mathcal{D}(\sigma_w^2,\set{S},a,b)=0\) and \(T<0\), as required by \Cref{lemma:aleatoric-hessian}. The same conclusion holds for all retained subsets if the condition is satisfied for each of them.
\end{lemma}

\begin{proof}
We consider the second derivative of the log MAP II objective w.r.t. the noise parameter, which is given by
\begin{equation}
    \frac{\partial^2\mathcal{L}(\sigma_w^2,\set{S},a,b)}{\partial(\sigma_w^2)^2} = -y^\top A^{-3}y + \frac{1}{2}\tr(A^{-2})-\frac{a-1}{(\sigma_w^2)^2},
\end{equation}
where \(A=K+\sigma_w^2I\). Strict concavity holds if
\begin{equation}
    a > 1 + (\sigma_w^2)^2\left(\frac{1}{2}\tr(A^{-2})-y^\top A^{-3}y\right).
\end{equation}
We now upper bound the right-hand side uniformly for \(\sigma_w^2\in[\xi_{\min},\xi_{\max}]\). Since \(K\) is positive semidefinite, we obtain
\begin{align}
    (\sigma_w^2)^2\tr(A^{-2}) &= \sum_{i=1}^{|\set{S}|}\left(\frac{\sigma_w^2}{\lambda_i(K)+\sigma_w^2}\right)^2\leq\sum_{i=1}^{|\set{S}|}\left(\frac{\xi_{\max}}{\lambda_i(K)+\xi_{\max}}\right)^2.
\end{align}
Moreover, \(\lambda_{\max}(A)\leq\lambda_{\max}(K)+\xi_{\max}\), so that
\begin{align}
    (\sigma_w^2)^2y^\top A^{-3}y &\geq\frac{(\sigma_w^2)^2\|y\|_2^2}{(\lambda_{\max}(K)+\xi_{\max})^3}\geq\frac{\xi_{\min}^2\|y\|_2^2}{(\lambda_{\max}(K)+\xi_{\max})^3}.
\end{align}
Combining both bounds yields
\begin{align}
    (\sigma_w^2)^2\left(\frac{1}{2}\tr(A^{-2})-y^\top A^{-3}y\right)
    \leq &\;\frac{1}{2}\sum_{i=1}^{|\set{S}|}\left(\frac{\xi_{\max}}{\lambda_i(K)+\xi_{\max}}\right)^2-\frac{\xi_{\min}^2\|y\|_2^2}{(\lambda_{\max}(K)+\xi_{\max})^3}.
\end{align}
Thus, the condition in the lemma implies \(\frac{\partial^2\mathcal{L}}{\partial(\sigma_w^2)^2}<0\) throughout the admissible interval. This proves strict concavity and uniqueness of the interval maximizer. If the maximizer is an interior stationary point, then \(T=\frac{\partial^2\mathcal{L}}{\partial(\sigma_w^2)^2}<0\), completing the proof.
\end{proof}

\begin{lemma}\label{lemma:upper_bound_comp_difference}
 Suppose that the negated objective \(- \mathcal{J}(\cdot)\) is restricted strongly concave on \(T\), as defined in \Cref{def:rsc}, then the difference in the objective for two considered subsets of \(T\) is bounded by
\begin{equation*}
   \mathcal{J}(\set{S}\setminus L) - \mathcal{J}(\set{S}\setminus (L \cup U)) \leq \frac{1}{2m_R} \|\nabla - \mathcal{J}(\set{S}\setminus L)_U\|_2^2 
\end{equation*}
where the sets satisfy \(L \subset L \cup U \subseteq T\).
\end{lemma}
\begin{proof}
    Let us begin with the restricted strong concavity criterion
    \begin{equation}
    - \frac{m_R}{2} \|h\|_2^2 \geq \mathcal{J}(\set{S}\setminus L) - \mathcal{J}(\set{S}\setminus (L \cup U))  - \langle \nabla - \mathcal{J}(\set{S}\setminus L),h\rangle.
\end{equation}
By bringing the inner product of the gradient to the other side, we obtain
\begin{equation}
     \langle \nabla - \mathcal{J}(\set{S}\setminus L),h\rangle - \frac{m_R}{2} \|h\|_2^2 \geq \mathcal{J}(\set{S}\setminus L) - \mathcal{J}(\set{S}\setminus (L \cup U)),
\end{equation}
which we can further upper bound by the best possible subset
\begin{equation}
   \max_{h} \langle \nabla - \mathcal{J}(\set{S}\setminus L),h\rangle - \frac{m_R}{2} \|h\|_2^2 \geq \mathcal{J}(\set{S}\setminus L) - \mathcal{J}(\set{S}\setminus (L \cup U)).
\end{equation}
The unrestricted quadratic expression in the variable z supported on $U$ attains its maximum at $z=\frac{1}{m_R} \nabla(-\set{J})_U$, yielding $\|\nabla(-\set{J})_U\|_2^2\frac{1}{2m_R}$.
\begin{equation}
     \mathcal{J}(\set{S}\setminus L) - \mathcal{J}(\set{S}\setminus (L \cup U)) \leq \frac{1}{2m_R} \|\nabla - \mathcal{J}(\set{S}\setminus L)_U\|_2^2 
\end{equation}
\end{proof}

\begin{lemma}\label{lemma:gradient-dominance-progress}
Suppose the conditions of Theorem~1 hold, and let
$T_k=\set{S}\setminus\set{S}^{(k)}$ for $k=0,\ldots,q$.
Then for every $k<q$,
\begin{equation}\label{eq:f2i-greedy-progress-uq}
\bar{\mathcal J}(T_{k+1})-\bar{\mathcal J}(T_k)
\geq \frac{\rho m_R}{qM_R}
\left[U_q-\bar{\mathcal J}(T_k)\right]_+.
\end{equation}
\end{lemma}
\begin{proof}
By the definition of the realization coefficient in
\Cref{assumption:gradient-dominance}, the actual improvement from the
$k$-th discrete deletion is
\begin{align}
\bar{\mathcal J}(T_{k+1})-\bar{\mathcal J}(T_k)
&=\mathcal J(\set{S}^{(k)})-\mathcal J(\set{S}^{(k+1)})\\
&=\frac{\rho_k}{2M_R}
\|u^{(k)}(\mathcal A)+v^{(k)}(\mathcal A)\|_\infty^2\\
&\geq\frac{\rho m_R}{qM_R}
\left[U_q-\bar{\mathcal J}(T_k)\right]_+,
\end{align}
where the last step follows from
\Cref{assumption:gradient-dominance} and $\rho_k\geq\rho$.
\end{proof}

\thrmforgettingoptimality*
\begin{proof}
    Let $T^*$ be an optimal removal set with $|T^*|=q$. By the
restricted strong concavity inequality of Definition~1 applied
with $L=\emptyset$ and $U=T^*$,
\begin{align}
\bar{\mathcal J}(T^*)
&\leq \sum_{i\in T^*}g_i^{(0)}-\frac{m_Rq}{2}\\
&\leq \sum_{i=1}^{q}g_{(i)}^{(0)}-\frac{m_Rq}{2}.
\end{align}
Since $\mathcal J(\set{S}\setminus T^*)\geq0$, we also have
$\bar{\mathcal J}(T^*)\leq\mathcal J(\set{S})$.
Consequently,
\begin{equation}\label{eq:opt-bounded-by-uq}
\bar{\mathcal J}(T^*)\leq U_q.
\end{equation}
By $\rho>0$, every executed deletion has strictly positive gain;
hence $\bar{\mathcal J}(T_k)\leq\bar{\mathcal J}(T')
\leq\bar{\mathcal J}(T^*)\leq U_q$ for every $k\leq q$.
Set $G_k=U_q-\bar{\mathcal J}(T_k)\geq0$ and
$c=\rho m_R/(qM_R)>0$.
By \Cref{eq:f2i-greedy-progress-uq},
\begin{equation}
G_{k+1}\leq (1-c)G_k.
\end{equation}
Since $\rho>0$ and $q\geq1$, the achieved gain is positive, so
$U_q\geq\bar{\mathcal J}(T^*)\geq\bar{\mathcal J}(T')>0$.
Thus $G_0>0$, and $G_1\geq0$ implies $0<c\leq1$.
Since $T_0=\emptyset$ and
$\bar{\mathcal J}(T_0)=0$, repeated application yields
\begin{align}
G_q &\leq (1-c)^qU_q
\leq e^{-cq}U_q
=e^{-\gamma}U_q.
\end{align}
Therefore,
\begin{align}
\bar{\mathcal J}(T')
&\geq (1-e^{-\gamma})U_q\\
&\geq (1-e^{-\gamma})\bar{\mathcal J}(T^*),
\end{align}
which proves the claim.
\end{proof}

\subsection{Empirical Diagnostics of the Near-Optimality Conditions}
\label{sec:appendix_optimality_diagnostics}

We empirically investigate the conditions underlying the near-optimality guarantee in \Cref{thrm:forgetting_optimality}: restricted strong concavity (RSC), positive realized improvement
under binary deletion, and restricted gradient dominance. We evaluate these conditions for the symmetric and asymmetric synthetic examples and the Twitter experiment, varying the Gamma-prior parameter $a$.

\paragraph{Restricted curvature.}
Following \Cref{lemma:rkhs-hessian}, we evaluate the Hessian of the negated uncertainty objective at each removal step $k$, denoted by \(H^{(k)} = H^{\mathrm{epi},(k)} + H^{\mathrm{ale},(k)}\).
Let $T$ denote the final set of removed observations and $R_k=T\cap\mathcal{S}^{(k)}$ those remaining at iteration $k$. We consider the principal submatrix $H_{R_k}^{(k)}$ and report its extremal eigenvalues,
\begin{equation}
    \label{eq:restrictedCurvature}
    \lambda_{\min}(H_{R_k}^{(k)}) \leq \lambda_{\max}(H_{R_k}^{(k)}).
\end{equation}
A negative largest eigenvalue indicates local strong concavity along the considered removal coordinates.
These eigenvalues provide empirical curvature diagnostics and show constraint violations, but do not alone establish the full RSC condition of \Cref{def:rsc}.

\begin{figure}[H]
    \centering
    \includegraphics[width=\textwidth]{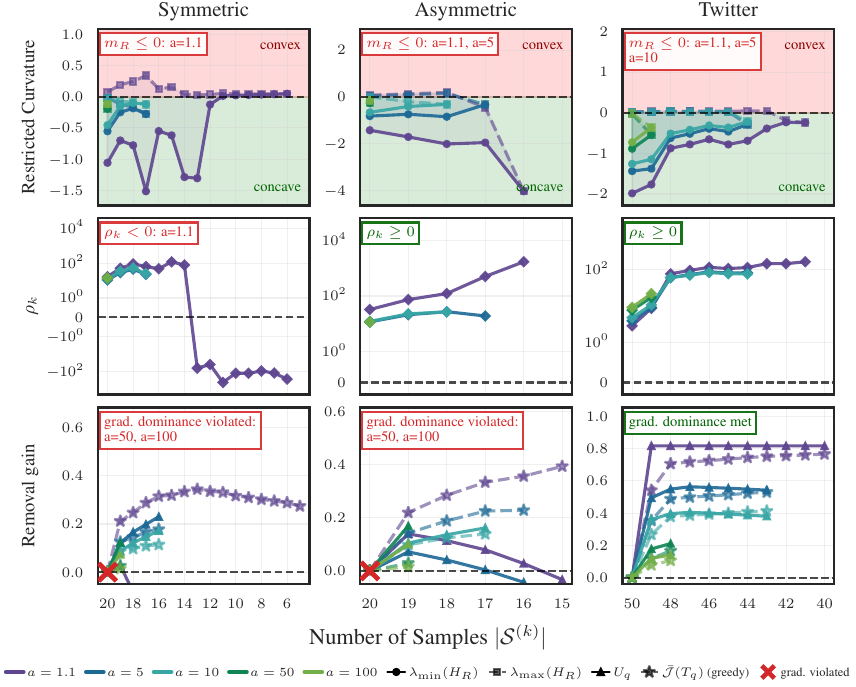}
    \caption{
    \emph{Empirical diagnostics of the near-optimality conditions} for the symmetric and asymmetric synthetic examples and the Twitter experiment, for different Gamma-prior parameters $a$.
    \emph{Top:} Extremal eigenvalues of the Hessian restricted to the remaining removal coordinates; positive largest eigenvalues indicate a violation of local concavity. \emph{Middle:} Removal coefficients $\rho_k$, where negative values indicate that discrete removal increases uncertainty.
    \emph{Bottom:} Upper-bound quantity $U_q$ and accumulated greedy removal gain $\bar{\mathcal{J}}(T_k)$. Red crosses mark detected violations of restricted gradient dominance. The diagnostics illustrate the dependence of the sufficient conditions on the prior parameter $a$; local curvature measurements do not certify global discrete RSC.
    }
    \label{fig:res_curv_twitter}
\end{figure}

\paragraph{Gradient dominance.}
To assess the effect of discrete removals, we compute the removal coefficient
\begin{equation}
    \rho_k = \frac{2M_R\left[\mathcal{J}(\mathcal{S}^{(k)})-\mathcal{J}(\mathcal{S}^{(k+1)})\right]}{\|u^{(k)}+v^{(k)}\|_\infty^2},
\end{equation}
where $\rho_k>0$ indicates an actual reduction in uncertainty. We additionally evaluate the restricted gradient-dominance condition
\begin{equation}
    \|u^{(k)}+v^{(k)}\|_\infty^2 \geq \frac{2m_R}{q} \left[U_q-\bar{\mathcal{J}}(T_k)\right]_+,
\end{equation}
where $U_q$ upper-bounds the optimal improvement under the discrete RSC condition. The bottom row of \Cref{fig:res_curv_twitter} compares $U_q$ with the accumulated greedy removal gain and marks iterations violating gradient dominance.

\paragraph{Observations.}\looseness=-1
As shown in \Cref{fig:res_curv_twitter}, the prior parameter $a$ substantially influences curvature and removal behavior. Increasing $a$ generally leads to more stable, locally concave curvature, but can also reduce the magnitude of the influence gradients, eventually violating gradient dominance for excessively large $a$. Hence, favorable prior parameters balance sufficient curvature with informative removal gradients. For small $a$, positive restricted curvature and even negative realized gains $\rho_k$ can occur, as observed in the symmetric example for $a=1.1$. Conversely, gradient dominance is violated for larger $a$ in the synthetic examples despite favorable curvature, while the Twitter experiments satisfy the gradient-dominance criterion even when local concavity is violated.

\looseness=-1
These observations highlight that restricted curvature, positive realized improvement, and gradient dominance capture distinct properties of the removal process. Importantly, we observe empirical improvements even when not all theoretical conditions are satisfied. Thus, these conditions are sufficient for the near-optimality guarantee, but are not necessary for successful data removal in practice.

\end{document}